\documentclass[final,5p,times,twocolumn,numbers]{elsarticle}

\usepackage{amssymb}
\usepackage{lipsum}
\usepackage{amsmath}
\usepackage{array}
\usepackage{tabularx}
\usepackage{supertabular}
\usepackage{pdflscape}
\usepackage{rotating}
\usepackage{threeparttable}
\usepackage{booktabs}
\usepackage{longtable}
\usepackage{xltabular}
\usepackage{url} 
\usepackage{tikz}
\usepackage{hyperref}
\usepackage{array}
\usepackage{lscape}
\usepackage{graphicx}
\usepackage{algorithm}
\usepackage{algpseudocode}
\usepackage{tikz}
\usetikzlibrary{shapes,arrows,positioning}
\usetikzlibrary{positioning, shapes, arrows.meta}
\usetikzlibrary{arrows.meta, positioning}
\usepackage{forest}   
\usepackage{xcolor}   

\usepackage{adjustbox}
\usepackage{afterpage}
\usepackage{xltabular}
\newcolumntype{Y}{>{\RaggedRight\arraybackslash}X}
\usepackage{float} 
\usepackage{pifont}   
\usepackage{amssymb}  
\usepackage{ragged2e} 
\usepackage{amsmath}  
\usepackage{fancyhdr}
\usetikzlibrary{mindmap,shadows,trees}
\usetikzlibrary{mindmap,trees} 

\journal{Information Fusion}
\usepackage{lineno}

\begin{document}
\begin{frontmatter}


\title{Vision-Language-Action (VLA) Models: Concepts, Progress, Applications and Challenges}

\author[1]{Ranjan Sapkota}
\author[2]{Yang Cao}
\author[3]{Konstantinos I. Roumeliotis}
\author[1]{Manoj Karkee}

\affiliation[1]{organization={Cornell University},
    addressline={Biological \& Environmental Engineering},
    city={Ithaca},
    state={New York},
    country={USA}}

\affiliation[2]{organization={The Hong Kong University of Science and Technology},
    addressline={Department of Computer Science and Engineering},
    country={Hong Kong}}

\affiliation[3]{organization={University of the Peloponnese},
    addressline={Department of Informatics and Telecommunications},
    country={Greece}}

\cortext[cor1]{Ranjan Sapkota}
\ead{rs2672@cornell.edu}

\begin{abstract}
Vision-Language-Action (VLA)  models mark a transformative advancement in artificial intelligence, aiming to unify perception, natural language understanding, and embodied action within a single computational framework. This foundational review presents a comprehensive synthesis of recent advancements in Vision-Language-Action models, systematically organized across five thematic pillars that structure the landscape of this rapidly evolving field. We begin by establishing the conceptual foundations of VLA systems, tracing their evolution from cross-modal learning architectures to generalist agents that tightly integrate vision-language models (VLMs), action planners, and hierarchical controllers. Our methodology adopts a rigorous literature review framework, covering over 80 VLA models published in the past three years. Key progress areas include architectural innovations, efficient training strategies, and real-time inference accelerations. We explore diverse application domains such as autonomous vehicles, medical and industrial robotics, precision agriculture, humanoid robotics, and augmented reality. The review further addresses major challenges across real-time control, multimodal action representation, system scalability, generalization to unseen tasks, and ethical deployment risks. Drawing from the state-of-the-art, we propose targeted solutions including agentic AI adaptation, cross-embodiment generalization, and unified neuro-symbolic planning. We outline a forward-looking roadmap where VLA models, VLMs, and agentic AI converge to strengthen socially aligned, adaptive, and general-purpose embodied agents. This work, therefore, is expected to serve as a foundational reference for advancing intelligent, real-world robotics and artificial general intelligence. The project repository is available on GitHub(\href{ https://github.com/Applied-AI-Research-Lab/Vision-Language-Action-Models-Concepts-Progress-Applications-and-Challenges }{Source Link})
\end{abstract}




\end{frontmatter}

\section{Introduction}
\label{introduction}
Before Vision-Language-Action (VLA) models were developed, progress in robotics and artificial intelligence happened mostly in separate domains: vision systems that could acquire, interpret and recognize images \cite{donahue2015long,hanson2014visions,radford2021learning}, language systems that could understand and generate text \cite{sutskever2011generating,radford2018improving}, and action systems that could control movement \cite{duarte2018action}. These isolated systems worked well on their own, but struggled to work together and generalize to novel scenarios or adapt to the complexity and unpredictability of  real-world challenges \cite{doveh2023teaching, cao2020behind}.

\begin{figure}[ht!]
\centering
\includegraphics[width=0.90\linewidth]{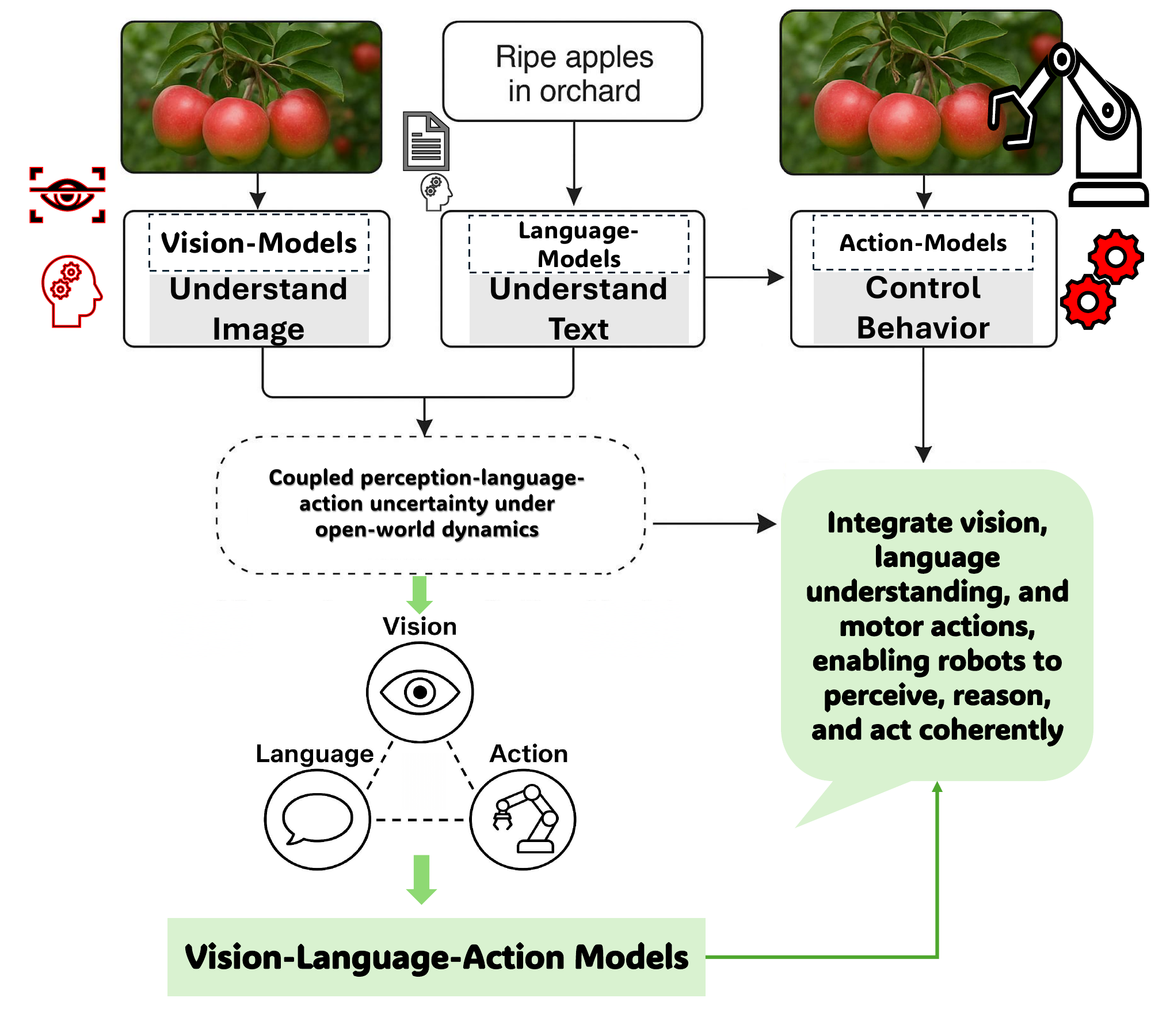}
\caption{\textbf{Evolution from isolated modalities to unified Vision–Language–Action models.} Integrated perception, language, and action enable adaptive, generalizable embodied intelligence.}
\label{fig:one}
\end{figure}

As illustrated in Figure \ref{fig:one}, traditional computer vision models, primarily based on convolutional neural networks (CNNs), were tailored for narrowly specified tasks such as object detection~\cite{cao2023coda,misra2021end,cao2024collaborative,3dgsdet,qi2019deep} or classification~\cite{simonyan2014very,he2016deep,huang2017densely,gao2019res2net}, requiring extensive labeled datasets and cumbersome retraining for even slight shifts in environment or objectives \cite{shin2016deep, gu2018recent}. These vision models could “see” (e.g., identifying apples in an orchard, as shown in Figure \ref{fig:one}) but lacked any understanding of language or the ability to convert visual insights into desired actions. Language models, particularly large language models (LLMs), revolutionized text-based understanding and generation \cite{chang2024survey}; however, they remained restricted to processing language without the capability to perceive or reason about the physical world \cite{huang2023language} (“Ripe apples in orchard” in Figure \ref{fig:one} exemplifies this limitation). Meanwhile, action-based systems in robotics, relying heavily on hand-crafted policies or reinforcement learning \cite{mohammed2020review}, enabled specific behaviors like object manipulation but demanded painstaking engineering and failed to generalize beyond specifically designed scenarios \cite{luo2024precise}.

Despite progress with VLMs, which achieved impressive multi-modal understanding by combining vision and language \cite{sapkota2025review, chen2024spatialvlm, Zhu2024UNIT, chen2024internvl, Qwen2.5-VL, sapkota2025object}, there remained a clear integration gap: the inability to generate or execute coherent actions based on multi-modal input \cite{ma2024survey, li2025benchmark}. As further visualized in Figure \ref{fig:one} , most AI systems are specialized in one or two modalities such as vision-language, vision-action, or language-action, which struggled to fully integrate all three into a unified, end-to-end frameworks. Consequently, robots could recognize objects visually (“apple”), understand a corresponding textual instruction (“pick the apple”), or perform a predefined motor action (grasping), yet integrating and performing all these abilities into fluid, adaptable behavior has been missing. The result was a pipeline that could not flexibly adapt to new tasks or environments, leading to brittle generalization and labor-intensive engineering efforts. This limitation highlighted a critical bottleneck in embodied AI: without systems that could jointly perceive, understand, and act, intelligent autonomous behavior remained a challenging goal.

\begin{figure}[h!]
\centering
\begin{tikzpicture}[
    scale=0.70,
    transform shape,
    mindmap,
    every node/.style={concept, circular drop shadow, font=\sffamily, text=black},
    root concept/.append style={
        font=\Large,
        minimum size=3.2cm,
        text width=2.8cm,
        fill=blue!20,
        line width=0.8pt,
        text=black
    },
    level 1 concept/.append style={
        sibling angle=60,
        level distance=4cm,
        text width=2.35cm,
        font=\scriptsize\sffamily,
        line width=0.6pt
    }
]

\node [root concept] {VLA\\Concepts}
  [clockwise from=270]
  child[concept color=green!30]   { node {Foundation:\\What are VLAs?} }
  child[concept color=violet!30]  { node {Evolution\\Timeline} }
  child[concept color=orange!25]  { node {Multimodal\\Integration} }
  child[concept color=red!25]     { node {Tokenization\\\& Encoding} }
  child[concept color=cyan!25]    { node {Learning\\Paradigms} }
  child[concept color=yellow!30]  { node {Adaptive\\Execution} };

\end{tikzpicture}
\caption{Mind map of core VLA concepts. Each color-coded branch highlights a foundational dimension: definitions (foundation), historical evolution, multimodal integration, tokenization and encoding, learning paradigms, and adaptive execution in embodied settings.}
\label{fig:vla_concepts}
\end{figure}

\begin{figure*}[h!]
\centering
\scalebox{0.85}{ 
\begin{tikzpicture}
  \path[mindmap,concept color=orange!30,text=black]
    node[concept] {Vision-Language-Action (VLA) Models}
    [clockwise from=0]
    
    child[concept color=green!30, level distance = 7 cm] {
      node[concept] {VLA Progress \& Training Efficiency}
      [clockwise from=20]
      child { node[concept] {Architectural Innovations} }
      child { node[concept] {VLA Applications} }
      child { node[concept] {Data-Efficient Learning} }
      child { node[concept] {Parameter Efficiency} }
      child { node[concept] {Acceleration Methods} }
    }
    
child[concept color=red!35, level distance=11cm] {
  node[concept] {VLA Challenges}
      [clockwise from=210]
      child { node[concept] {Inference Constraints} }
      child { node[concept] {Multimodal Action} }
      child { node[concept] {Safety \& Bias} }
      child { node[concept] {System Complexity} }
      child { node[concept] {Compute Demands} }
      child { node[concept] {Generalization Gaps} }
      child { node[concept] {Environmental Robustness} }
      child { node[concept] {Ethics \& Society} }
    };
\end{tikzpicture}
}
\caption{Mind map illustrating VLA model ecosystem: progress in training efficiency (architectural innovations, data/parameter efficiency, acceleration) alongside key challenges (inference constraints, multimodal action, safety, generalization, ethics) that must be overcome for scalable real-world deployment.}
\label{fig:vla_progress_challenges}
\end{figure*}
The pressing need to bridge these gaps catalyzed the emergence of VLA models. VLA models, conceptualized around 2021-2022, and pioneered by efforts such as Google DeepMind’s Robotic Transformer 2 (RT-2) \cite{zitkovich2023rt}, introduced a transformative architecture that unified perception, reasoning, and control within a single framework. As a solution to the limitations outlined in Figure \ref{fig:vla_concepts}, VLAs integrate vision inputs, language comprehension, and motor control capabilities, enabling embodied agents to perceive their surroundings, understand complex instructions, and execute appropriate actions dynamically. Early VLA approaches achieved this integration by extending vision-language models to include action tokens numerical or symbolic representations of robot motor commands, thereby allowing the model to learn from paired vision, language, and trajectory data \cite{ma2024survey}. This methodological innovation dramatically improved robots' ability to generalize to unseen objects, interpret novel language commands, and perform multi-step reasoning in unstructured environments \cite{jeong2024survey}.

VLA models represent a transformative step in the development of unified multi-modal intelligence, overcoming the long-standing limitations of treating vision, language, and action as separate domains \cite{ma2024survey}. By leveraging internet-scale datasets that integrate visual, linguistic, and behavioral information, VLAs empower robots to not only recognize and describe their environments but also to reason contextually and execute appropriate actions in complex, dynamic settings \cite{xu2024survey}. The progression illustrated in Figures \ref{fig:vla_concepts} and  \ref{fig:vla_progress_challenges} from isolated vision, language, and action systems to an integrated VLA paradigm-captures a fundamental shift toward the development of truly adaptive and generalizable embodied agents. In light of the transformative potential of this paradigm, a comprehensive and critically informed review of the current literature is both timely and essential. First, such a review is necessary to clarify the foundational concepts and architectural principles that distinguish VLAs from their predecessors. Second, it provides a structured account of the rapid progress and key milestones in the field, enabling researchers and practitioners to appreciate the trajectory of algorithmic and technological advancements. Third, an in-depth review is essential for mapping the diverse range of real-world applications - from household robotics to industrial automation and assistive technologies-where VLAs are already demonstrating transformative potential. Furthermore, by critically examining the current challenges, such as data efficiency, safety, generalization, and ethical considerations, the review identifies barriers that must be addressed for widespread deployment. Finally, synthesizing these insights helps to inform the broader AI and robotics communities about emerging research directions and practical considerations, fostering collaboration and innovation. 

In this review, we systematically analyze the foundational principles of VLA models. Additionally, we discuss their developmental progress and technical challenges. Our objective is to consolidate the current understanding and applications of VLAs while identifying limitations and proposing future directions for their evolution.The review begins with a detailed examination of key conceptual foundations (Figure~\ref{fig:vla_concepts}), including the definition of VLA models, their historical evolution, mechanisms for multimodal integration, and unified tokenization and representation strategies spanning vision, language, and action. These conceptual descriptions set the stage for understanding how VLAs are structured and function across modalities.

Building upon this description, we present a unified view of recent progress and training efficiency strategies (Figure~\ref{fig:vla_progress_challenges}). This includes architectural innovations adopted and extended within VLA models, along with data-efficient learning frameworks, parameter-efficient modeling techniques, and model acceleration strategies originally developed in broader machine learning and robotics contexts. Together, these advances are critical for scaling VLA systems to real-world applications.

Following this, we present a comprehensive discussion of the current limitations encountered by VLA systems (Figure~\ref{fig:vla_progress_challenges}), many of which reflect broader challenges in embodied AI and robotics but arise in distinct and compounded forms due to the tight integration of vision, language, and action.
The limitations to be discussed include inference bottlenecks, safety concerns, high computational demands, limited generalization, and ethical implications. We not only highlight these pressing challenges but also provide an analytical discussion on potential solutions to address them.

Together, these three figures offer a visual depiction that lays out the framework and supports the textual analysis presented in this review. By outlining the conceptual landscape, recent innovations, and open challenges, this work aims to guide future research and encourage the development of more robust, efficient, and ethically grounded VLA systems.

Figure~\ref{fig:vla_paper_architecture} summarizes the overall structure and logical flow of this review and illustrates how the manuscript is organized to provide a comprehensive and systematic analysis of VLA research. As depicted in the figure, the paper progresses from foundational concepts to recent advances, applications, challenges, and future research directions, ensuring a coherent narrative across sections. To construct this architecture, an extensive and rigorous literature search was conducted using the primary keywords “Vision–Language–Action” and “Vision–Language Models,” together with the commonly used abbreviation “VLA.” These keywords were employed to retrieve candidate studies from major academic and technical repositories, including Hugging Face, arXiv, ScienceDirect, Nature, IEEE Xplore, Wiley, and Springer Nature. The resulting corpus was further refined through manual screening to ensure relevance, technical depth, and alignment with the scope of this review. Only articles that directly contributed to the understanding of VLA concepts, methodological progress, application domains, and open challenges were retained. This multi-stage filtering process enabled a balanced coverage of both foundational and state-of-the-art works while avoiding peripheral or loosely related studies. As a result, Figure~\ref{fig:vla_paper_architecture} not only reflects the thematic organization of the paper but also embodies the underlying review methodology. 

This paper follows a structured, hierarchical organization that systematically develops the foundations, evolution, and implications of VLA models, as summarized in Figure~\ref{fig:vla_paper_architecture}. The presentation begins with an introduction that motivates embodied intelligence and the emergence of VLAs, followed by a concepts section that establishes core principles including multimodal integration, tokenization, learning paradigms, and real-time control. Building on these foundations, the progress section examines architectural innovations, training and efficiency advancements, and parameter-efficient acceleration strategies. The applications section then grounds these developments in real-world domains, including humanoid robotics, autonomous vehicles, healthcare, agriculture, industrial systems, and interactive AR navigation. This is followed by a focused analysis of key challenges, including real-time inference, safety, generalization, system integration, and ethical considerations. The paper concludes with a future roadmap that distills cross-cutting research directions in continual learning, scalability, interpretability, and embodied intelligence.

\begin{figure}[H]
\centering
\footnotesize

\definecolor{cRoot}{RGB}{35,55,90}
\definecolor{cIntro}{RGB}{173,216,230} 
\definecolor{cConcept}{RGB}{190,225,170} 
\definecolor{cProgress}{RGB}{255,218,185} 
\definecolor{cApps}{RGB}{221,204,255} 
\definecolor{cChallenges}{RGB}{255,200,200}
\definecolor{cDiscuss}{RGB}{255,242,178} 
\definecolor{cFuture}{RGB}{200,235,210} 
\definecolor{cEnd}{RGB}{200,200,200} 

\resizebox{0.95\columnwidth}{!}{%
\begin{forest}
  for tree={
    grow=east,
    draw,
    rounded corners,
    line width=0.55pt,
    anchor=west,
    parent anchor=east,
    child anchor=west,
    edge={-Latex, line width=0.75pt},
    edge path={
      \noexpand\path[\forestoption{edge}] (!u.parent anchor) -- +(.55em,0) |- (.child anchor)\forestoption{edge label};
    },
    font=\sffamily\footnotesize,
    minimum height=1.05em,
    s sep=4.2mm,
    l sep=7mm,
    text width=3.05cm, 
    align=center,
    inner sep=1.6pt
  },
  [,phantom 
    [{\shortstack{Paper Architecture}}, fill=cRoot, text=white, font=\sffamily\bfseries\footnotesize, text width=3.25cm
      [{\shortstack{Conclusion}}, fill=cEnd, text=black]
      [{\shortstack{Future\\Roadmap}}, fill=cFuture, text=black
        [{\shortstack{Cross-cutting themes:\\continual learning, recovery, \\ interaction}}, fill=cFuture!85, text width=3.80cm]
        [{\shortstack{Safety, ethics,\& human-\\centered alignment}}, fill=cFuture!85, text width=3.80cm]
        [{\shortstack{Evaluation beyond task success \\ : safety/energy/recovery}}, fill=cFuture!85, text width=3.80cm]
        [{\shortstack{Cross-embodiment transfer\\\& morphology-agnostic skills}}, fill=cFuture!85, text width=3.80cm]
        [{\shortstack{Efficiency \& scalability\\for real-time edge deployment}}, fill=cFuture!85, text width=3.80cm]
        [{\shortstack{World models for\\physical/causal reasoning}}, fill=cFuture!85, text width=3.80cm]
        [{\shortstack{Hierarchical neuro-symbolic \\ planning forinterpretability}}, fill=cFuture!85, text width=3.80cm]
        [{\shortstack{Agentic self-supervised\\lifelong learning \& adaptation}}, fill=cFuture!85, text width=3.80cm]
        [{\shortstack{Foundation models as `cortex''\\ for embodied perception}}, fill=cFuture!85, text width=3.80cm]
      ]
      [{\shortstack{Discussion}}, fill=cDiscuss, text=black
        [{\shortstack{Potential Solutions}}, fill=cDiscuss!85, text width=3.80cm]
      ]
      [{\shortstack{Challenges \&\\Limitations}}, fill=cChallenges, text=black
        [{\shortstack{Robustness \&ethical challenges\\in deployment}}, fill=cChallenges!85, text width=3.80cm]
        [{\shortstack{System integration complexity \&\\computational demands}}, fill=cChallenges!85, text width=3.80cm]
        [{\shortstack{Dataset bias, grounding,\\\& generalization\\to unseen tasks}}, fill=cChallenges!85, text width=3.80cm]
        [{\shortstack{Multimodal action representation \\ \&safety assurance}}, fill=cChallenges!85, text width=3.80cm]
        [{\shortstack{Real-time inference constraints}}, fill=cChallenges!85, text width=3.80cm]
      ]
      [{\shortstack{Applications of\\VLA Models}}, fill=cApps, text=black
        [{\shortstack{Interactive AR navigation}}, fill=cApps!85, text width=3.80cm]
        [{\shortstack{Precision \&  agriculture}}, fill=cApps!85, text width=3.80cm]
        [{\shortstack{Healthcare \& medical robotics}}, fill=cApps!85, text width=3.80cm]
        [{\shortstack{Industrial robotics}}, fill=cApps!85, text width=3.80cm]
        [{\shortstack{Autonomous vehicle systems}}, fill=cApps!85, text width=3.80cm]
        [{\shortstack{Humanoid robotics}}, fill=cApps!85, text width=3.80cm]
      ]
      [{\shortstack{Progress in VLA Models}}, fill=cProgress, text=black
        [{\shortstack{Parameter-efficient methods \&\\acceleration}}, fill=cProgress!85, text width=3.80cm]
        [{\shortstack{Training \& efficiency \\ advancements}}, fill=cProgress!85, text width=3.80cm]
        [{\shortstack{Architectural innovations}}, fill=cProgress!85, text width=3.80cm]
      ]
      [{\shortstack{Concepts of\\VLA Models}}, fill=cConcept, text=black
        [{\shortstack{Adaptive control \& real-time \\ execution}}, fill=cConcept!85, text width=3.80cm]
        [{\shortstack{Learning paradigms: data \\ sources  \& training strategies}}, fill=cConcept!85, text width=3.80cm]
        [{\shortstack{Tokenization \& representation}}, fill=cConcept!85, text width=3.80cm]
        [{\shortstack{Multimodal integration: from \\ pipelines to\\unified agents}}, fill=cConcept!85, text width=3.80cm]
        [{\shortstack{Evolution \& timeline}}, fill=cConcept!85, text width=3.80cm]
      ]
      [{\shortstack{Introduction}}, fill=cIntro, text=black]
    ]
  ]
\end{forest}%
} 

\vspace{-0.4em}
\caption{Flow and structure of this paper, from conclusion and future roadmap back through applications, progress, concepts, to introduction.}
\label{fig:vla_paper_architecture}
\vspace{-0.6em}
\end{figure}

\section{Concepts of Vision-Language-Action Models}
VLA models represent a class of intelligent systems that jointly process visual inputs, interpret natural language instructions, and generate executable action representations that can be instantiated on physical robotic hardware operating in dynamic environments. Technically, VLAs combine vision encoders (e.g., CNNs, ViTs), language models (e.g., LLMs, transformers), and policy modules or planners to achieve task-conditioned control. These models typically build upon multimodal fusion techniques established in vision-language models, such as cross-attention, concatenated embeddings, or token unification, and extend them to align sensory observations, linguistic instructions, and action representations.

\begin{figure*}[ht!]
\centering
\includegraphics[width=0.80\linewidth]{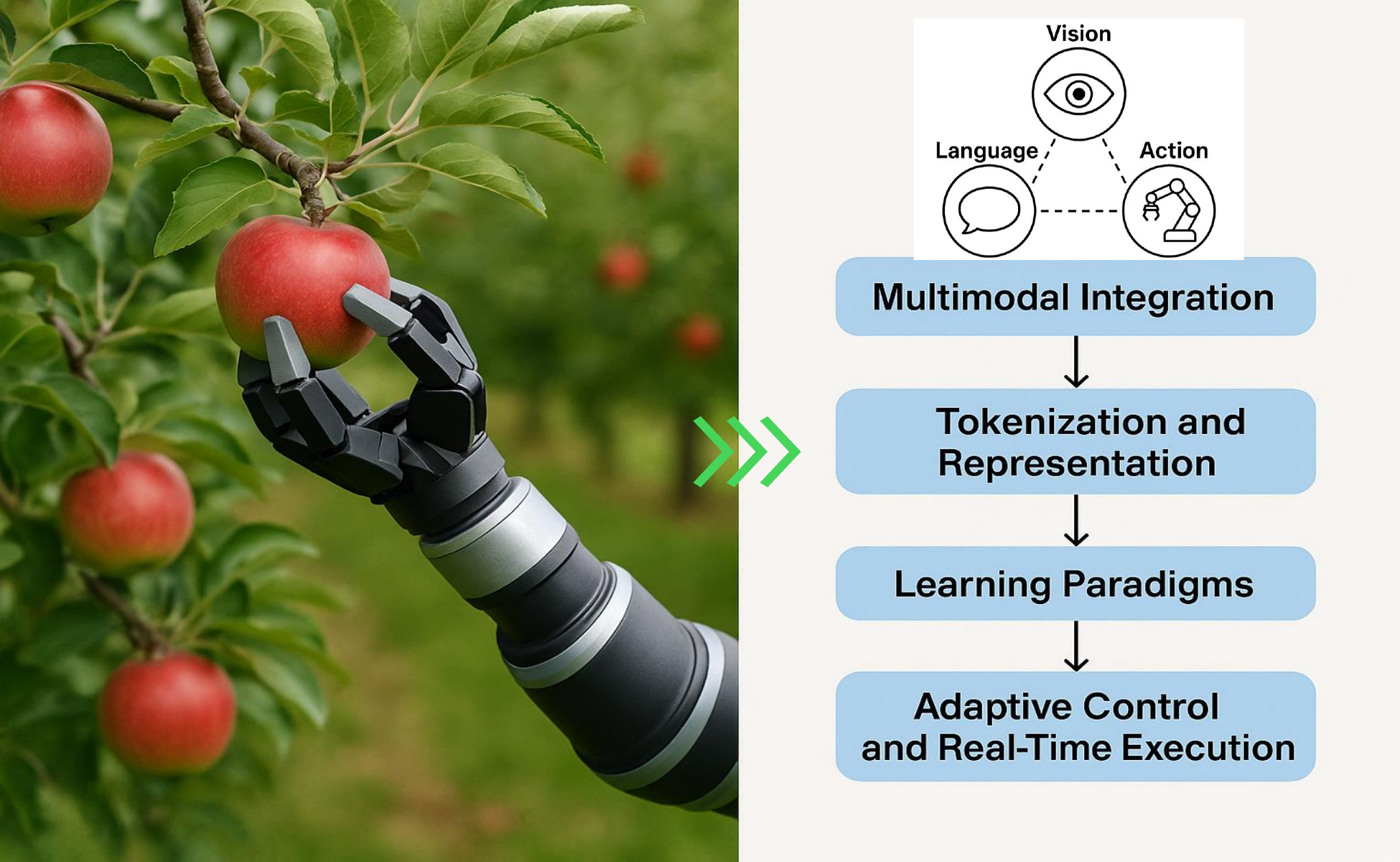}
\caption{\textbf{Foundational Concept of VLA Models (in an Apple-Picking Scenario)} This illustration depicts a robotic arm autonomously picking a ripe apple in an orchard, guided by a VLA model. On the right, a flowchart outlines the four key stages of VLA models: Multimodal Integration, Tokenization and Representation, Learning Paradigms, and Adaptive Control and Real-Time Execution.}
\label{fig:concepts}
\end{figure*}

Unlike traditional visuomotor pipelines, VLAs support semantic grounding \cite{lyre2024understandingaisemanticgrounding}, enabling context-aware reasoning \cite{Waite3722033}, affordance detection \cite{GU202136}, and temporal planning \cite{liu2025codrivevlmvlmenhancedurbancooperative}. A typical VLA model observes the environment through camera or sensor data, interprets goals expressed in language (e.g., “pick up the red apple”) (Figure \ref{fig:concepts}), and outputs low-level or high-level action sequences that could be implemented by automated systems to perform the action. Recent advancements integrate imitation learning, reinforcement learning, or retrieval-augmented modules to improve sample efficiency and generalization. This review examines how VLA models have evolved from foundational fusion architectures to general-purpose agents capable of real-world deployment across robotics, navigation, and human-AI collaboration.

VLA models are multi-modal artificial intelligence systems that unify visual perception, language comprehension, and physical action generation into a single framework. These models enable robots or AI agents to interpret sensory inputs (e.g., images, text), understand contextual meaning, and autonomously execute tasks in real-world environments - all through end-to-end learning and action rather than isolated subsystems. As shown conceptually in Figure \ref{fig:concepts}, VLA models bridge the historical disconnect between visual recognition, language comprehension, and/or motor execution that limited the capabilities of earlier robotic and AI systems. 

\subsection{Evolution and Timeline}
The rapid development of VLA models from 2022-2025 demonstrates three distinct evolutionary phases:
\begin{enumerate}
    \item  \textbf{Foundational Integration (2022–2023).} Early VLAs established basic visuomotor coordination through multi-modal fusion architectures. \cite{shridhar2022cliport} first combined CLIP embeddings with motion primitives, while \cite{reed2022generalist} demonstrated generalist capabilities across 604 tasks. \cite{brohan2022rt} achieved 97\% success rates in manipulation through scaled imitation learning, and \cite{jiang2022vima} introduced temporal reasoning via transformer-based planners. By 2023, \cite{zitkovich2023rt} enabled visual chain-of-thought reasoning, and \cite{chi2023diffusion} advanced stochastic action prediction through diffusion processes. These foundations addressed low-level control but lacked compositional reasoning, the ability to decompose complex tasks into reusable, semantically grounded sub-actions and recombine them across novel contexts-prompting subsequent innovations in affordance grounding \cite{zhao2023learning,huang2023voxposer}.

    \item \textbf{Specialization and Embodied Reasoning (2024).} Second-generation VLAs incorporated domain-specific inductive biases. \cite{yue2024deer} enhanced few-shot adaptation through retrieval-augmented training, while \cite{zhang2024uni} optimized navigation via 3D scene-graph integration. \cite{dey2024revla} introduced reversible architectures for memory efficiency, and \cite{wei2024occllama} addressed partial observability with physics-informed attention. Simultaneously, \cite{arai2025covla} improved compositional understanding through object-centric disentanglement, and \cite{zhou2025opendrivevla} extended applications to autonomous driving via multi-modal sensor fusion. However, these advances required new benchmarking methodologies \cite{xu2024survey}.

    \item  \textbf{Generalization and Safety-Critical Deployment (2025).} Latest systems prioritize robustness and human alignment. \cite{zhang2025safevla} integrated formal verification for risk-aware decisions, while \cite{ding2025humanoid} demonstrated whole-body control through hierarchical VLAs. \cite{budzianowski20edgevla} optimized compute efficiency for embedded deployment, and \cite{li2024cogact} combined neural-symbolic reasoning for causal inference. Emerging paradigms like \cite{li2024improving}'s affordance chaining and \cite{bjorck2025gr00t}'s sim-to-real transfer learning address cross-embodiment challenges, while \cite{lin2024showui} bridges VLAs with human-in-the-loop interfaces through natural language grounding.

\end{enumerate}

Figure~\ref{fig:vla-timeline} presents a comprehensive timeline highlighting the evolution of 45 VLA models developed between 2022 and 2025. The earliest VLA systems, including CLIPort~\cite{shridhar2022cliport}, Gato~\cite{reed2022generalist}, RT-1~\cite{brohan2022rt}, and VIMA~\cite{jiang2022vima}, laid the foundation by combining pretrained vision-language representations with task-conditioned policies for manipulation and control. These early VLA systems were followed by ACT~\cite{zhao2023learning}, RT-2~\cite{zitkovich2023rt}, and VoxPoser~\cite{huang2023voxposer}, which integrated visual chain-of-thought reasoning and affordance grounding. Models like Diffusion Policy~\cite{chi2023diffusion} and Octo~\cite{team2024octo} introduced stochastic modeling and scalable data pipelines. In 2024, systems such as Deer-VLA~\cite{yue2024deer}, ReVLA~\cite{dey2024revla}, and Uni-NaVid~\cite{zhang2024uni} added domain specialization and memory-efficient designs, while Occllama~\cite{wei2024occllama} and ShowUI~\cite{lin2024showui} tackled partial observability and user interaction. The trajectory continued with robotics-focused VLAs like Quar-VLA~\cite{ding2024quar} and RoboMamba~\cite{liu2024robomamba}. Recent innovations emphasize generalization and deployment: SafeVLA~\cite{zhang2025safevla}, Humanoid-VLA~\cite{ding2025humanoid}, and MoManipVLA~\cite{wu2025momanipvla} incorporate verification, full-body control, and memory systems. Models such as Gr00t N1~\cite{bjorck2025gr00t} and SpatialVLA~\cite{qu2025spatialvla} further bridge sim-to-real transfer and spatial grounding. This timeline illustrates how VLAs have advanced from modular learning to general-purpose, safe, and embodied intelligence.

\begin{figure*}[ht!]
\centering
\scalebox{0.82}{
\begin{tikzpicture}[
    timeline/.style={very thick, draw=black!75},
    continue/.style={densely dashed, very thick, draw=black!45},
    yearTick/.style={very thick, draw=black!70},
    yearText/.style={font=\bfseries\small, text=black!85},
    label/.style={align=center, font=\scriptsize, text=black!90, text width=2.75cm, inner sep=1pt},
    event2022/.style={circle, draw=black!85, fill=black!15, very thick, minimum size=8mm},
    event2023/.style={circle, draw=blue!90!black, fill=blue!18, very thick, minimum size=8mm},
    event2024/.style={circle, draw=red!85!black, fill=red!18, very thick, minimum size=8mm},
    event2025/.style={circle, draw=green!70!black, fill=green!20, very thick, minimum size=8mm},
    legendbox/.style={rounded corners=1.5pt, draw=black!70, very thick, inner sep=4pt, fill=white},
    legitem/.style={font=\scriptsize, text=black!90},
    node distance=6mm and 6mm
  ]

\node[legendbox, anchor=north east] (leg) at (21.2,-0.7) {

\begin{tikzpicture}[baseline]
\node[event2022] (l22) at (0,0) {};
\node[legitem, right=4pt of l22] {2022 (Foundation)};
\node[event2023] (l23) at (0,-0.55) {};
\node[legitem, right=4pt of l23] {2023 (Scaling / Policies)};
\node[event2024] (l24) at (0,-1.10) {};
\node[legitem, right=4pt of l24] {2024 (Specialization)};
\node[event2025] (l25) at (0,-1.65) {};
\node[legitem, right=4pt of l25] {2025 (Generalization / Advanced)};
\end{tikzpicture}
};

\draw[timeline] (0,0) -- (21,0);

\node[event2022] (cliport) at (1,0) {};
\node[label, above=10pt of cliport] {CLIPort\\\cite{shridhar2022cliport}};

\node[event2022] (gato) at (3,0) {};
\node[label, above=10pt of gato] {Gato\\\cite{reed2022generalist}};

\node[event2022] (rt1) at (5,0) {};
\node[label, above=10pt of rt1] {RT-1\\\cite{brohan2022rt}};

\node[event2022] (vima) at (7,0) {};
\node[label, above=10pt of vima] {VIMA\\\cite{jiang2022vima}};

\node[event2023] (act) at (9,0) {};
\node[label, above=10pt of act] {ACT\\\cite{zhao2023learning}};

\node[event2023] (rt2) at (11,0) {};
\node[label, above=10pt of rt2] {RT-2\\\cite{zitkovich2023rt}};

\node[event2023] (voxposer) at (13,0) {};
\node[label, above=10pt of voxposer] {VoxPoser\\\cite{huang2023voxposer}};

\node[event2023] (diffpolicy) at (15,0) {};
\node[label, above=10pt of diffpolicy] {Diffusion Policy\\\cite{chi2023diffusion}};

\draw[continue] (20,0) -- (3,-4);

\draw[timeline] (3,-4) -- (21,-4);

\node[event2024] (octo) at (3,-4) {};
\node[label, below=10pt of octo] {Octo\\\cite{team2024octo}};

\node[event2024] (openvla) at (5,-4) {};
\node[label, below=10pt of openvla] {OpenVLA\\\cite{kim2024openvla}};

\node[event2024] (deer) at (7,-4) {};
\node[label, below=10pt of deer] {Deer-VLA\\\cite{yue2024deer}};

\node[event2024] (uni) at (9,-4) {};
\node[label, below=10pt of uni] {Uni-NaVid\\\cite{zhang2024uni}};

\node[event2024] (revla) at (11,-4) {};
\node[label, below=10pt of revla] {ReVLA\\\cite{dey2024revla}};

\node[event2024] (occllama) at (13,-4) {};
\node[label, below=10pt of occllama] {Occllama\\\cite{wei2024occllama}};

\node[event2024] (pi0) at (15,-4) {};
\node[label, below=10pt of pi0] {Pi-0\\\cite{black2024pi_0}};

\node[event2024] (rdt) at (17,-4) {};
\node[label, below=10pt of rdt] {RDT-1B\\\cite{liu2024rdt}};

\node[event2024] (cogact) at (19,-4) {};
\node[label, below=10pt of cogact] {CogAct\\\cite{li2024cogact}};

\draw[continue] (21,-4) -- (3,-8);

\draw[timeline] (3,-8) -- (21,-8);

\node[event2024] (edgevla) at (3,-8) {};
\node[label, below=10pt of edgevla] {EdgeVLA\\\cite{budzianowski20edgevla}};

\node[event2024] (showui) at (5,-8) {};
\node[label, below=10pt of showui] {ShowUI\\\cite{lin2024showui}};

\node[event2024] (navila) at (7,-8) {};
\node[label, below=10pt of navila] {NaviLa\\\cite{cheng2024navila}};

\node[event2024] (quar) at (9,-8) {};
\node[label, below=10pt of quar] {Quar-VLA\\\cite{ding2024quar}};

\node[event2024] (bivla) at (11,-8) {};
\node[label, below=10pt of bivla] {Bi-VLA\\\cite{gbagbe2024bi}};

\node[event2024] (robomamba) at (13,-8) {};
\node[label, below=10pt of robomamba] {RoboMamba\\\cite{liu2024robomamba}};

\node[event2025] (otter) at (15,-8) {};
\node[label, below=10pt of otter] {Otter\\\cite{huang2025otter}};

\node[event2025] (pointvla) at (17,-8) {};
\node[label, below=10pt of pointvla] {PointVLA\\\cite{li2025pointvla}};

\node[event2025] (hybridvla) at (19,-8) {};
\node[label, below=10pt of hybridvla] {HybridVLA\\\cite{liu2025hybridvla}};

\draw[continue] (21,-8) -- (3,-12);

\draw[timeline] (3,-12) -- (21,-12);

\node[event2025] (covla) at (3,-12) {};
\node[label, below=10pt of covla] {CoVLA\\\cite{arai2025covla}};

\node[event2025] (opendrivevla) at (5,-12) {};
\node[label, below=10pt of opendrivevla] {OpenDriveVLA\\\cite{zhou2025opendrivevla}};

\node[event2025] (orion) at (7,-12) {};
\node[label, below=10pt of orion] {ORION\\\cite{fu2025orion}};

\node[event2025] (objectvla) at (9,-12) {};
\node[label, below=10pt of objectvla] {ObjectVLA\\\cite{zhu2025objectvla}};

\node[event2025] (conrft) at (11,-12) {};
\node[label, below=10pt of conrft] {ConRFT\\\cite{chen2025conrft}};

\node[event2025] (hirobot) at (13,-12) {};
\node[label, below=10pt of hirobot] {Hi Robot\\\cite{shi2025hi}};

\node[event2025] (tla) at (15,-12) {};
\node[label, below=10pt of tla] {TLA\\\cite{hao2025tla}};

\node[event2025] (racevla) at (17,-12) {};
\node[label, below=10pt of racevla] {RaceVLA\\\cite{serpiva2025racevla}};

\node[event2025] (dexvla) at (19,-12) {};
\node[label, below=10pt of dexvla] {DexVLA\\\cite{wen2025dexvla}};

\draw[continue] (21,-12) -- (3,-16);

\draw[timeline] (3,-16) -- (21,-16);

\node[event2025] (humanoid) at (3,-16) {};
\node[label, below=10pt of humanoid] {Humanoid-VLA\\\cite{ding2025humanoid}};

\node[event2025] (safevla) at (5,-16) {};
\node[label, below=10pt of safevla] {SafeVLA\\\cite{zhang2025safevla}};

\node[event2025] (momanipvla) at (7,-16) {};
\node[label, below=10pt of momanipvla] {MoManipVLA\\\cite{wu2025momanipvla}};

\node[event2025] (vlacache) at (9,-16) {};
\node[label, below=10pt of vlacache] {VLA-Cache\\\cite{xu2025vla}};

\node[event2025] (tinyvla) at (11,-16) {};
\node[label, below=10pt of tinyvla] {TinyVLA\\\cite{wen2025tinyvla}};

\node[event2025] (gr00t) at (13,-16) {};
\node[label, below=10pt of gr00t] {Gr00t N1\\\cite{bjorck2025gr00t}};

\node[event2025] (nora) at (15,-16) {};
\node[label, below=10pt of nora] {NORA\\\cite{hung2025nora}};

\node[event2025] (spatialvla) at (17,-16) {};
\node[label, below=10pt of spatialvla] {SpatialVLA\\\cite{qu2025spatialvla}};

\node[event2025] (molevla) at (19,-16) {};
\node[label, below=10pt of molevla] {MoLe-VLA\\\cite{zhang2025mole}};

\draw[continue] (21,-16) -- (3,-20);

\draw[timeline] (3,-20) -- (21,-20);

\node[event2025] (longvla) at (3,-20) {};
\node[label, below=10pt of longvla] {Long-VLA\\\cite{fan2025long}};

\node[event2025] (retovla) at (5,-20) {};
\node[label, below=10pt of retovla] {RetoVLA\\\cite{koo2025retovla}};

\node[event2025] (vlaser) at (7,-20) {};
\node[label, below=10pt of vlaser] {Vlaser\\\cite{yang2025vlaser}};

\node[event2025] (ddvla) at (9,-20) {};
\node[label, below=10pt of ddvla] {Discrete\\Diffusion VLA\\\cite{liang2025discrete}};

\node[event2025] (beingh0) at (11,-20) {};
\node[label, below=10pt of beingh0] {Being-H0\\\cite{luo2025being}};

\node[event2025] (egovla) at (13,-20) {};
\node[label, below=10pt of egovla] {EgoVLA\\\cite{yang2025egovla}};

\node[event2025] (stereovla) at (15,-20) {};
\node[label, below=10pt of stereovla] {StereoVLA\\\cite{deng2025stereovla}};

\node[event2025] (geovla) at (17,-20) {};
\node[label, below=10pt of geovla] {GeoVLA\\\cite{sun2025geovla}};

\node[event2025] (efficientvla) at (19,-20) {};
\node[label, below=10pt of efficientvla] {EfficientVLA\\\cite{yang2025efficientvla}};

\end{tikzpicture}
}
\caption{Comprehensive timeline of Vision--Language--Action (VLA) models (2022--2025), organized by year with high-contrast color coding and thematic grouping.}
\label{fig:vla-timeline}
\end{figure*}

\subsection{Multimodal Integration: From Isolated Pipelines to Unified Agents} 
A central advancement in the emergence of VLA models lies in their ability to perform multi-modal integration, the joint processing of vision, language, and action within a unified architecture. Traditional robotic systems treated perception, natural language understanding, and control as discrete modules, often linked through manually defined interfaces or data transformations \cite{lin2019automatic, cangelosi2010integration, tellex2020robots}. For instance, classic pipeline-based frameworks required a perception model to output symbolic labels, which were then mapped by a planner to specific actions frequently with domain-specific hand engineering \cite{rawal2025intelligent, katiyar2023model}. These approaches lacked adaptability, failed in ambiguous or unseen environments, and could not generalize instructions beyond pre-defined templates.

In contrast, modern VLAs fuse modalities end-to-end using large-scale pretrained encoders and transformer-based architectures \cite{wu2024visionllm}. This shift enables the model to interpret visual observations and linguistic instructions within the same computational space, allowing flexible, context-aware reasoning \cite{li2023intentqa}. For example, in the task ``Pick the ripe apples'' (Figure~\ref{fig:concepts}), the vision encoder—typically a Vision Transformer (ViT) or ConvNeXt—parses the scene to localize and categorize relevant objects (e.g., fruit, foliage, background) and infer ripeness-related visual cues based on learned texture, shape, and contextual features rather than fixed color assumptions \cite{woo2023convnext}. Meanwhile, the language model, often a variant of T5, GPT, or BERT, encodes the instruction into a high-dimensional embedding. These representations are then fused via cross-attention or joint tokenization schemes, producing a unified latent space that informs the action policy \cite{han2024review}.

This multimodal synergy was first effectively demonstrated in CLIPort~\cite{shridhar2022cliport}, which takes an RGB image of a tabletop scene and a natural language instruction (e.g., ``place the blue block on the red square'') as inputs, encodes them using CLIP for semantic grounding, and outputs pixel-level pick-and-place action distributions via a convolutional transport decoder. By directly conditioning visuomotor policies on language embeddings, CLIPort eliminates explicit language parsing and enables end-to-end language-conditioned manipulation. Similarly, VIMA \cite{jiang2022vima} advanced this approach by employing a transformer encoder to jointly process object-centric visual tokens and instruction tokens, enabling few-shot generalization across spatial reasoning tasks.

Recent developments push this fusion further by incorporating temporal and spatial grounding. VoxPoser~\cite{huang2023voxposer} employs voxel-level reasoning to resolve ambiguities in 3D object selection by composing pretrained vision-language models and classical motion planners, notably achieving zero-shot manipulation without task-specific training data. In contrast, RT-2~\cite{zitkovich2023rt} fuses visual-language tokens and action representations within a unified transformer, co-trained on large-scale internet vision-language corpora and over 100{,}000 real-robot demonstrations from the RT-1 dataset, enabling zero-shot generalization to unseen instructions. Another noteworthy contribution is Octo~\cite{team2024octo}, which introduces a memory-augmented transformer trained on more than four million robot trajectories collected across diverse robots and environments via the Open X-Embodiment dataset, supporting long-horizon decision-making and demonstrating the scalability of joint perception--language--action learning.

Crucially, VLAs offer robust solutions to challenges faced in real-world grounding. For example, Occllama \cite{wei2024occllama} handles occluded object references through attention-based mechanisms, while ShowUI \cite{lin2024showui} demonstrates natural language interfaces that allow non-expert users to command agents through voice or typed input. These capabilities are only possible because the integration is not limited to surface-level fusion; rather, it captures semantic, spatial, and temporal alignment across modalities.

\subsection{Tokenization and Representation: How VLAs Encode the World} 
A core innovation that sets VLA models apart from conventional vision-language architectures lies in their token-based representation framework, which enables holistic reasoning over perceptual \cite{ni2024peria, zhao2025cot}, linguistic, and physical action spaces \cite{li2025visual}. Inspired by autoregressive generative models like transformers, modern VLAs encode the world using discrete tokens that unify all modalities vision, language, state, and action into a shared embedding space \cite{liu2025hybridvla}. This allows the model to not only understand “what needs to be done” (semantic reasoning), but also “how to do it” (control policy execution) in a fully learnable and compositional way \cite{xiong2024autoregressive, lu2024unified, tian2024visual}.

\begin{itemize}
    \item \textbf{Prefix Tokens: Encoding Context and Instruction:} Prefix tokens serve as the contextual backbone of VLA models \cite{xu2025vla, jeong2024survey}. These tokens encode the environmental scene (via images or video) and the accompanying natural language instruction into compact embeddings that prime the model's internal representations \cite{bordes2024introduction}. 
    
    For instance, as depicted in Figure \ref{fig:prefix} in a task such as “stack the green blocks on the red tray,” the image of a cluttered tabletop is processed through a vision encoder like ViT or ConvNeXt, while the instruction is embedded by an LLM (e.g., T5 or LLaMA). These are then transformed into a sequence of prefix tokens that establish the model’s initial understanding of the goal and environmental layout. This shared representation enables cross-modal grounding, allowing the system to resolve spatial references (e.g., “on the left,” “next to the blue cup”) and object semantics (“green blocks”) across both modalities.

\begin{figure}[ht!]
\centering
\includegraphics[width=0.99\linewidth]{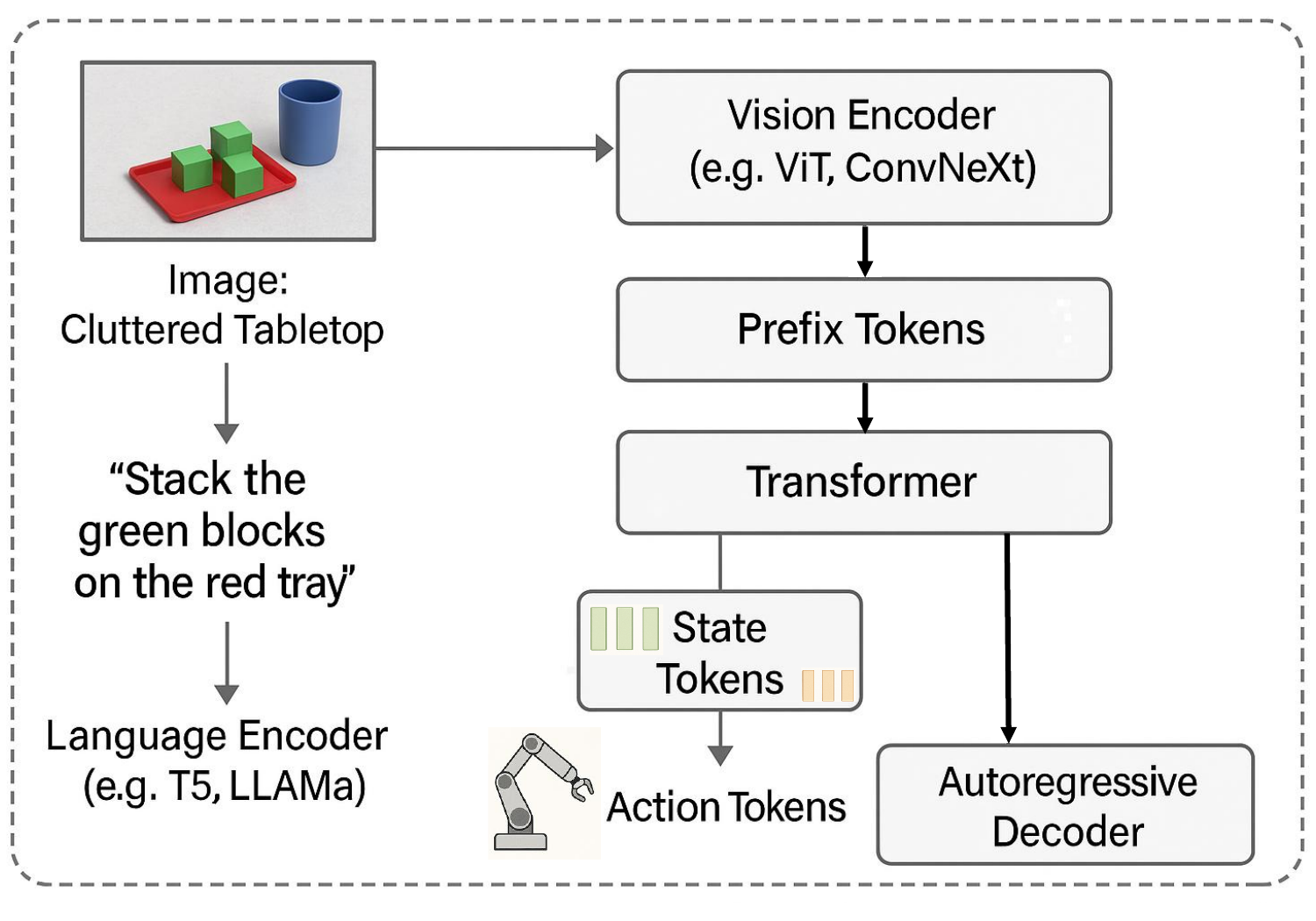}
\caption{A diagram illustrating the end-to-end tokenization and representation process in VLA models. Visual input (e.g., cluttered tabletop) is encoded by a vision encoder (e.g., ViT), while natural language instructions (e.g., “stack the green blocks”) are processed by a language encoder (e.g., T5). The system fuses prefix, state, and action tokens through a transformer and autoregressively predicts motor actions.}
\label{fig:prefix}
\end{figure}

\begin{figure}[ht!]
\centering
\includegraphics[width=0.99\linewidth]{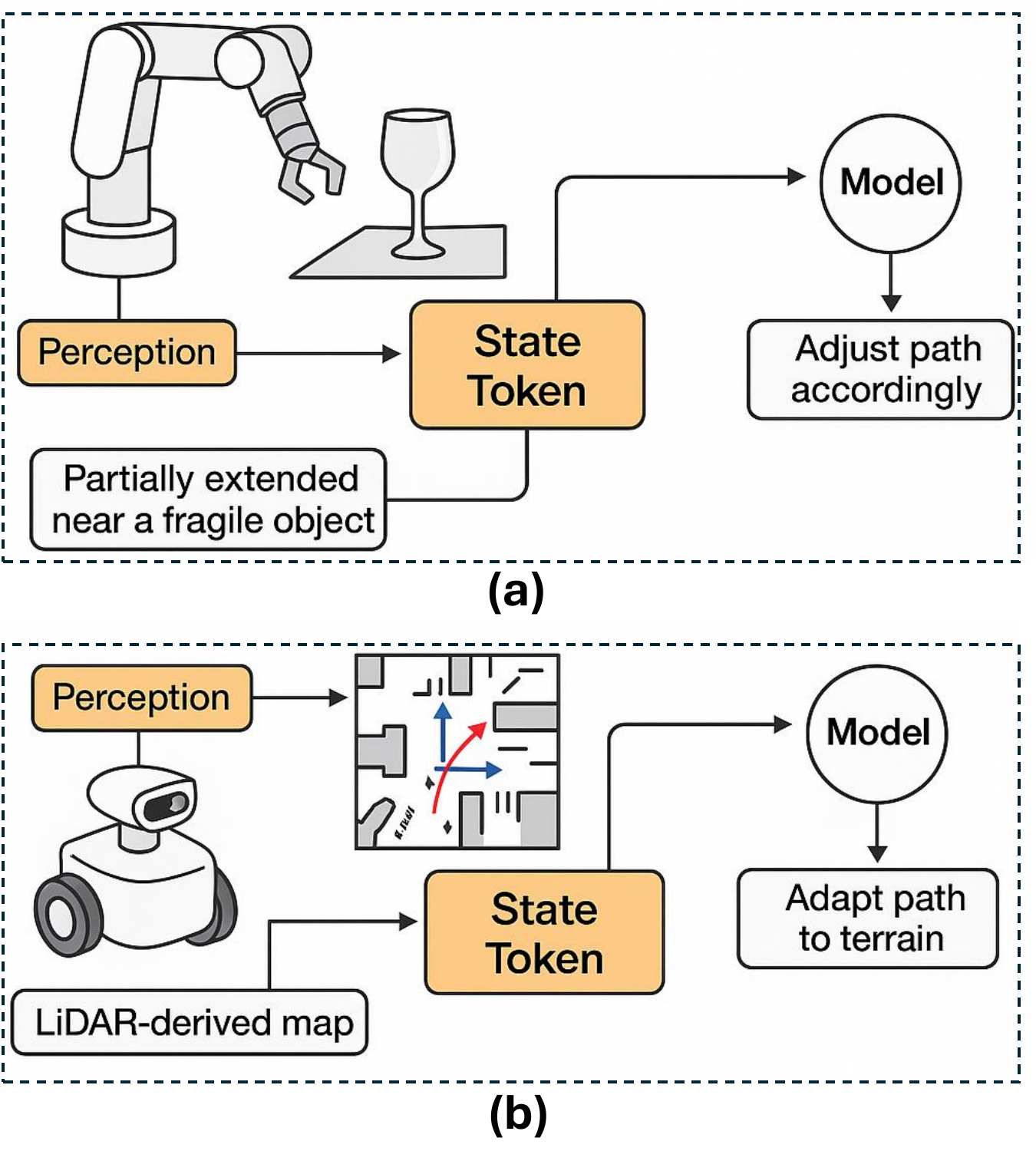}
\caption{Illustrating how VLA models utilize prefix, state, and action tokens in real-world scenarios. In robotic manipulation, state tokens detect arm extension near fragile objects, enabling path adjustment. In navigation, they represent LiDAR and odometry data. The apple-picking task shows how prefix tokens guide goal understanding, while action tokens generate motion sequences for targeted grasping and execution.}
\label{fig:prefixutilize}
\end{figure}
  \item \textbf{State Tokens: Embedding the Robot's Configuration:} In addition to perceiving external stimuli, VLAs must be aware of their internal physical state \cite{wen2025tinyvla, liu2024robomamba}. This is achieved through the use of state tokens, which encode real-time information about the agent's configuration joint positions, force-torque readings, gripper status, end-effector pose, and even the locations of nearby objects \cite{li2024foundation}. These tokens are crucial for ensuring situational awareness and safety, especially during manipulation or locomotion \cite{sun2025review, imran2025foundation}. 

    Figure~\ref{fig:prefixutilize} illustrates how VLA models utilize state tokens to enable dynamic, context-aware decision-making in both manipulation and navigation settings. In Figure~\ref{fig:prefixutilize}a, a robot arm is shown partially extended near a fragile object. In such scenarios, state tokens play a critical role by encoding real-time proprioceptive information, such as joint angles, gripper pose, and end-effector proximity. These tokens are continuously fused with visual and language-based prefix tokens, allowing the transformer to reason about physical constraints. The model can thus infer that a collision is imminent and adjust the motor commands accordingly e.g., rerouting the arm trajectory or modulating force output. In mobile robotic platforms, as depicted in Figure~\ref{fig:prefixutilize}b, state tokens encapsulate spatial features such as odometry, LiDAR scans, and inertial sensor data. These are essential for terrain-aware locomotion and obstacle avoidance. The transformer model integrates this state representation with environmental and instructional context to generate navigation actions that dynamically adapt to changing surroundings. Whether grasping objects in cluttered environments or autonomously navigating uneven terrain, state tokens provide a structured mechanism for situational awareness, enabling the autoregressive decoder to produce precise, context-informed action sequences that reflect both internal robot configuration and external sensory data.
    
    \item \textbf{Action Tokens: Autoregressive Control Generation:} 
    The final layer of the VLA token pipeline involves action tokens \cite{kim2025fine, kim2024openvla}, which are autoregressively generated by the model to represent the next step in motor control \cite{wen2025tinyvla}. Each token corresponds to a low-level control signal, such as joint angle updates, torque values, wheel velocities, or high-level movement primitives \cite{gu2025humanoid}. During inference, the model decodes these tokens one step at a time, conditioned on prefix and state tokens, effectively turning VLA models into language-driven policy generators \cite{firoozi2023foundation, song2025accelerating}. This formulation allows seamless integration with real-world actuation systems, supports variable-length action sequences \cite{bartoccioni2025vavim, huang2025decision}, and enables model fine-tuning via reinforcement or imitation learning frameworks \cite{zhao2025more}. Notably, models like RT-2~\cite{zitkovich2023rt} and PaLM-E~\cite{driess2023palm} exemplify this design, where perception, instruction, and embodiment are merged into a unified token stream.

For instance, in the apple-picking task as depicted in Figure \ref{fig:workflow}, the model may receive prefix tokens that include the image of the orchard and the text instruction. The state tokens describe the robot's current arm posture and whether the gripper is open or closed. Action tokens are then predicted step by step to guide the robotic arm toward the apple, adjust the gripper orientation, and execute a grasp with appropriate force. The beauty of this approach is that it allows transformers, which are traditionally used for text generation, to now generate sequences of physical actions in a manner similar to generating a sentence only here, the sentence is the motion.
\end{itemize}

\begin{figure}[ht!]
\centering
\includegraphics[width=0.99\linewidth]{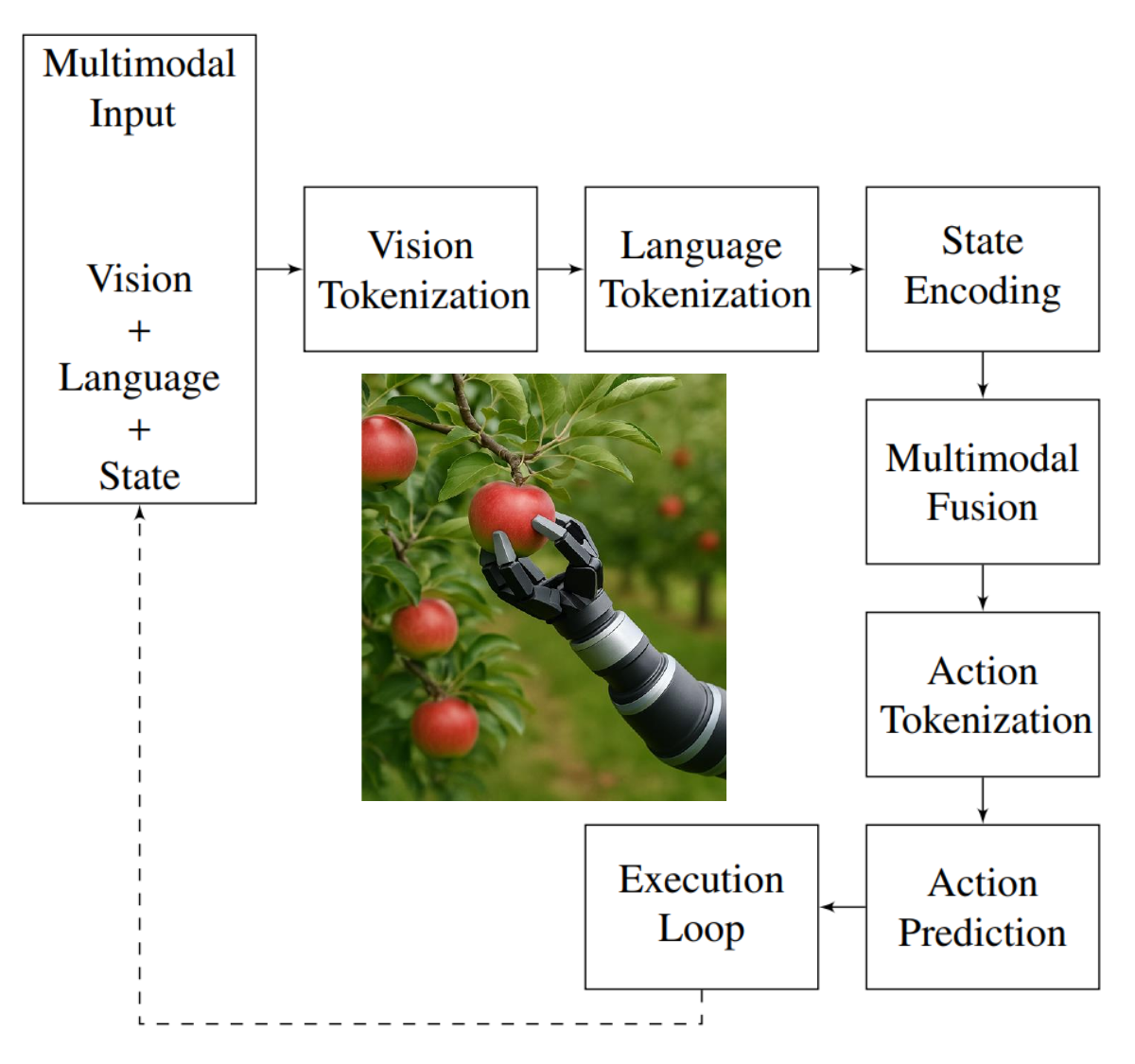}
\caption{\textbf{Illustrating the process of how VLAs Encode the World.} VLAs encode the world by converting vision, language, and sensor inputs into tokens, fusing them through cross-attention, predicting action sequences via transformers, and executing tasks with real-time feedback - enabling robots to interpret scenes, follow instructions, and adapt actions dynamically.}
\label{fig:workflow}
\end{figure}

To operationalize the VLA paradigm in robotics, we present in Figure~\ref{fig:workflow} a structured pipeline that demonstrates how multimodal information specifically vision, language, and proprioceptive state is encoded, fused, and converted into executable action sequences. This end-to-end loop allows a robot to interpret complex tasks like “pick the ripe apple near the green leaf” and execute precise, context-sensitive manipulations. The system begins with \textbf{multimodal input acquisition}, where three distinct data streams are collected: visual observations (e.g., RGB-D frames), natural language commands, and real-time robot state information (e.g., joint angles or velocity). These are independently tokenized into discrete embeddings using pretrained modules \cite{ding2024understanding, zhang2025generative}. As depicted in the diagram, the image is processed through a Vision Transformer (ViT) backbone to generate \textit{vision tokens}, the instruction is parsed by a language model such as BERT or T5 to produce \textit{language tokens}, and state inputs are transformed via a lightweight MLP encoder into compact \textit{state tokens}.

These tokens are then fused using a cross-modal attention mechanism, where the model jointly reasons over object semantics, spatial layout, and physical constraints \cite{ghosh2024exploring}. This fused representation forms the contextual basis for decision-making \cite{huangearly, lu2025probing}. In Figure~\ref{fig:workflow}, this is denoted as the \textbf{multi-modal fusion} step. The fused embedding is passed into an autoregressive decoder typically a transformer that generates a series of action tokens. These tokens may correspond to joint displacements, gripper force modulation, or high-level motor primitives (e.g., “move to grasp pose”, “rotate wrist”). The predicted action tokens are subsequently translated into low-level control commands and executed by an external, hardware-dependent execution loop, which interfaces with the robot controller to close the perception-action cycle by feeding back updated state observations for the next VLA inference step. This closed-loop mechanism enables the model to dynamically adapt to perturbations, object shifts, or occlusions in real time \cite{zhang2025gevrm, lyu2025dywa, xu2025embodied}.

To provide clear and specific implementation details, Algorithm~1 formalizes the VLA tokenization process. Given an RGB-D frame $I$, natural language instruction $T$, and joint angle vector $\theta$, the algorithm produces a set of action tokens that can be executed in sequence. The image $I$ is processed via a ViT to produce $V$, a set of 400 visual tokens. In parallel, the instruction $T$ is encoded by a BERT model to yield $L$, a sequence of 12 semantic language tokens. Simultaneously, the robot state $\theta$—including joint angles, end-effector pose, and proprioceptive signals—is encoded by a multilayer perceptron into a compact 64-dimensional state embedding $S$, providing the model with real-time awareness of the robot’s configuration and physical constraints during action generation. These tokens are then fused via a cross-attention module to produce a shared 512-dimensional representation $F$, capturing the semantics, intent, and situational awareness needed for grounded action. Finally, a policy decoder such as FAST~\cite{pertsch2025fast} maps the fused features into 50 discrete action tokens, which can then be decoded into motor commands $\tau_{1:N}$.

The decoding process is implemented using a transformer-based architecture, as shown in the code snippet titled \textit{Action Prediction Code}. A \texttt{Transformer} decoder is instantiated with 12 layers, a model dimension of 512, and 8 attention heads. The fused multimodal tokens are provided as context, and the decoder autoregressively predicts action tokens one step at a time, where each predicted token represents the next control decision conditioned on the full multimodal context and all previously generated actions. The resulting action-token sequence is then detokenized into a continuous motor command trajectory for execution. This implementation mirrors how text generation works in LLMs, but here the “sentence” is a motion trajectory a novel repurposing of natural language generation techniques for physical action synthesis.

Together, Figure~\ref{fig:workflow}, Algorithm~\ref{alg:vla_tokenization}, and the pseudocode illustrate how VLAs unify perception, instruction, and embodiment within a coherent and interpretable token space. This modularity allows the framework to generalize across tasks and robot morphologies, facilitating rapid deployment in real-world applications like apple picking, household tasks, and mobile navigation. Importantly, the clarity and separability of the tokenization steps make the architecture extensible, enabling further research on token learning, hierarchical planning, or symbolic grounding in VLA systems.
\vspace{0.5em}
\begin{algorithm}[H]
\caption{VLA Tokenization Pipeline}
\label{alg:vla_tokenization}
\begin{algorithmic}[1]
\State \textbf{Input}: RGB-D frame $I$, text command $T$, joint angles $\theta$
\State $V \gets \text{ViT}(I)$ \Comment{400 vision tokens}
\State $L \gets \text{BERT}(T)$ \Comment{12 language tokens}
\State $S \gets \text{MLP}(\theta)$ \Comment{64-dim state encoding}
\State $F \gets \text{CrossAttention}(V,L,S)$ \Comment{512-dim fused token}
\State $A \gets \text{FAST}(F)$ \Comment{50 action tokens}
\State \textbf{Output}: Motor commands $\tau_{1:N}$
\end{algorithmic}
\end{algorithm}
\vspace{0.5em}

\textbf{Action Prediction Code}
\begin{verbatim}
# Python-like pseudocode
def predict_actions(fused_tokens):
    transformer = Transformer(
        num_layers=12,
        d_model=512,
        nhead=8
    )
    action_tokens = transformer.decode(
        fused_tokens,
        memory=fused_tokens
    )
    return detokenize(action_tokens)
\end{verbatim}

\subsection{Learning Paradigms: Data Sources and Training Strategies}

\begin{figure*}[ht!]
\centering
\includegraphics[width=0.68\linewidth]{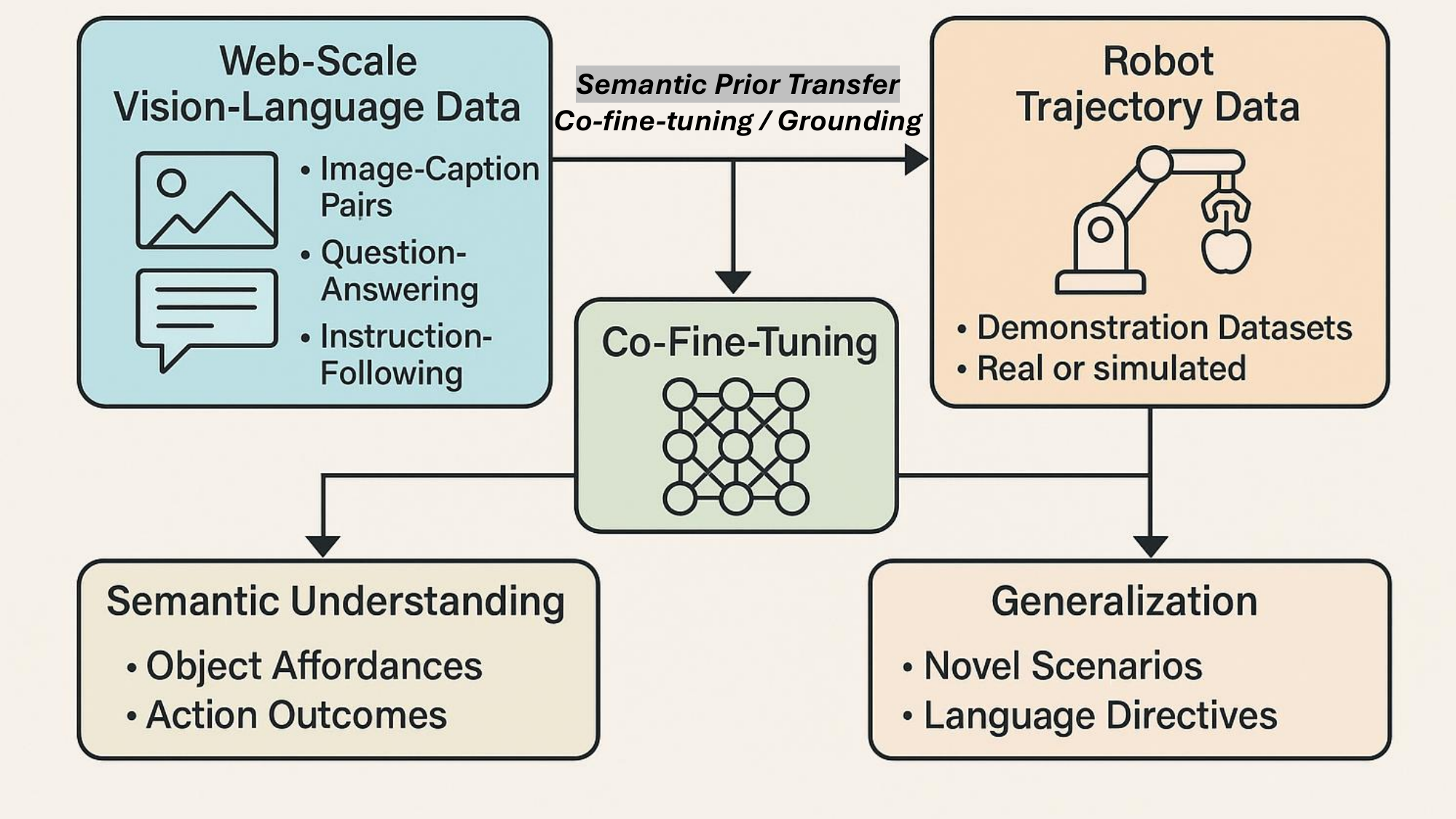}
\caption{\textbf{Learning Paradigms: Data Sources and Training Strategies for VLAs.}  }
\label{fig:learningparadm}
\end{figure*}
Training VLA models requires a hybrid learning paradigm that integrates both semantic knowledge from the web and task-grounded information from robotics datasets \cite{chen2025vision}. As shown in prior sections, the multi-modal architecture of VLAs must be exposed to diverse forms of data that support language understanding, visual recognition, and motor control. This is typically achieved through two primary data sources.

First, as depicted in figure \ref{fig:learningparadm}, large-scale internet-derived corpora form the backbone of the model’s semantic prior. These datasets include image-caption pairs (e.g., COCO, LAION-400M), instruction-following datasets (e.g., HowTo100M, WebVid), and visual question-answering corpora (e.g., VQA, GQA). Such datasets enable pretraining of the visual and language encoders, helping the model acquire general representations of objects, actions, and concepts \cite{agarwal2024methods}. This phase often uses contrastive or masked modeling objectives, such as CLIP-style contrastive learning or language modeling losses, to align vision and language modalities within a shared embedding space \cite{sameni2024building, yang2023attentive}. Importantly, this stage gives VLAs a foundational “understanding of the world” that facilitates compositional generalization, object grounding, and zero-shot transfer \cite{chen2025language, bolya2025perception}.

However, semantic understanding alone is insufficient for physical task execution \cite{dang2025ecbench, wang2025learn, li2025benchmark}. Thus, the second phase focuses on grounding the model in embodied experience \cite{wang2025learn}. Robot trajectory datasets collected either from real-world robots or high-fidelity simulators are used to teach the model how language and perception translate into action \cite{firoozi2023foundation}. These include datasets like RoboNet \cite{dasari2019robonet}, BridgeData \cite{ebert2021bridge}, and RT-X \cite{vuong2023open}, which provide video-action pairs, joint trajectories, and environment interactions under natural language instructions \cite{moroncelli2024integrating}. Demonstration data may come from kinesthetic teaching, teleoperation, or scripted policies \cite{karamcheti2021learning, belkhale2024rt}. This phase typically employs supervised learning (e.g., behavior cloning) \cite{foster2024behavior}, reinforcement learning (RL), or imitation learning to train the autoregressive policy decoder to predict action tokens based on fused visual-language-state embeddings \cite{guo2025improving}.

Recent works increasingly adopt \textit{multistage or multitask training strategies}. For example, models are often pretrained on vision-language datasets using masked language modeling, then fine-tuned on robot demonstration data using token-level autoregressive loss \cite{kim2024openvla, zhou2025chatvla, xu2025vla}. Others use curriculum learning, where simpler tasks (e.g., object pushing) precede more complex ones (e.g., multistep manipulation) \cite{zhen20243d}. Some approaches further leverage domain adaptation such as in OpenVLA \cite{kim2024openvla} or sim-to-real transfer to bridge the gap between synthetic and real-world distributions \cite{li2025pointvla}. By unifying semantic priors with task execution data, these learning paradigms allow VLA models to generalize across tasks, domains, and embodiments forming the backbone of scalable, instruction-following agents capable of robust real-world operation.

Through co-fine-tuning, these datasets are brought into alignment \cite{wang2025roboflamingo, fan2024language}. The model learns to map from visual and linguistic inputs to appropriate action sequences \cite{qu2025spatialvla}. This training paradigm not only helps the model understand object affordances (e.g., apples can be grasped) and action outcomes (e.g., lifting requires force and trajectory), but also promotes generalization to novel scenarios \cite{li2024improving}. A model trained on kitchen manipulation tasks may be able to infer how to pick an apple in an outdoor orchard if it has learned general principles of object localization, grasping, and following language directives.

Recent architectures, such as Google DeepMind’s RT-2 (Robotic Transformer 2) \cite{zitkovich2023rt}, have demonstrated this principle in action. RT-2 treats action generation as a form of text generation, where each action token corresponds to a discrete command in a robot’s control space. Because the model is trained on both web-scale multi-modal data and thousands of robot demonstrations, it can flexibly interpret novel instructions and perform zero-shot generalization to new objects and tasks something that was largely impossible with traditional control systems or even with early multi-modal models.

\subsection{Adaptive Control and Real-Time Execution}
Another strength of VLAs lies in their ability to perform adaptive control, using real-time feedback from sensors to adjust behavior on the fly \cite{serpiva2025racevla}. This is particularly important in dynamic, unstructured environments like orchards, homes, or hospitals, where unexpected changes (e.g., wind moving an apple, lighting changes, human presence) can alter the task parameters. During execution, state tokens are updated in real-time, reflecting sensor inputs and joint feedback \cite{xu2025vla}. The model can then revise its planned actions accordingly. For instance, in the apple-picking scenario, if the target apple shifts slightly or another apple enters the field of view, the model dynamically reinterprets the scene and adjusts the grasp trajectory. This capability mimics human-like adaptability and is a core advantage of VLA systems over pipeline-based robotics.

\section{Progress in Vision-Language-Action Models}
The inception of VLA models was catalyzed by the remarkable success of transformer-based LLMs, notably ChatGPT, released in November 2022, which demonstrated unprecedented semantic reasoning capabilities (ChatGPT) \cite{ray2023chatgpt}. This breakthrough inspired researchers to extend language models to multimodal domains, integrating perception and action for robotics. By 2023, GPT-4 introduced multimodal capabilities by processing both text and images, which spurred subsequent efforts to extend language-centric multimodal foundation models toward incorporating physical action representations and control interfaces \cite{achiam2023gpt}. Concurrently, VLMs like CLIP (2022) \cite{shridhar2022cliport} and Flamingo (2022) \cite{alayrac2022flamingo} had established robust visual-text alignment through contrastive learning, enabling zero-shot object recognition and laying the groundwork for VLM models (such as CLIP). These models leveraged large-scale labeled datasets to align images with textual descriptions, a critical precursor to integrating actions. 

A pivotal development was the creation of large-scale robotic datasets, such as RT-1’s 130,000 demonstrations, which provided action-grounding data essential for co-training vision, language, and action components \cite{brohan2022rt}. These datasets captured diverse tasks and environments, enabling models to learn generalizable behaviors. Architectural breakthroughs followed with Google’s RT-2 in 2023 \cite{brohan2023rt}, a landmark VLA model that unified vision, language, and action tokens, treating robotic control as an autoregressive sequence prediction task (RT-2 Blog(\href{https://deepmind.google/blog/rt-2-new-model-translates-vision-and-language-into-action/}{Source Link})). RT-2 discretized actions using Discrete Cosine Transform (DCT) compression and Byte-Pair Encoding (BPE), achieving a 63\% improvement in performance on novel objects. Multimodal fusion techniques, such as cross-attention transformers, integrated Vision Transformer (ViT)-processed images (e.g., 400 patch tokens) with language embeddings, enabling robots to execute complex commands like “Pick the red cup left of the bowl.” Additionally, UC Berkeley’s Octo model (2023) introduced an open-source approach with 93M parameters and diffusion decoders, trained on 800,000 robot demonstrations from the OpenX-Embodiment dataset, further broadening the research landscape \cite{team2024octo}.

\subsection{Architectural Innovations in VLA Models}
From 2023 to 2024, VLA models underwent significant architectural advancements and refined training methodologies. Dual-system architectures emerged as a key innovation, exemplified by NVIDIA’s GR00T N1 (2025) \cite{bjorck2025gr00t}, which combined System 1 (fast diffusion policies with 10ms latency for low-level control) and System 2 (LLM-based planners for high-level task decomposition). This separation enabled efficient coordination between strategic planning and real-time execution, enhancing adaptability in dynamic environments. Other models, like Stanford’s OpenVLA (2024) \cite{kim2024openvla}, introduced a 7B-parameter open-source VLA trained on 970k real-world robot demonstrations, using dual vision encoders (DINOv2 \cite{oquab2023dinov2} and SigLIP \cite{zhai2023sigmoid}) and a Llama 2 language model \cite{touvron2023llama}, outperforming larger models like RT-2-X (55B) \cite{kim2024openvla}. Training paradigms evolved to leverage co-fine-tuning on web-scale vision-language data (e.g., LAION-5B) \cite{schuhmann2022laion} and robotic trajectory data (e.g., RT-X) \cite{vuong2023open}, aligning semantic knowledge with physical constraints \cite{schuhmann2022laion}. Synthetic data generation tools like UniSim addressed data scarcity by creating photorealistic scenarios, such as occluded objects, crucial for robust training (UniSim \cite{yang2023unisim}). Parameter efficiency was enhanced through Low-Rank Adaptation (LoRA) adapters for fine-tuning \cite{hu2022lora}, which allowed domain adaptation without full retraining, reducing GPU hours by 70\%. The introduction of diffusion-based policies, as seen in Physical Intelligence’s pi 0 model (2024) \cite{black2024pi_0}, offered improved action diversity but required significant computational resources. These advancements democratized VLA technology, fostering collaboration and accelerating innovation. 

Recent VLA models have converged toward three major architectural paradigms that balance efficiency, modularity, and robustness: early fusion models, dual-system architectures, and self-correcting frameworks. Each of these innovations addresses specific challenges in grounding, generalization, and action reliability in real-world robotic systems.

\textbf{1. Early Fusion Models:}  
One class of VLA approaches focuses on fusing vision and language representations at the input stage before passing them to the policy module. Huang et al.’s EF-VLA model \cite{huangearly}, presented at International Conference on Learning Representations (ICLR 2025), exemplifies this trend by retaining the representational alignment established by CLIP \cite{shridhar2022cliport}. EF-VLA accepts image-text pairs, encodes them with CLIP’s frozen encoders, and fuses the resulting embeddings early in the transformer backbone prior to action prediction. This design ensures that the semantic consistency learned during CLIP pretraining is preserved, reducing overfitting and enhancing generalization. Notably, EF-VLA demonstrated a 20\% performance improvement on compositional manipulation tasks and reached 85\% success on previously unseen goal descriptions. By keeping the vision-language backbone frozen, this approach preserves computational efficiency and avoids catastrophic forgetting, while domain-specific training is confined to lightweight policy or action modules, enabling task adaptation without sacrificing the model’s general-purpose visual-semantic representations.

\textbf{2. Dual-System Architectures:}  
Inspired by dual-process theories of human cognition, models like NVIDIA’s GR00T N1 (2025) \cite{bjorck2025gr00t} implement two complementary subsystems: a fast-reactive module (System 1) and a slow-reasoning planner (System 2). System 1 comprises a diffusion-based control policy that operates at 10 ms latency, ideal for fine-grained, low-level control such as end-effector stabilization or adaptive grasping. In contrast, System 2 uses a LLM for task planning, skill composition, and high-level sequencing. The planner parses long-horizon goals (e.g., “clean the table”) into atomic subtasks, while the low-level controller ensures real-time execution. This decomposition enables multi-timescale reasoning and improved safety, especially in environments where rapid reaction and deliberation must co-exist. In benchmark tests on multi-stage household manipulation, GR00T N1 outperformed monolithic models such as RT-1, RT-2, and OpenVLA by 17\% in success rate and reduced collision failures by 28\%.

\textbf{3. Self-Correcting Frameworks:}  
A third architectural evolution is the emergence of self-correcting VLA models, which augment conventional inference pipelines with explicit failure detection and recovery mechanisms. SC-VLA (2024) retains a standard fast inference path similar to earlier end-to-end or hierarchical VLA designs, but introduces an additional, slower correction path that is selectively activated to re-evaluate decisions and generate recovery actions when execution failures or inconsistencies are detected. In this framework, the default behavior is to predict poses or actions directly from the fused embedding using a lightweight transformer. When failures are detected e.g., unsuccessful grasps or obstacle collisions, the model invokes a secondary process that performs chain-of-thought reasoning \cite{zhang2024learning, zawalski2024robotic}. This path queries an internal LLM (or external expert system) to diagnose failure modes and generate correction strategies \cite{duan2024aha}. For example, if the robot repeatedly misidentifies an occluded object, the LLM may suggest an active viewpoint change or gripper reorientation. In closed-loop experiments, SC-VLA reduced task failure rates by 35\% and significantly improved recoverability in cluttered and adversarial environments.

\textbf{4. Architectural Design Space of VLA Models}
VLA models exhibit a rich diversity of architectural designs and functional emphases, which can be systematically organized along the dimensions of end-to-end versus modular pipelines, hierarchical versus flat policy structures, and the balance between low-level control and high-level planning (Table \ref{tab:vla-classification-continued}). End-to-end VLAs, such as CLIPort~\cite{shridhar2022cliport}, RT-1~\cite{brohan2022rt}, and OpenVLA~\cite{kim2024openvla}, process raw sensory inputs directly into motor commands via a single unified network. By contrast, component-focused models like VLATest~\cite{wang2024towards} and Chain-of-Affordance~\cite{li2024improving} decouple perception, language grounding, and action modules, enabling targeted improvements in individual submodules.

Hierarchical architectures have emerged to tackle complex, long-horizon tasks by separating strategic decision making from reactive control. For instance, CogACT~\cite{li2024cogact} and NaVILA~\cite{cheng2024navila} employ a two-tier hierarchy where an LLM-based planner issues subgoals to a low-level controller, thereby combining the strengths of System 2 reasoning and System 1 execution. Similarly, ORION~\cite{fu2025orion} integrates a QT-Former for long-term context aggregation with a generative trajectory planner in a cohesive framework.

Low-level policy emphasis is typified by diffusion-based controllers (e.g.\ Pi-0~\cite{black2024pi_0}, DexGraspVLA~\cite{zhong2025dexgraspvla}), which excel at producing smooth, diverse motion distributions but often incur higher computational cost. In contrast, high-level planners (e.g.\ FAST Pi-0 Fast~\cite{pertsch2025fast}, CoVLA~\cite{arai2025covla}) focus on rapid subgoal generation or coarse trajectory prediction, delegating fine-grained control to specialized modules or traditional motion planners. End-to-end dual-system models like HybridVLA~\cite{liu2025hybridvla} and Helix~\cite{team2025gemini} blur these distinctions by jointly training both components while preserving modular interpretability.

Table \ref{tab:vla-classification-continued} further highlights how recent VLAs balance these trade-offs. Models such as OpenDriveVLA~\cite{zhou2025opendrivevla} and CombatVLA~\cite{chen2025combatvla} prioritize hierarchical planning in dynamic, safety-critical domains, whereas lightweight, edge-targeted systems like Edge VLA~\cite{budzianowski20edgevla} and TinyVLA~\cite{wen2025tinyvla} emphasize real-time low-level policies at the expense of high-level reasoning. This classification framework not only clarifies the design space of VLAs but also guides future development by pinpointing under-explored combinations such as fully end-to-end, hierarchical models optimized for embedded deployment that promise to advance both the capabilities and the applicability of VLA systems across robotics, autonomous driving, and beyond.

The classification in Table \ref{tab:vla-classification-continued} is significant because it provides a clear framework for comparing diverse VLA architectures, highlighting how design choices such as end-to-end integration versus hierarchical decomposition impact task performance, scalability, and adaptability. By categorizing models along dimensions such as low-level policy execution and high-level planning, researchers can more clearly identify the strengths and limitations of existing approaches and uncover opportunities for architectural innovation. For example, agricultural robotics tasks such as high-speed fruit harvesting or precision spraying benefit from architectures emphasizing fast, reactive low-level controllers, whereas applications like orchard navigation, multi-row coverage planning, or long-horizon crop monitoring require stronger high-level planning and reasoning capabilities. This taxonomy therefore aids in selecting appropriate VLA architectures for specific use cases and guides future development toward hybrid systems that balance responsiveness with cognitive planning, ultimately accelerating progress in embodied AI.

Additionally, to synthesize recent advancements in VLA models, Table~\ref{tab:vla_models_compact} presents a summary of notable systems developed from 2022 through 2025. Building upon architectural innovations such as early fusion, dual-system processing, and self-correcting feedback loops, these models incorporate diverse design philosophy and training strategies. Each table entry explicitly enumerates the model’s architectural components—namely the vision encoder, language encoder, and action decoder together with the primary training datasets used to ground and evaluate the model’s capabilities. Models like \textit{CLIPort} \cite{shridhar2022cliport} and \textit{RT-2} \cite{zitkovich2023rt} laid early foundations by aligning semantic embeddings with action policies, while more recent frameworks like \textit{Pi-Zero}, \textit{CogACT} \cite{li2024cogact}, and \textit{GR00T N1} \cite{bjorck2025gr00t} introduce scalable architectures with diffusion-based or high-frequency controllers. Several models leverage multi-modal pretraining with internet-scale vision-language corpora and robot trajectory datasets, enhancing generalization and zero-shot capabilities \cite{zhu2025objectvla, zhong2025dexgraspvla, zheng2025universal, yang2025fp3}. This tabulated comparison serves as a reference point for researchers seeking to understand the functional diversity, domain applicability, and emerging trends in VLA design across real and simulated environments.

\begin{table*}[h!]
\centering
\caption{Taxonomy of  VLA  models showing structured classification based on architectural paradigms and scientific priorities. We differentiate models by their support for end-to-end execution, hierarchical planning–control decomposition, or component-focused modularity, and further by their emphasis on low-level motor policies versus high-level task planners.} 
\label{tab:vla-classification-continued}

\begin{tabular}{
  p{5cm} 
  p{1.8cm}   
  p{1.5cm} 
  p{1.6cm} 
  p{1.6cm} 
  p{1.6cm} 
  p{1.7cm} 
}
\toprule
\textbf{Model Name} & \textbf{Year} & \textbf{End-to-End} & \textbf{Hie rarc hical} & \textbf{Comp onent Focused} & \textbf{Low-Level Policy} & \textbf{High-Level Planner} \\
\midrule
CLIPort~\cite{shridhar2022cliport}           & 2022 & \checkmark & \ding{55} & \ding{55} & \checkmark & \ding{55} \\
RT-1~\cite{brohan2022rt}                     & 2022 & \checkmark & \ding{55} & \ding{55} & \checkmark & \ding{55} \\
Gato~\cite{reed2022generalist}               & 2022 & \checkmark & \ding{55} & \ding{55} & \checkmark & \ding{55} \\
VIMA~\cite{jiang2022vima}                    & 2022 & \checkmark & \ding{55} & \ding{55} & \checkmark & \ding{55} \\
Diffusion Policy~\cite{chi2023diffusion}      & 2023 & \checkmark & \ding{55} & \ding{55} & \checkmark & \ding{55} \\
ACT~\cite{zhao2023learning}                  & 2023 & \checkmark & \ding{55} & \ding{55} & \checkmark & \ding{55} \\
VoxPoser~\cite{huang2023voxposer}            & 2023 & \checkmark & \ding{55} & \ding{55} & \checkmark & \ding{55} \\
Seer~\cite{gu2023seer}                       & 2023 & \checkmark & \ding{55} & \ding{55} & \checkmark & \ding{55} \\
Octo~\cite{team2024octo}                     & 2024 & \checkmark & \ding{55} & \ding{55} & \checkmark & \ding{55} \\
OpenVLA~\cite{kim2024openvla}                & 2024 & \checkmark & \ding{55} & \ding{55} & \checkmark & \ding{55} \\
CogACT~\cite{li2024cogact}                   & 2024 & \ding{55}   & \checkmark & \ding{55} & \checkmark & \checkmark \\
VLATest~\cite{wang2024towards}               & 2024 & \ding{55}   & \ding{55}   & \checkmark & \ding{55}   & \ding{55}   \\
NaVILA~\cite{cheng2024navila}                & 2024 & \ding{55}   & \checkmark & \ding{55} & \checkmark & \checkmark \\
RoboNurse-VLA~\cite{li2024robonurse}         & 2024 & \checkmark & \ding{55} & \ding{55} & \checkmark & \ding{55} \\
Mobility VLA~\cite{chiang2024mobility}       & 2024 & \ding{55}   & \checkmark & \ding{55} & \checkmark & \checkmark \\
RevLA~\cite{dey2024revla}                    & 2024 & \ding{55}   & \ding{55}   & \checkmark & \ding{55}   & \ding{55}   \\
Uni-NaVid~\cite{zhang2024uni}                & 2024 & \ding{55}   & \checkmark & \ding{55} & \checkmark & \checkmark \\
RDT-1B~\cite{liu2024rdt}                     & 2024 & \checkmark & \ding{55} & \ding{55} & \checkmark & \ding{55} \\
RoboMamba~\cite{liu2024robomamba}            & 2024 & \checkmark & \ding{55} & \ding{55} & \checkmark & \ding{55} \\
Chain-of-Affordance~\cite{li2024improving}    & 2024 & \ding{55}   & \ding{55}   & \checkmark & \ding{55}   & \ding{55}   \\
Edge VLA~\cite{budzianowski20edgevla}        & 2024 & \ding{55}   & \ding{55}   & \checkmark & \ding{55}   & \ding{55}   \\
ShowUI-2B~\cite{lin2024showui}               & 2024 & \checkmark & \ding{55} & \ding{55} & \checkmark & \ding{55} \\
Pi-0~\cite{black2024pi_0}                    & 2024 & \checkmark & \ding{55} & \ding{55} & \checkmark & \ding{55} \\
FAST (Pi-0 Fast)~\cite{pertsch2025fast}       & 2025 & \ding{55}   & \ding{55}   & \checkmark & \checkmark & \ding{55}   \\
OpenVLA-OFT~\cite{kim2025fine}               & 2025 & \checkmark & \ding{55} & \ding{55} & \checkmark & \ding{55} \\
CoVLA~\cite{arai2025covla}                   & 2025 & \ding{55}   & \checkmark & \ding{55} & \checkmark & \checkmark \\
OpenDriveVLA~\cite{zhou2025opendrivevla}     & 2025 & \ding{55}   & \checkmark & \ding{55} & \checkmark & \checkmark \\
ORION~\cite{fu2025orion}                     & 2025 & \ding{55}   & \checkmark & \ding{55} & \checkmark & \checkmark \\
UAV-VLA~\cite{sautenkov2025uav}              & 2025 & \ding{55}   & \checkmark & \ding{55} & \checkmark & \checkmark \\
CombatVLA~\cite{chen2025combatvla}           & 2025 & \checkmark & \ding{55} & \ding{55} & \checkmark & \ding{55} \\
HybridVLA~\cite{liu2025hybridvla}            & 2025 & \ding{55}   & \checkmark & \ding{55} & \checkmark & \checkmark \\
NORA~\cite{hung2025nora}                     & 2025 & \checkmark & \ding{55} & \ding{55} & \checkmark & \ding{55} \\
SpatialVLA~\cite{qu2025spatialvla}           & 2025 & \ding{55}   & \ding{55}   & \checkmark & \checkmark & \ding{55}   \\
MoLe-VLA~\cite{zhang2025mole}                & 2025 & \ding{55}   & \ding{55}   & \checkmark & \checkmark & \ding{55}   \\
JARVIS-VLA~\cite{li2025jarvis}               & 2025 & \checkmark & \ding{55} & \ding{55} & \checkmark & \ding{55} \\
UP-VLA~\cite{zhang2025up}                    & 2025 & \checkmark & \ding{55} & \ding{55} & \checkmark & \ding{55} \\
Shake-VLA~\cite{khan2025shake}               & 2025 & \ding{55}   & \ding{55}   & \checkmark & \checkmark & \ding{55}   \\
DexGraspVLA~\cite{zhong2025dexgraspvla}      & 2025 & \ding{55}   & \checkmark & \ding{55} & \checkmark & \checkmark \\
DexVLA~\cite{wen2025dexvla}                  & 2025 & \ding{55}   & \checkmark & \ding{55} & \checkmark & \checkmark \\
Humanoid-VLA~\cite{ding2025humanoid}         & 2025 & \checkmark & \ding{55} & \ding{55} & \checkmark & \ding{55} \\
ObjectVLA~\cite{zhu2025objectvla}            & 2025 & \checkmark & \ding{55} & \ding{55} & \checkmark & \ding{55} \\
Long-VLA~\cite{fan2025long}                 & 2025 & \checkmark & \ding{55}   & \ding{55}   & \checkmark & \ding{55} \\
RetoVLA~\cite{koo2025retovla}                  & 2025 & \ding{55}   & \ding{55}   & \checkmark & \checkmark & \ding{55} \\
Vlaser~\cite{yang2025vlaser}                   & 2025 & \ding{55}   & \checkmark & \ding{55}   & \checkmark & \checkmark \\
Discrete Diffusion VLA~\cite{liang2025discrete}   & 2025 & \checkmark & \ding{55}   & \ding{55}   & \checkmark & \ding{55} \\
Being-H0~\cite{luo2025being}                 & 2025 & \checkmark & \ding{55}   & \ding{55}   & \checkmark & \ding{55} \\
EgoVLA~\cite{yang2025egovla}                   & 2025 & \checkmark & \ding{55}   & \ding{55}   & \checkmark & \ding{55} \\
StereoVLA~\cite{deng2025stereovla}             & 2025 & \checkmark & \ding{55}   & \ding{55}   & \checkmark & \ding{55} \\
GeoVLA~\cite{sun2025geovla}                   & 2025 & \ding{55}   & \checkmark & \ding{55}   & \checkmark & \checkmark \\
EfficientVLA~\cite{yang2025efficientvla}     & 2025 & \ding{55}   & \ding{55}   & \checkmark & \checkmark & \ding{55} \\
\bottomrule
\end{tabular}
\end{table*}

\clearpage
\onecolumn

\begin{center}
\setlength{\tabcolsep}{4.0pt} 
\renewcommand{\arraystretch}{1.12}
\scriptsize
\begin{longtable}{p{3.2cm} p{4.6cm} p{3.6cm} p{5.8cm}}
\caption{\textbf{Compact summary of representative Vision--Language--Action (VLA) models.} Each row reports the primary encoders/decoders, training data, and the main distinctive capability.}
\label{tab:vla_models_compact}\\

\toprule
\textbf{Model (Ref.)} & \textbf{Architecture (vision / language / action)} & \textbf{Training data} & \textbf{Key strength / uniqueness} \\
\midrule
\endfirsthead

\multicolumn{4}{c}{\small\sl Continued from previous page}\\
\toprule
\textbf{Model (Ref.)} & \textbf{Architecture (vision / language / action)} & \textbf{Training data} & \textbf{Key strength / uniqueness} \\
\midrule
\endhead

\midrule
\multicolumn{4}{r}{\small\sl Continued on next page}\\
\bottomrule
\endfoot

\bottomrule
\endlastfoot

CLIPort~\cite{shridhar2022cliport} &
CLIP-ResNet50 + Transporter-ResNet / CLIP-GPT / LingUNet &
Self-collected [SC] &
Aligns semantic CLIP features with Transporter spatial reasoning for precise SE(2) manipulation.\\
\addlinespace

RT-1~\cite{brohan2022rt} &
EfficientNet / Universal Sentence Encoder / Transformer (discretized actions) &
RT-1-Kitchen [SC] &
Early large-scale transformer policy for multi-task kitchen manipulation with tokenized actions.\\
\addlinespace

RT-2~\cite{zitkovich2023rt} &
ViT-22B or ViT-4B / PaLI-X or PaLM-E / symbol-tuning (action tokens) &
VQA + RT-1-Kitchen &
Co-finetunes internet-scale VQA with robot data, yielding emergent generalization for embodied tasks.\\
\addlinespace

Gato~\cite{reed2022generalist} &
ViT / SentencePiece / Transformer (unified token stream) &
Self-collected [SC] &
Generalist agent unifying robotics, language, and Atari via shared tokenization and a single transformer.\\
\addlinespace

VIMA~\cite{jiang2022vima} &
ViT + Mask R-CNN / T5 / Transformer &
VIMA-Data [SC] &
Prompt-driven VL grounding across multiple compositional task types (six prompt modalities).\\
\addlinespace

ACT~\cite{zhao2023learning} &
ResNet-18 / --- / CVAE-Transformer &
ALOHA [SC] &
Temporal ensembling enables smooth bimanual imitation with fine control precision.\\
\addlinespace

Octo~\cite{team2024octo} &
CNN / T5-base / Diffusion Transformer &
Open X-Embodiment (OXE) &
Large multi-robot policy trained on 4M+ trajectories spanning many robot embodiments.\\
\addlinespace

VoxPoser~\cite{huang2023voxposer} &
ViLD + MDETR / GPT-4 / MPC (LLM-guided planning) &
Zero-shot &
Composes LLM+VLM for constraint-aware motion planning without task-specific training.\\
\addlinespace

Diffusion Policy~\cite{chi2023diffusion} &
ResNet-18 / --- / U-Net or Transformer diffusion &
Self-collected [SC] &
Diffusion modeling captures multimodal action distributions for robust visuomotor control.\\
\addlinespace

OpenVLA~\cite{kim2024openvla} &
DINOv2 + SigLIP / Prismatic-7B / symbol-tuning &
OXE + DROID &
Open-source RT-2-like VLA; supports efficient LoRA adaptation and broad generalization.\\
\addlinespace

$\pi^0$ (Pi-Zero)~\cite{black2024pi_0} &
PaliGemma VLM / PaliGemma (multimodal) / 300M diffusion action model &
Pi-Cross-Embodiment &
Lightweight general robot controller (reported $\sim$3B total) with strong cross-robot, open-world generalization and bimanual skills.\\
\addlinespace

$\pi^0$-Fast~\cite{pertsch2025fast} &
PaliGemma VLM / PaliGemma / autoregressive transformer with FAST tokenization &
Pi-Cross-Embodiment &
High-frequency real-time control via compressed frequency-space action tokens (reported up to 15$\times$ faster inference).\\
\addlinespace

OpenVLA-OFT~\cite{kim2025fine} &
SigLIP + DINOv2 (multi-view) / Llama-2 7B / parallel decoding + action chunking (L1 regression) &
LIBERO; bimanual ALOHA &
Fine-tuning recipe with parallel decoding and chunked actions; reported 97.1\% LIBERO success and 26$\times$ faster inference for high-frequency bimanual control.\\
\addlinespace

RDT-1B~\cite{liu2024rdt} &
Multi-view RGB encoder / transformer language module / Diffusion Transformer (unified action space) &
46 datasets ($>$1M episodes) + ALOHA fine-tune &
1.2B diffusion foundation model for dexterous bimanual manipulation with strong language conditioning and zero-shot transfer.\\
\addlinespace

Helix\footnote{\url{https://www.figure.ai/news/helix}} &
System~2: open-source VLM for multimodal reasoning (7--9\,Hz) / integrated semantics / System~1: transformer visuomotor policy (200\,Hz, full upper-body) &
Figure robot E2E (pixels+language$\rightarrow$actions) &
Humanoid-focused dual-rate VLA enabling real-time high-DoF control, dexterity, and collaborative multi-robot manipulation with zero-shot generalization.\\
\addlinespace

CogACT~\cite{li2024cogact} &
DINOv2 ViT-L/14 + SigLIP ViT-So400M/14 / Llama-2 via Prismatic-7B / DiT-Base (300M diffusion) &
OXE subset; Realman \& Franka tasks &
Componentized VLA with diffusion action transformer; reported +59.1\% real-world success vs. OpenVLA and strong adaptation to unseen robots/objects.\\
\addlinespace

Chain-of-Affordance (CoA)~\cite{li2024improving} &
Affordance-aware visual encoder / transformer reasoning prompts / autoregressive + diffusion policy (affordance-conditioned) &
LIBERO; real+sim manipulation &
Sequential affordance reasoning (object$\rightarrow$grasp$\rightarrow$spatial$\rightarrow$motion) improves spatial planning and obstacle avoidance; reported stronger LIBERO performance than OpenVLA.\\
\addlinespace

Edge VLA (EVLA)~\cite{budzianowski20edgevla} &
SigLIP + DINOv2 / Qwen2 (0.5B) / non-autoregressive joint control prediction &
Bridge; OXE; 1.2M text--image pairs &
Edge-optimized VLA (e.g., Jetson-class) with reported 30--50\,Hz inference and OpenVLA-comparable performance under low power.\\
\addlinespace

ShowUI-2B~\cite{lin2024showui} &
UI-guided visual token selection / interleaved V--L--A streaming / transformer GUI action predictor &
256K GUI instruction-following &
Compact 2B VLA for digital automation; strong screenshot grounding and GUI/web navigation with efficient token selection.\\
\addlinespace

GR00T N1~\cite{bjorck2025gr00t} &
NVIDIA Eagle-2 VLM / integrated high-level planning / diffusion transformer (DiT) &
Human demos + robot trajectories + simulation + internet video &
Generalist humanoid dual-system design combining planning and diffusion execution for dexterous multi-step control and broad embodiment generalization.\\
\addlinespace

Seer~\cite{gu2023seer} &
Grounding-optimized visual backbone / transformer language / autoregressive action head &
LIBERO &
Strong visual grounding for manipulation; competitive on LIBERO but typically below newer fine-tuned variants (e.g., OpenVLA-OFT).\\
\addlinespace

DiffusionVLA~\cite{wen2024diffusion} &
Transformer visual encoder / autoregressive reasoning / diffusion action head &
LIBERO; factory sorting; zero-shot bin-picking &
Diffusion control improves robustness and interpretability; reported weaker spatial generalization than CoA in some configurations.\\
\addlinespace

NaVILA~\cite{cheng2024navila} &
CLIP + CNN / LLaMA-2 / hierarchical control: topological planner + RL locomotion &
Real-world legged robot nav demos &
Modular hierarchy for terrain generalization; reported 88\% real-world navigation success from natural language.\\
\addlinespace

RoboNurse-VLA~\cite{li2024robonurse} &
SAM2 + RGB-D / LLaMA-2 + voice-to-text / pose regression + gripper classifier &
Surgical handover videos + voice prompts &
Real-time surgical tool handover robust to novel tools and dynamic operating-room scenes.\\
\addlinespace

Mobility VLA~\cite{chiang2024mobility} &
Long-context ViT + goal-image encoder / T5-based instruction encoder / graph planner + visual goal localization &
MINT dataset (VL instruction tours) &
Topological mapping from multimodal tours enables navigation generalization across large unseen spaces.\\
\addlinespace

TinyVLA~\cite{wen2025tinyvla} &
FastViT / compact 128-d language / diffusion decoder (50M) &
Mini-ALOHA + SC tasks &
No large-scale pretraining required; reported faster inference (5$\times$) with strong precision under tight compute budgets.\\
\addlinespace

QUAR-VLA~\cite{ding2024quar} &
CLIP + proprioceptive embedding / BERT + grounding adapter / transformer full-body decoder &
QUART (locomotion+manipulation) &
Quadruped-centric VLA with strong sim-to-real transfer and fine-grained instruction alignment.\\
\addlinespace

ChatVLA~\cite{zhou2025chatvla} &
Phase-aligned vision encoder / Prismatic MoE LLM / unified V--L--A planner &
Unified chat-action (web+robot) &
Joint VQA+planning; mitigates forgetting and supports conversational task execution with manipulation.\\
\addlinespace

PointVLA~\cite{li2025pointvla} &
CLIP + 3D point cloud fusion / LLaMA-2 / transformer with spatial token fusion &
Few-shot spatial tasks (real+sim) &
Improves long-horizon spatial reasoning by injecting 3D structure while preserving pretrained 2D knowledge.\\
\addlinespace

VLA-Cache~\cite{xu2025vla} &
SigLIP + token memory buffer / Prismatic-7B / transformer with dynamic token reuse &
ALOHA + sim/real fusion &
Caches static visual tokens for efficiency; reported 40--50\% faster inference with negligible performance loss.\\
\addlinespace

HybridVLA~\cite{liu2025hybridvla} &
CLIP + DINOv2 / LLaMA-2 / hybrid diffusion + autoregressive ensemble &
RT-X + synthetic fusion &
Dynamic ensemble improves robustness in multi-arm settings and supports stronger sim-to-real generalization.\\
\addlinespace

MoLe-VLA~\cite{zhang2025mole} &
Multi-stage ViT + STAR router / CogKD-enhanced transformer / sparse transformer (dynamic routing) &
RLBench + real-world manipulation &
Selective layer activation yields efficiency (reported 5.6$\times$ speedup) and higher success (reported +8\%).\\
\addlinespace

UAV-VLA~\cite{sautenkov2025uav} &
ViT for aerial imagery / GPT instruction parsing / transformer path planner &
Satellite + UAV imagery instructions &
Zero-shot aerial task planning with scalable language grounding for large unmapped environments.\\
\addlinespace

DexGraspVLA~\cite{zhong2025dexgraspvla} &
Object-centric spatial ViT / transformer grasp reasoning / diffusion grasp controller &
Dexterous grasping benchmark (sim+real) &
Reported $>$90\% zero-shot success across diverse objects; robust to lighting/background variation and unseen conditions.\\
\addlinespace

GraspVLA~\cite{deng2025graspvlagraspingfoundationmodel} &
Multi-view DINOv2 + SigLIP / VLM predicts boxes+grasps / flow-matching action expert (PAG) &
SynGrasp-1B; GRIT &
Synthetic-pretrained grasping VLA; improves sim-to-real transfer and supports zero-/few-shot generalization to long-tail object classes and preferences.\\
\addlinespace

Interleave-VLA~\cite{fan2025interleavevlaenhancingrobotmanipulation} &
InternVL2.5 + OWLv2 / Qwen2.5 / continuous action predictor (OpenVLA+$\pi^0$-style, diffusion controller) &
Open Interleaved X-Embodiment (210k eps., 11 datasets) &
End-to-end interleaved image--text instruction following; reported 2--3$\times$ out-of-domain gains and zero-shot execution from sketches and novel multimodal prompts.\\
\addlinespace

Long-VLA~\cite{fan2025long} &
End-to-end VLA for long-horizon tasks / phase-aware input masking + transformer policy (subtask phase segmentation) &
Long-horizon multi-step robot manipulation demonstrations (task sequences) &
Targets long-horizon execution by explicitly separating ``moving'' vs. ``interaction'' phases, improving subtask compatibility and robustness over extended task horizons.\\
\addlinespace

RetoVLA~\cite{koo2025retovla} &
VLM-based policy / reuses register tokens as spatial context for action prediction &
Real-robot manipulation on a 7-DoF arm + task-specific demonstrations &
Lightweight spatial-reasoning upgrade by repurposing register tokens; reported sizable success-rate gains on complex manipulation with minimal architectural overhead.\\
\addlinespace

Vlaser~\cite{yang2025vlaser} &
VLM-to-VLA pipeline / synergistic embodied reasoning + policy learning (reasoning-aware VLA fine-tuning) &
Vlaser-6M embodied reasoning dataset + VLA fine-tuning data (robot demonstrations) &
Bridges embodied reasoning and control: strong performance across embodied grounding/QA/planning while improving transfer to policy learning under domain shift.\\
\addlinespace

Discrete Diffusion VLA~\cite{liang2025discrete} &
Single-transformer VLA / discretized action chunks + discrete diffusion refinement (CE training; remasking) &
LIBERO + SimplerEnv (Fractal/Bridge) style VLA benchmarks &
Unifies diffusion-style refinement with discrete token interfaces: adaptive decoding order, error correction via remasking, and reduced autoregressive bottlenecks with strong benchmark success rates.\\
\addlinespace

Being-H0~\cite{luo2025being} &
Dexterous VLA pretrained from human videos / explicit hand-motion modeling + VL grounding for action &
Large-scale human manipulation videos (egocentric/third-person) + transfer to robot control &
Scales dexterous manipulation by leveraging diverse human video data; improves generalization to novel tasks/scenes compared with small teleop-only robot datasets.\\
\addlinespace

EgoVLA~\cite{yang2025egovla} &
VLM pretraining on egocentric human manipulation / unified human--robot action space + robot fine-tuning &
Large-scale egocentric human videos + small set of robot demonstrations &
Uses abundant egocentric videos for scalable pretraining, then aligns embodiments via unified action space to enable practical robot transfer with limited robot data.\\
\addlinespace

StereoVLA~\cite{deng2025stereovla} &
Stereo-enhanced VLA / geometric-semantic fusion from stereo pairs (+ auxiliary depth cues) for action decoding &
Stereo robot manipulation demonstrations (stereo RGB with language-conditioned actions) &
Explicitly exploits stereo geometry to improve spatial precision (depth-sensitive grasping/manipulation) and robustness to viewpoint/camera variations.\\
\addlinespace

EfficientVLA~\cite{yang2025efficientvla} &
Training-free VLA acceleration/compression / structured redundancy removal across VLA pipeline (token/interface optimizations) &
Applies to existing VLAs (evaluation on standard manipulation benchmarks; e.g., SIMPLER-style settings) &
Practical deployability: speeds up and reduces compute/memory of large VLA policies with minimal accuracy drop, enabling closer-to-real-time inference on constrained hardware.\\
\addlinespace

\end{longtable}
\end{center}

\clearpage
\twocolumn

\subsection{Training Efficiency Advancements in Vision–Language–Action Models}

VLA models have seen rapid progress in training and optimization techniques to reconcile multimodal inputs, reduce compute requirements, and enable real-time control. Key areas of advancement include:

\begin{itemize}
  \item \textbf{Data-Efficient Learning.}  
    \begin{itemize}
      \item \emph{Co-fine-tuning} on massive vision–language corpora (e.g.\ LAION-5B) and robotic trajectory collections (e.g.\ Open X-Embodiment) aligns semantic understanding with motor skills. OpenVLA (7 B parameters) achieves a 16.5 \% higher success rate than a 55 B-parameter RT-2 variant, demonstrating that co-fine-tuning yields strong generalization with fewer parameters~\cite{schuhmann2022laion, vuong2023open, kim2024openvla} compared to compared to scaling model size alone.  
      \item \emph{Synthetic Data Generation} via UniSim produces photorealistic scenes including occlusions and dynamic lighting to augment rare edge-case scenarios, improving model robustness in cluttered environments by over 20 \%~\cite{yang2023unisim, team2024octo}.  
      \item \emph{Self-Supervised Pretraining} adopts contrastive objectives (à la CLIP) to learn joint visual–text embeddings before action fine-tuning, reducing reliance on task-specific labels. Qwen2-VL leverages self-supervised alignment to accelerate downstream grasp-and-place convergence by 12 \%~\cite{radford2018improving, huang2023language}.  
    \end{itemize}

  \item \textbf{Parameter-Efficient Adaptation.}  
    Low-Rank Adaptation (LoRA) inserts lightweight adapter matrices into frozen transformer layers, cutting trainable weights by up to 70 \% while retaining performance~\cite{hu2022lora}. The Pi-0 Fast variant uses merely 10 M adapter parameters atop a static backbone to deliver continuous 200 Hz control with negligible accuracy loss~\cite{pertsch2025fast}.

  \item \textbf{Inference Acceleration.}  
    \begin{itemize}
      \item \emph{Compressed Action Tokens} (FAST) and \emph{parallel decoding} in dual-system frameworks (e.g.\ GR00T N1) achieve up to 2.5$\times$ faster policy inference, reducing per-step latency to below 5\,ms compared to standard autoregressive decoding in single-policy VLAs evaluated on real-time manipulation and humanoid control benchmarks. This acceleration comes at a modest cost in trajectory smoothness, manifested as slightly higher action discretization error and reduced fine-grained motion continuity under high-frequency control \cite{bjorck2025gr00t, song2025accelerating}.

    \end{itemize}
\end{itemize}

Together, these methods have transformed VLAs into practical agents capable of handling language-conditioned, vision-guided tasks in dynamic, real-world settings.

\subsection{Parameter-Efficient Methods and Acceleration Techniques in VLA Models}

Beyond data-efficient learning strategies, a complementary line of research in VLA models focuses on reducing model size, memory footprint, and inference latency to enable deployment on real robotic platforms with limited computational and power resources. Unlike training-centric efficiency methods discussed earlier, the techniques in this subsection primarily target \emph{parameter efficiency at adaptation time} and \emph{runtime acceleration during policy inference}.

\begin{enumerate}

\item \textbf{Parameter-Efficient Adaptation via Low-Rank Modules:}
Rather than full fine-tuning, many VLAs adopt parameter-efficient adaptation mechanisms such as Low-Rank Adaptation (LoRA), which was introduced earlier in the context of training efficiency. In this section, we emphasize its role in \emph{reducing the effective parameter footprint during deployment}. For example, OpenVLA employs lightweight LoRA adapters (approximately 20\,M parameters) on top of a frozen 7\,B-parameter backbone, allowing task adaptation with minimal memory overhead and avoiding duplication of full model weights across tasks~\cite{hu2022lora, kim2024openvla}. This design enables multiple task-specialized policies to coexist on resource-constrained systems while preserving the general-purpose visual--semantic representations learned during pretraining.

\item \textbf{Quantization for Edge Deployment:}
Model quantization reduces numerical precision to improve inference throughput and memory efficiency. Experiments with OpenVLA show that INT8 quantization on embedded platforms such as NVIDIA Jetson Orin preserves approximately 97\,\% of full-precision task success on pick-and-place benchmarks, with only minor degradation in fine-grained dexterous manipulation~\cite{schuhmann2022laion, kim2024openvla}. Post-training quantization and per-channel calibration further mitigate accuracy loss under high-dynamic-range sensory inputs~\cite{oquab2023dinov2}. These optimizations enable sustained control frequencies of up to 30\,Hz within strict power budgets typical of mobile robots.

\item \textbf{Model Pruning and Architectural Slimming:}
Structured pruning removes redundant architectural components such as attention heads or feed-forward sublayers to reduce memory and compute requirements. Although less explored in VLAs than in standalone vision or language models, early studies on diffusion-based visuomotor policies indicate that pruning up to 20\,\% of convolutional vision encoders results in negligible degradation in grasp stability~\cite{chi2023diffusion}. Similar pruning strategies applied to transformer-based VLAs (e.g.\ RDT-1B) can reduce memory footprint by approximately 25\,\% with less than 2\,\% drop in task success, enabling sub-4\,GB deployments~\cite{liu2024rdt, li2024cogact}.

\item \textbf{Compressed Action Tokenization:}
To address inference bottlenecks arising from long-horizon control sequences, compressed action representations have been proposed. FAST reformulates continuous action trajectories into compact frequency-domain tokens, substantially reducing decoding length. The Pi-0 Fast variant achieves up to 15$\times$ faster inference by compressing 1000\,ms action windows into 16 discrete tokens, enabling control rates of up to 200\,Hz on desktop GPUs~\cite{pertsch2025fast}. This method trades minimal trajectory granularity for large speedups, making it well suited for high-frequency, reactive manipulation tasks.

\item \textbf{Parallel Decoding and Action Chunking:}
Standard autoregressive VLAs decode actions sequentially, leading to cumulative latency. Parallel decoding strategies, as employed in dual-system architectures such as GR00T~N1, generate groups of spatio-temporal action tokens concurrently, achieving approximately 2.5$\times$ reduction in end-to-end inference latency on 7-DoF robotic arms operating at 100\,Hz~\cite{bjorck2025gr00t, song2025accelerating}. Action chunking further abstracts multi-step routines (e.g.\ \emph{pick-and-place}) into single high-level tokens, reducing inference steps by up to 40\,\% in long-horizon manipulation tasks such as kitchen workflows~\cite{jiang2022vima}.

\item \textbf{Hardware-Aware Compilation and Runtime Optimization:}
Finally, hardware-aware optimizations leverage compiler-level graph rewrites, kernel fusion, and accelerator-specific primitives to maximize throughput. Frameworks such as TensorRT-LLM exploit tensor cores, fused attention kernels, and pipelined memory transfers to accelerate both transformer inference and diffusion sampling. In OpenVLA-OFT, such optimizations reduce inference latency by approximately 30\,\% and lower energy consumption per inference by 25\,\% on RTX-class GPUs compared to standard PyTorch execution~\cite{kim2025fine}. These system-level optimizations are critical for deploying real-time VLAs on mobile robots, aerial platforms, and humanoid systems with strict power constraints.

\end{enumerate}

\textbf{Discussion:}  
Parameter‐efficient adaptation and inference acceleration techniques collectively democratize VLA deployment:

\begin{itemize}
  \item LoRA and quantization empower smaller labs and research/development programs to fine‐tune and operate billion‐parameter VLAs on consumer‐grade hardware, unlocking cutting‐edge semantic understanding for robots~\cite{hu2022lora, kim2024openvla}.  
  \item Pruning and FAST tokenization compress model and action representations, enabling sub‐4 GB, sub‐5 ms control loops without sacrificing precision in dexterous tasks~\cite{liu2024rdt, pertsch2025fast}.  
  \item Parallel decoding and action chunking overcome sequential bottlenecks of autoregressive policies, supporting 100–200 Hz decision rates needed for agile manipulation and legged locomotion~\cite{bjorck2025gr00t, song2025accelerating}.  
  \item Hybrid RL‐SL training stabilizes exploration in complex environments, while hardware‐aware compilation ensures real‐time performance on edge accelerators~\cite{moroncelli2024integrating, kim2025fine}.  
\end{itemize}

Together, these advances make it practical to embed VLA models across industrial manipulators, assistive drones, and consumer robots, bridging the gap from research prototypes to real‐world autonomy.  

\subsection{Applications of Vision-Language-Action Models}

VLA models are rapidly emerging as foundational building blocks for embodied intelligence, integrating perception, natural language understanding, and motor control within a unified architecture. By encoding visual and linguistic modalities into shared semantic spaces and generating contextually grounded actions, VLA models enable seamless interaction between agents and their environments \cite{li2024cogact, zhou2025opendrivevla}. By encoding visual and linguistic modalities into shared semantic spaces and generating contextually grounded actions, VLA models enable seamless interaction between agents and their environment. This multimodal capacity has positioned VLAs as transformative agents across a wide spectrum of real-world applications. In humanoid robotics, systems like Helix and RoboNurse-VLA combine vision, language, and dexterous manipulation to assist with domestic tasks and surgical operations, demonstrating real-time reasoning and safety-aware control \cite{li2024robonurse, wen2025tinyvla}. In autonomous vehicles, models such as OpenDriveVLA and ORION process dynamic visual streams and natural language instructions to make transparent, adaptive driving decisions in complex urban environments \cite{fu2025orion, zhou2025opendrivevla}. Industrial deployments leverage VLA architectures for high-precision assembly, inspection, and collaborative manufacturing \cite{li2024cogact}. In agriculture, VLA-powered robotic systems could enable vision-guided fruit harvesting, plant monitoring, and anomaly detection, reducing labor dependency and increasing sustainability. Furthermore, recent advances in interactive augmented reality systems utilize VLA models for real-time, language-conditioned spatial navigation, guiding users in indoor and outdoor settings based on voice or visual cues \cite{sautenkov2025uav, gbagbe2024bi}. Across these domains, VLAs offer a unified framework for robust, adaptable, and semantically aligned task execution, marking a pivotal shift toward embodied generalist agents.

Table \ref{tab:vla_comparison} shows the recent VLA models by summarizing their methodologies, application domains, and key innovations.

\begin{table*}[h!]
\centering
\scriptsize
\setlength{\tabcolsep}{3.5pt}
\renewcommand{\arraystretch}{1.15}
\caption{Comparison of representative Vision-Language-Action (VLA) methodologies, application areas, and key innovations.}
\label{tab:vla_comparison}
\begin{tabular}{p{2.3cm} p{5.0cm} p{3.4cm} p{6.7cm}}
\toprule
\textbf{Reference (Year)} & \textbf{VLA methodology} & \textbf{Application area} & \textbf{Strength / key innovation} \\
\midrule
CogACT \cite{li2024cogact} (2024) &
Modular VLA with a specialized action module based on diffusion transformers.&
Industrial robotics; language-guided manipulation. &
Robust action modeling with fast adaptation and strong generalization, improving task success in diverse industrial settings. \\
\addlinespace

VLATest \cite{wang2024towards} (2024) &
Automated large-scale testing framework for VLA manipulation evaluation. &
Robotic manipulation benchmarking (robustness/reliability). &
Systematic multi-scene and multi-task evaluation that exposes robustness gaps and supports targeted VLA improvements. \\
\addlinespace

NaVILA \cite{cheng2024navila} (2024) &
Two-level VLA: high-level vision-language generates mid-level navigation commands; RL locomotion executes. &
Legged navigation from natural language in cluttered real-world scenes. &
Modular high/low-level split with strong generalization and high real-world success across challenging terrains. \\
\addlinespace

RoboNurse-VLA \cite{li2024robonurse} (2024) &
Vision module (SAM2) + language module (Llama2) with real-time voice-to-action pipeline. &
Surgical assistance (instrument grasp and handover). &
Accurate real-time assistance, robustness to unseen tools, and stable performance in dynamic operating room conditions. \\
\addlinespace

Mobility VLA \cite{chiang2024mobility} (2024) &
Hierarchical VLA with long-context VLM for goal localization and topological graph navigation. &
Multimodal instruction navigation with demonstration tours (MINT). &
Leverages demonstrations for scalable navigation in large spaces; robust to novel language+image queries. \\
\addlinespace

CoVLA \cite{arai2025covla} (2025) &
CLIP-based vision, Llama-2 language, trajectory prediction for action. &
Autonomous driving (dataset + VLA training). &
Large-scale richly annotated dataset enabling interpretable scene understanding and robust path planning. \\
\addlinespace

OpenDriveVLA \cite{zhou2025opendrivevla} (2025) &
Hierarchical alignment of 2D/3D visual tokens and language embeddings; autoregressive agent--environment--ego modeling. &
End-to-end autonomous driving. &
Unified semantic space and interaction modeling that improves planning and QA performance in complex traffic scenes. \\
\addlinespace

ORION \cite{fu2025orion} (2025) &
QT-Former for history context, LLM reasoning, generative planner for trajectory prediction. &
Holistic end-to-end autonomous driving. &
Aligns reasoning and action spaces with unified optimization for VQA and planning, yielding stronger closed-loop performance. \\
\addlinespace

QUAR-VLA \cite{ding2024quar} (2025) &
QUART-based fusion of vision and language for action generation. &
Quadruped navigation, manipulation, whole-body tasks. &
Tight VL-A coupling and instruction alignment with strong sim-to-real generalization for legged platforms.\\
\addlinespace

TinyVLA \cite{wen2025tinyvla} (2025) &
Compact multimodal backbone with diffusion-policy decoder. &
Fast, data-efficient manipulation control. &
Reduced inference cost with strong generalization; improves efficiency without requiring heavy pre-training. \\
\addlinespace

UAV-VLA \cite{sautenkov2025uav} (2025) &
Modular pipeline: GPT for goal extraction, VLM for object search, GPT for action generation. &
UAV mission planning from language + satellite imagery. &
Efficient flight/action planning with minimal prior training; supports intuitive human--UAV interaction and benchmarking. \\
\addlinespace

Bi-VLA \cite{gbagbe2024bi} (2025) &
Multimodal transformer linking vision, language, and bimanual action modules. &
Bimanual household manipulation. &
High adaptability and robust real-world bimanual execution through explicit action-module integration. \\
\addlinespace

ChatVLA \cite{zhou2025chatvla} (2025) &
Phased alignment training with Mixture-of-Experts for VL-A integration.&
Unified multimodal understanding and robot control. &
Reduces forgetting/interference, improving both VQA and manipulation with efficient specialization. \\
\addlinespace

RoboMamba \cite{liu2024robomamba} (2025) &
Mamba/SSM-based VLA with co-trained vision encoder and SE(3) action modeling. &
Efficient robotic reasoning and manipulation. &
Linear-complexity inference and minimal fine-tuning enable fast pose-aware reasoning and efficient control. \\
\addlinespace

OTTER \cite{huang2025otter} (2025) &
Text-aware feature extraction using frozen pre-trained VLMs. &
Manipulation with zero-shot generalization. &
Preserves VL semantic alignment without VLM fine-tuning via task-relevant feature selection for strong zero-shot transfer. \\
\addlinespace

PointVLA \cite{li2025pointvla} (2025) &
Injects 3D point-cloud features into frozen VLA via modular skip-blocks. &
Spatial reasoning; few-shot and long-horizon manipulation. &
Adds 3D geometry while preserving 2D knowledge, improving spatial grounding without full retraining. \\
\addlinespace

VLA-Cache \cite{xu2025vla} (2025) &
Adaptive token caching with selective reuse of static visual tokens. &
Real-time efficient manipulation inference. &
Layer-wise token reuse yields major speedups with minimal accuracy loss, practical for on-robot deployment. \\
\addlinespace

CombatVLA \cite{chen2025combatvla} (2025) &
Video-action AoT training with truncated AoT for fast inference. &
Real-time combat decision-making in 3D games. &
Large inference acceleration with improved tactical reasoning and high success in real-time interactive environments. \\
\addlinespace

HybridVLA \cite{liu2025hybridvla} (2025) &
Unified LLM with collaborative diffusion and autoregressive action policies. &
Single-/dual-arm manipulation across sim and real tasks. &
Adaptive action ensembling improves robustness and generalization on complex manipulations. \\
\addlinespace

NORA \cite{hung2025nora} (2025) &
3B-parameter VLA using Qwen-2.5-VL-3B backbone with FAST+ tokenizer. &
Generalist embodied robotics (sim + real). &
Low overhead with strong visual reasoning and fast action decoding; competitive with larger VLAs. \\
\addlinespace

SpatialVLA \cite{qu2025spatialvla} (2025) &
Ego3D position encoding and adaptive action grids for spatially-aware VLA. &
Cross-robot, multi-task, zero-shot manipulation. &
Explicit 3D spatial integration and adaptive discretization improve transfer and generalization; open-sourced. \\
\addlinespace

MoLe-VLA \cite{zhang2025mole} (2025) &
Mixture-of-Layers with dynamic layer-skipping via router and distillation. &
Efficient manipulation on RLBench + real robots. &
Selective layer activation provides strong speedups while maintaining cognition and improving success. \\
\addlinespace

JARVIS-VLA \cite{li2025jarvis} (2025) &
Post-trained large VLMs with VL guidance and an action head for keyboard/mouse control.&
Open-world visual games (e.g., Minecraft), 1k+ tasks. &
Self-supervised post-training yields strong generalization and world knowledge; open-sourced task suite and models. \\
\addlinespace

UP-VLA \cite{zhang2025up} (2025) &
Unified VLA with joint multimodal understanding and future prediction objectives. &
Embodied manipulation with precise spatial reasoning. &
Joint semantics+dynamics learning improves performance on long-horizon tasks requiring fine spatial control. \\
\addlinespace

Shake-VLA \cite{khan2025shake} (2025) &
Modular stack with vision, speech-to-text, RAG, anomaly detection, and bimanual arms. &
Bimanual cocktail mixing in clutter/noise. &
Robust end-to-end deployment with reliable ingredient handling, recipe adaptation, and real-world task completion. \\
\addlinespace

MoRE \cite{zhao2025more} (2025) &
Sparse MoE with LoRA modules and RL-based Q-function training. &
Quadruped multi-task locomotion, navigation, manipulation. &
Scalable fine-tuning on mixed-quality data with strong multi-task and OOD generalization in sim and real. \\
\addlinespace

DexGraspVLA \cite{zhong2025dexgraspvla} (2025) &
Hierarchical planner (pre-trained VL) + diffusion low-level controller. &
General dexterous grasping in diverse conditions. &
Domain-invariant representations support strong zero-shot generalization across objects, lighting, and backgrounds. \\
\addlinespace

DexVLA \cite{wen2025dexvla} (2025) &
Plug-in diffusion action expert with embodiment curriculum learning. &
General robot control: single-arm, bimanual, dexterous hand. &
Cross-embodiment action modeling and fast adaptation enable strong long-horizon performance without task-specific tuning. \\
\bottomrule
\end{tabular}
\end{table*}

The following subsections chronologically explore the application areas in depth as shown in Figure \ref{fig:applications-mindmap}.

\begin{figure}[ht!]
  \centering
  \begin{tikzpicture}[
      mindmap,
      grow cyclic,
      clockwise from=90,
      every node/.style=concept,
      concept color=blue!20,
      level 1/.append style={
        level distance=4cm,
        sibling angle=60
      }
    ]
    \node[concept color=blue!40] {Applications of VLA}
      child { node {Humanoid Robotics} }
      child { node {Autonomous Vehicle Systems} }
      child { node {Industrial Robotics} }
      child { node {Healthcare \& Medical Robotics} }
      child { node {Precision \& Automated Agriculture} }
      child { node {Interactive AR Navigation} };
  \end{tikzpicture}
  \caption{Mind-map of application domains for Vision--Language--Action models, with Humanoid Robotics positioned at the top and remaining domains arranged clockwise to match the order of discussion in this section.}
  \label{fig:applications-mindmap}
\end{figure}

\subsubsection{Humanoid Robotics}
Humanoid robots, designed to mimic the form and functionality of the human body, represent one of the most demanding yet impactful domains for the deployment of VLA models. These platforms must seamlessly perceive complex environments, understand spoken or written natural language, and perform intricate physical tasks with human-level dexterity \cite{roychoudhury2023perception, cao2024ai}. The core strength of VLA models lies in their ability to unify perception, cognition, and control into a single, end-to-end trainable framework allowing humanoid robots to interpret visual inputs (e.g., RGB-D imagery of cluttered scenes), comprehend linguistic instructions (e.g., “place the spoon in the drawer”), and generate precise motor trajectories \cite{lu2023towards,zhu2020robot}.

Recent advances have significantly accelerated the deployment of VLAs in humanoid robotics. For example, Helix\footnote{\url{https://www.figure.ai/news/helix}}, a humanoid robot developed by Figure AI, leverages a fully integrated VLA model to perform full-body manipulation at high frequency, controlling arms, hands, torso, and even fine-grained finger motion in real time. The architecture follows a dual-system design: a multimodal transformer processes inputs such as language commands and vision streams, while a real-time motor policy outputs dense action vectors at 200 Hz. This allows Helix to generalize across previously unseen objects and tasks, adapting fluidly to changing environments without the need for task-specific retraining.

The key advantage of VLAs in humanoid systems is their ability to scale across diverse tasks using shared representations \cite{asuzu2025human}. Unlike traditional robotic systems that rely on task-specific programming or modular pipelines, VLA-powered humanoids operate under a unified token-based framework. Vision inputs are encoded via pretrained vision-language models like DINOv2 or SigLIP, while instructions are processed using LLMs such as LLaMA or GPT-style encoders. These representations are fused into prefix tokens that capture the full context of the scene and task. Action tokens are then generated autoregressively, similar to language decoding, but represent motor commands for the robot’s joints and end-effectors.

This capability enables humanoid robots to operate effectively in human-centric spaces, such as households, hospitals, and retail environments. In domestic settings, VLA-powered robots can clean surfaces, prepare simple meals, or organize objects simply by interpreting voice commands \cite{lu2023towards, zhu2020robot}. In healthcare, systems like RoboNurse-VLA \cite{li2024robonurse} have demonstrated the ability to perform precise instrument handovers to surgeons using real-time voice and visual cues. In retail, humanoid platforms equipped with VLAs can assist with customer queries, restock shelves, and navigate store layouts without explicit pre-programming \cite{asuzu2025human}.

What distinguishes modern humanoid VLAs is their ability to run on embedded, low-power hardware, making real-world deployment viable. For instance, systems such as TinyVLA \cite{wen2025tinyvla} and MoManipVLA \cite{wu2025momanipvla} demonstrate efficient inference pipelines that run on Jetson-class GPUs, enabling mobile deployment without compromising performance. These models exploit techniques like diffusion-based policies, LoRA-based fine-tuning, and dynamic token caching to minimize compute cost while retaining high precision and generalization.

In logistics and manufacturing, VLA-enabled humanoids are already making a commercial impact. Robots like Figure 01 are deployed in warehouses to perform repetitive, physically intensive tasks such as picking, sorting, and shelving alongside human workers. Their ability to handle novel object categories and dynamically changing scenes is powered by continual learning and robust multimodal grounding \cite{xu2025vla, li2024cogact}. 

As VLA models continue to advance in their capacity for diverse action generation, spatial reasoning, and real-time adaptation, humanoid robots are emerging as highly capable assistants across homes, industrial settings, and public spaces. Their strength lies in their ability to unify perception, language comprehension, and motor control through a shared token-based architecture enabling seamless, context-aware behavior in unstructured human environments.

\begin{figure*}[ht!]
\centering
\includegraphics[width=0.75\linewidth]{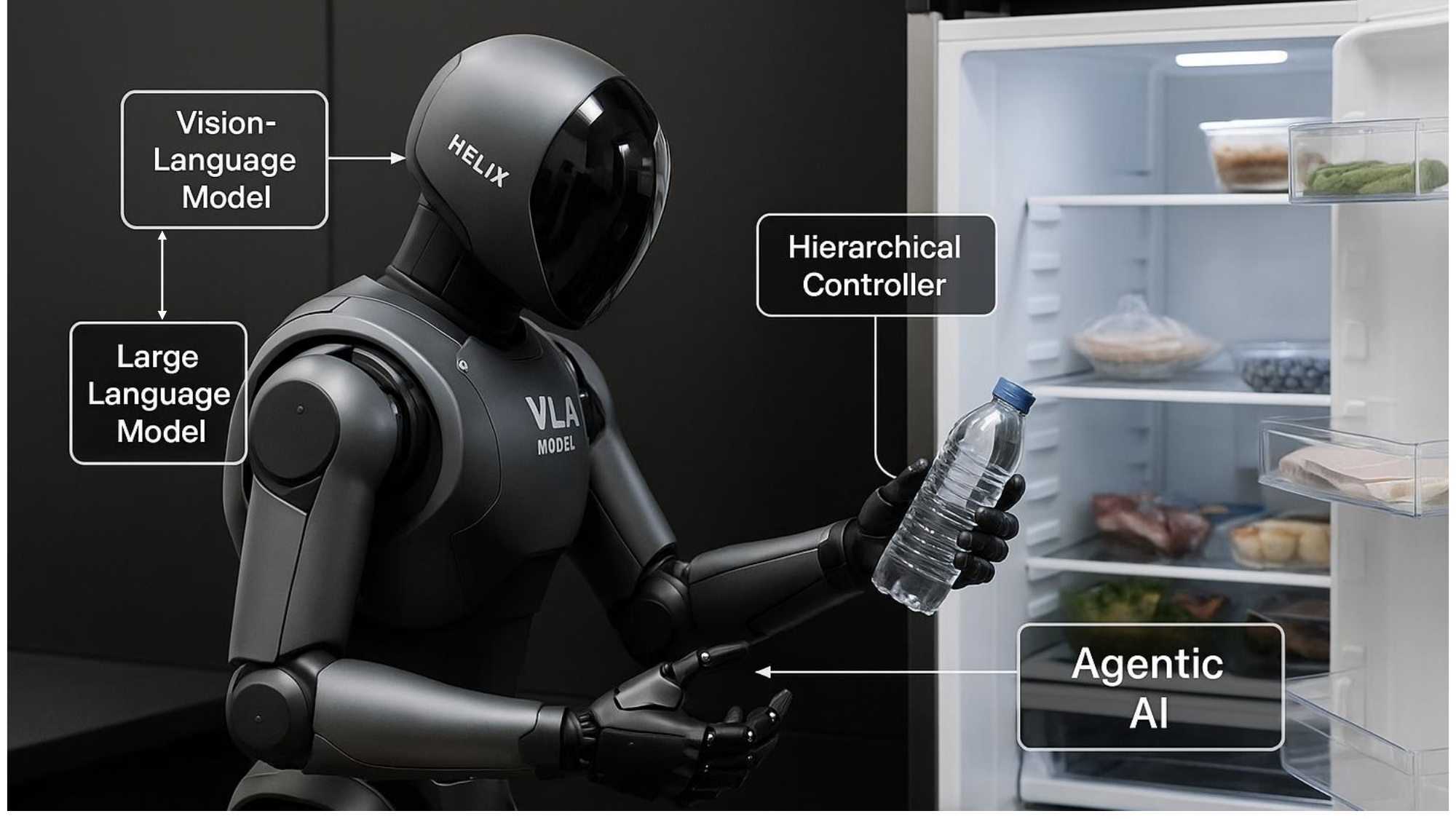}
\caption{This figure illustrates “Helix,” a next-generation humanoid robot executing a household task using a  VLA  framework. Upon receiving a verbal command, Helix integrates a vision-language model (e.g., SigLIP) and a language model (e.g., LLaMA-4) to jointly perceive and interpret the environment. A hierarchical VLA controller plans and executes sub-tasks opening the fridge, grasping a bottle while an agentic AI module adapts actions in real time. This demonstrates VLA-based generalist robotics with dynamic task adaptation and safe, semantically grounded manipulation.}
\label{fig:humanoidhelix}
\end{figure*}

For example, as depicted in the figure \ref{fig:humanoidhelix}, consider 'Helix', a state-of-the-art humanoid robot equipped with a next-generation VLA model. When instructed verbally, “Please take the water bottle from the fridge,” Helix activates its integrated perception system, where a foundation vision-language model (e.g., SigLIP or DINOv2) segments the visual scene to identify the refrigerator, its handle, and the bottle. The language input is processed by an LLM such as LLaMA-4, which tokenizes the instruction and fuses it with the visual context. This fused representation is passed to a hierarchical controller: the high-level policy plans the task sequence (locate handle, pull door, identify bottle, grasp), while a mid-level planner defines motor primitives, such as grasp type and joint trajectories. The low-level VLA controller, often based on diffusion policy networks, executes these actions with sub-second latency. Upon encountering variations (e.g., a tilted bottle or slippery grip), Helix's agentic AI module performs micro-policy refinement in real time, adjusting its grip based on feedback. This example illustrates the transformative potential of VLA-enabled humanoid. From kitchens to clinics, these systems not only interpret complex instructions and execute physical tasks with dexterity but also adapt to environmental unpredictability. By embedding agentic reasoning and safety alignment mechanisms, modern humanoid robots powered by VLAs are transitioning from narrow-task performers to generalist, trustworthy collaborators. As energy-efficient models like TinyVLA and MoManipVLA mature, deployment on mobile, low-power platforms becomes increasingly practical ushering in a new era of embodied, socially aligned AI.

\subsubsection{Autonomous Vehicle Systems}

Autonomous vehicles (AVs), including self-driving cars, trucks, and aerial drones, represent a frontier application domain for VLA models, where safety-critical decision-making demands tightly coupled perception, semantic understanding, and real-time action generation. Unlike traditional modular AV pipelines that explicitly separate perception, planning, and control, VLA frameworks explore tighter architectural coupling by jointly processing visual observations, high-level semantic cues, and internal state representations within a unified model. While such end-to-end formulations have shown promising results in simulation and controlled benchmarks for instruction-conditioned navigation and reasoning, large-scale commercial systems (e.g., Tesla Autopilot (\href{https://www.tesla.com/fsd}{Source Link})) continue to rely on modular or hybrid pipelines, with recent industry efforts focusing on integrating vision-language reasoning components rather than fully deploying VLA-style action generation in safety-critical driving.

VLA models empower AVs to comprehend complex environments beyond pixel-level object recognition. For instance, a self-driving car navigating an urban setting must detect traffic signs, understand pedestrian behavior, and interpret navigation commands such as “take the second right after the gas station.” These tasks involve fusing visual and linguistic signals to understand spatial relationships, predict intent, and generate context-aware driving actions. VLAs encode this information through token-based representations, where visual encoders (e.g., ViT, CLIP), language models (e.g., LLaMA-4), and trajectory decoders operate in a coherent semantic space, enabling the vehicle to reason about high-level goals and translate them into low-level motion.

A notable contribution in this direction is \textbf{CoVLA} \cite{arai2025covla}, which provides a comprehensive dataset pairing over 80 hours of real-world driving videos with synchronized sensor streams (e.g., LiDAR, odometry), detailed natural language annotations, and high-resolution driving trajectories. This dataset enables training VLA models to align perceptual and linguistic features with physical actions. CoVLA employs CLIP for visual grounding, LLaMA-2 for instruction embedding, and trajectory decoders for motion prediction. This configuration allows AVs to interpret verbal cues (e.g., “yield to ambulance”) and environmental conditions (e.g., merging traffic) to make transparent and safe driving decisions.

\textbf{OpenDriveVLA} \cite{zhou2025opendrivevla} advances the state of VLA modeling by integrating hierarchical alignment of 2D/3D multi-view vision tokens with natural language inputs. Its architecture leverages both egocentric spatial perception and external scene understanding to construct a dynamic agent-environment-ego interaction model. Through autoregressive decoding, OpenDriveVLA generates both action plans (e.g. steering angle, acceleration) and trajectory visualizations interpretable to humans. Its end-to-end framework achieves leading performance on public autonomous driving benchmarks, including planning and trajectory prediction tasks on the nuScenes and Waymo Open Motion datasets, as well as vision--language question-answering benchmarks for driving scenes, demonstrating strong robustness in urban navigation and behavioral prediction.

Another seminal model, \textbf{ORION} \cite{fu2025orion}, pushes the boundaries of closed-loop autonomous driving by incorporating a QT-Former to retain long-horizon visual context, an LLM to reason over traffic narratives, and a generative trajectory planner. ORION excels at aligning the discrete reasoning space of vision-language models with the continuous control space of AV motion. This unified optimization results in accurate visual question answering (VQA) and trajectory planning, crucial for scenarios involving ambiguous human instructions or occluded obstacles (e.g., “take the exit behind the red truck”).

\begin{figure}[ht!]
\centering
\includegraphics[width=0.99\linewidth]{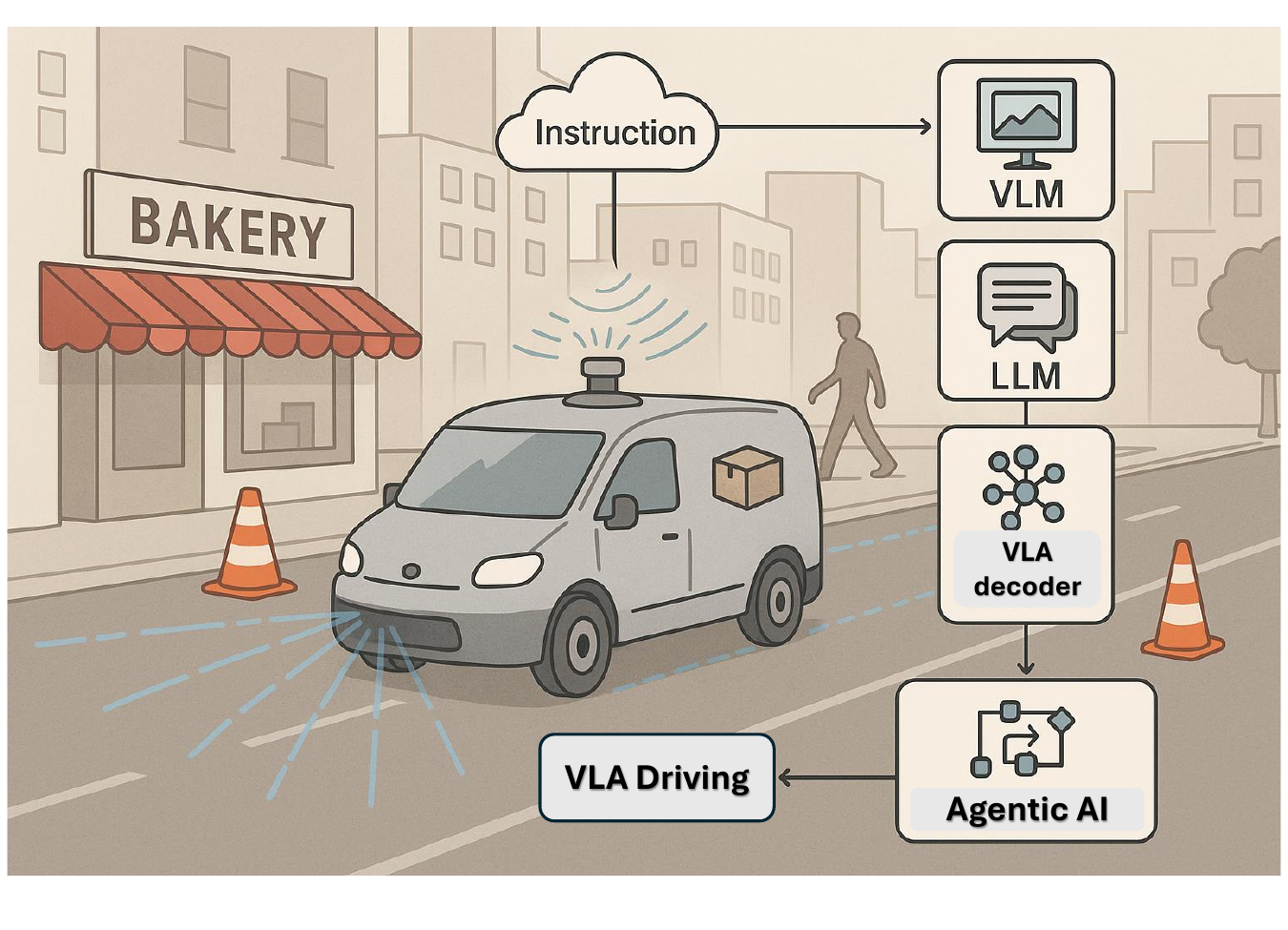}
\caption{This illustration depicts an autonomous delivery vehicle powered by a VLA system, integrating VLMs for visual grounding, LLMs for instruction parsing, and a VLA decoder for path planning. Agentic AI enables adaptive trajectory refinement in dynamic environments, exemplifying how multi-modal integration drives safe, interpretable, and autonomous decision-making in real-world navigation tasks.}
\label{fig:selfdrive}
\end{figure}

For example, as depicted in Figure \ref{fig:selfdrive}, consider an autonomous delivery vehicle, “AutoNav,” operating in a dense urban environment using a next-generation VLA architecture. As AutoNav receives a cloud-based instruction “Drop off the package near the red awning beside the bakery, then return to base avoiding construction zones”, its onboard VLM (e.g., CLIP or SigLIP) parses the visual stream from multiple cameras, identifying dynamic landmarks such as bakery signs, red awnings, and traffic cones. Simultaneously, the LLM module grounded in LLaMA-4 decodes the instruction and fuses it with real-time sensory context including LiDAR, GPS, and inertial odometry. A hierarchical control stack processes these multi-modal signals via an autoregressive VLA decoder that integrates egocentric views and world-centric maps to plan adaptive paths. As the vehicle approaches the delivery location, unexpected pedestrian activity prompts an agentic submodule to trigger trajectory re-planning using a reinforcement learning-inspired policy refinement routine. At the same time, AutoNav audibly warns pedestrians and recalibrates its speed to maintain safety margins. This interplay of semantic understanding, perceptual grounding, and adaptive control exemplifies the power of VLA-based systems in achieving interpretable, human-aligned behavior in safety-critical scenarios. 

This scenario illustrates how tightly integrated VLA architectures can surpass traditional perception-planning-control pipelines by enabling end-to-end semantic reasoning, rapid cross-module adaptation, and interpretable decision making. Unlike modular pipelines, where perception outputs, planner updates, and control adjustments are handled by loosely coupled components with limited semantic feedback, the VLA-based system jointly reasons over language intent, visual context, and embodied state, allowing it to dynamically replan trajectories, communicate safety-relevant intentions to humans, and adjust control policies in real time. As a result, the system exhibits greater autonomy, improved transparency through human-interpretable outputs, and enhanced decision-making agility in safety-critical environments.

In aerial robotics, VLAs enhance the capabilities of drones or UAVs in delivery and other tasks. Models such as \textbf{UAV-VLA} \cite{sautenkov2025uav} combine satellite imagery, natural language mission descriptions, and onboard sensing to execute high-level commands (e.g., “deliver to the rooftop pad with the blue tarp”). These systems use modular VLA architectures, where a vision-language planner parses global context and a flight controller executes precise waypoints, supporting applications in logistics, disaster response, and military reconnaissance.

As autonomous systems increasingly operate in unstructured environments, VLAs provide a scalable, interpretable, and data-efficient alternative to traditional pipelines. By learning from large-scale multimodal datasets and modeling decision-making as token prediction, VLAs align human-level semantics with robotic motion, paving the way for safer, smarter autonomous driving and navigation technologies.

\subsubsection{Industrial Robotics}

Industrial robotics is undergoing a paradigm shift with the integration of VLA models, enabling a new generation of intelligent robots capable of high-level reasoning, flexible task execution, and natural communication with human operators \cite{chen2025human, assres2025state}. Traditional industrial robots typically operate in highly structured environments using rigid programming, often requiring extensive reconfiguration and manual intervention when adapting to new assembly lines or product variants \cite{asif2025rapid, rodriguez2021human}. Such systems lack the semantic grounding and adaptability required for modern dynamic manufacturing settings.

VLA models, by contrast, offer a more human-interpretable and generalizable framework. Through the joint embedding of visual inputs (e.g., component layout or conveyor belt state), natural language instructions (e.g., “tighten the screw on the red module”), and robot state, VLAs can infer context and execute appropriate control commands in real-time \cite{li2025robotic, gao2025taxonomy, ma2024survey}. Vision transformers (e.g., ViT, DINOv2), LLMs (e.g., LLaMA-4), and autoregressive or diffusion-based action decoders form the backbone of these systems, allowing the robot to parse multi-modal instructions and perform actions grounded in its environment.

One of the most significant contributions in this domain is \textbf{CogACT} \cite{li2024cogact}, a modular VLA framework explicitly designed for industrial robotic manipulation. Unlike early VLAs that relied on frozen language-vision embeddings followed by direct action quantization, CogACT introduces a diffusion-based action transformer that models action sequences more robustly and adaptively. The system uses a visual-language encoder (e.g., Prismatic-7B) to extract high-level scene and instruction embeddings, which are then passed to a diffusion transformer (DiT-Base) to generate fine-grained motor actions. This modular separation enables superior generalization to unseen tools, parts, and layouts while preserving interpretability and robustness under real-world constraints.

Furthermore, CogACT demonstrates rapid adaptation across different robot embodiments such as 6-DoF arms or bimanual systems through efficient fine-tuning, making it suitable for deployment across heterogeneous factory environments \cite{li2024cogact}. Empirical evaluations show that CogACT outperforms prior models like OpenVLA by over 28\% in real-world task success rates \cite{kim2024openvla}, especially in complex, high-precision tasks such as multi-step assembly, screw fastening, and part sorting \cite{li2024cogact, yang2025efficientvla}.

As manufacturing shifts toward Industry 4.0 paradigms, VLAs promise to reduce programming overhead, support voice-commanded robot programming, and facilitate real-time human-robot collaboration on mixed-initiative tasks. While execution precision, safety guarantees, and latency optimizations remain areas of active research, the use of VLA models marks a substantial step toward autonomous, intelligent, and adaptable robots transforming the factories.

\subsubsection{Healthcare and Medical Robotics}
Healthcare and medical robotics represent a high-stakes domain where precision, safety, and adaptability are critical qualities that VLA models are increasingly well-suited to provide \cite{li2024robonurse, schmidgall2024gp}. Traditional medical robotic systems rely heavily on teleoperation or pre-programmed behaviors \cite{pantalone2021robot, si2021review}, limiting their autonomy and responsiveness in dynamic surgical or care environments. In contrast, VLA models offer a flexible framework that integrates real-time visual perception, language comprehension, and fine-grained motor control, enabling medical robots to understand high-level instructions and autonomously perform intricate procedures or assistance tasks \cite{li2024cogact, ding2024quar, verbaan2024perception}.

\begin{figure}[ht!]
\centering
\includegraphics[width=0.99\linewidth]{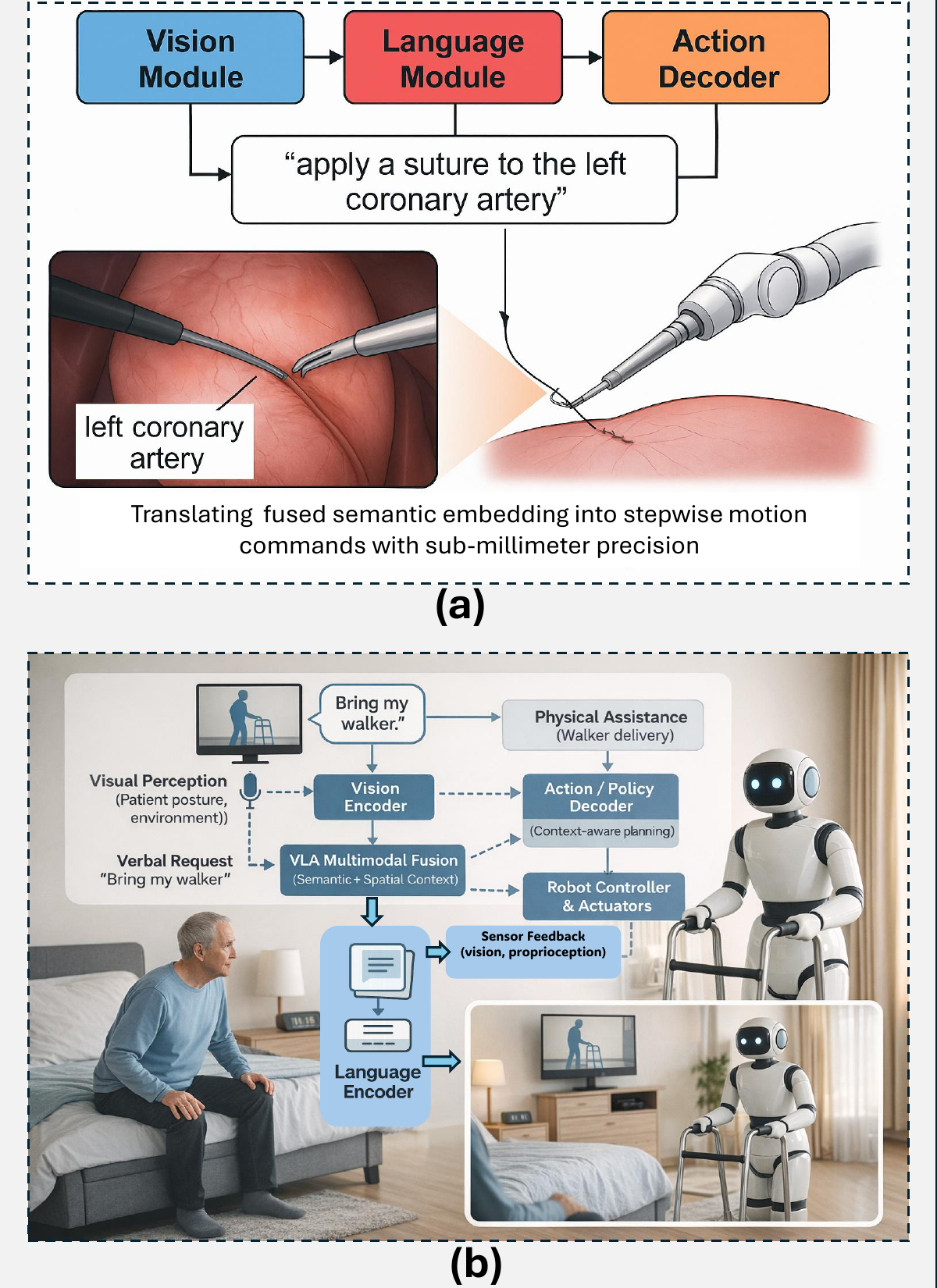}
\caption{a) This figure illustrates a  VLA  surgical system executing the task “apply a suture to the left coronary artery.” The vision module identifies anatomical targets, the language model interprets the instruction, and the action decoder generates precise motor commands, enabling adaptive tool control, real-time feedback, and safe autonomous operation; b) A VLA-powered assistive robot perceives patient behavior, processes verbal requests (e.g., “bring my walker”), and autonomously executes context-aware motion plans, enabling real-time assistance in eldercare, rehabilitation, and hospital logistics without relying on predefined scripts or manual oversight. }
\label{fig:conceptshealcare}
\end{figure}

In surgical robotics, VLAs can dramatically enhance capabilities in minimally invasive surgeries \cite{ding2025visual, wang2024non}. These systems can fuse laparoscopic video feeds \cite{li2024llava}, anatomical maps \cite{liu2025survey, ding2025visual}, and voice commands into a unified tokenized representation using vision encoders (e.g., ViT, SAM-2) and language models (e.g., LLaMA, T5) \cite{wang2025multimodal}. For instance, as depicted in Figure \ref{fig:conceptshealcare}a, in a task like “apply a suture to the left coronary artery,” the vision module identifies the anatomical target, while the language module contextualizes the instruction. The action decoder then translates the fused semantic embedding into stepwise motion commands with sub-millimeter precision. This closed-loop fusion of visual perception, language-grounded intent, and action-level control enables the robot to adaptively reposition tools, apply dynamic force feedback, and avoid critical anatomical structures, thereby reducing the need for surgeon micromanagement and minimizing the risk of human error.

Beyond the operating room, VLA models are powering a new generation of patient-assistive robots in elderly care, rehabilitation, and hospital logistics. These systems can autonomously perceive patient behavior, understand spoken or gestural input, and execute responsive tasks such as retrieving medication, guiding mobility aids, or notifying caregivers during emergencies. For example, as depicted in Figure \ref{fig:conceptshealcare}b, a VLA-enabled robot can visually detect a patient attempting to rise from bed, interpret a verbal request such as “bring my walker,” and generate a context-appropriate motion plan to assist without predefined scripts or constant supervision.

Recent VLA frameworks such as RoboNurse-VLA~\cite{li2024robonurse} highlight the real-world feasibility of this approach. RoboNurse employs SAM-2 for semantic scene segmentation and LLaMA-2 for command comprehension, integrated into a real-time voice-to-action pipeline that enables robots to assist with surgical instrument handovers in operating rooms \cite{li2024robonurse}. The system demonstrates robustness to diverse tools, varied lighting conditions, and noisy environments - common challenges in clinical settings.

Additionally, VLA architectures offer advantages in explainability and auditability, both critical in regulated medical domains \cite{trivedi2024explainable, liu2025screens}. Scene grounding and trajectory prediction can be visualized and reviewed post-hoc \cite{zhang2024vla}, which could facilitate clinical trust and enabling FDA-style validation pipelines. LoRA-based fine-tuning allows adaptation to specific hospital environments or procedural workflows with minimal data and computational infrastructure \cite{ayaz2024medvlm, wang2024surgical, liu2025survey}.

Importantly, the multi-modal foundation of VLA models enables cross-domain transferability: the same model trained on surgical tool manipulation can be adapted to patient mobility tasks with modest retraining \cite{dong2025advances}. This modularity significantly reduces development time and cost compared to task-specific automation systems \cite{hu2025vision}. As medical robotics transitions from teleoperated assistance to semi-autonomous and collaborative systems, VLA models stand at the core of this transformation.

As also discussed earier in other application areas, VLA's capability in combining high-level semantic understanding with low-level control is instrumental to provide a unified solution for scalable, human-aligned, and adaptive robotic healthcare \cite{xu2024mlevlm, zhou2025chatvla, zhang2025up}. As healthcare systems face increasing demand and workforce shortages, VLA-driven robotics will play a crucial role in enhancing medical precision, operational efficiency, and patient-centered care.

\subsubsection{Precision and Automated Agriculture}
\begin{figure*}[ht!]
\centering
\includegraphics[width=0.85\linewidth]{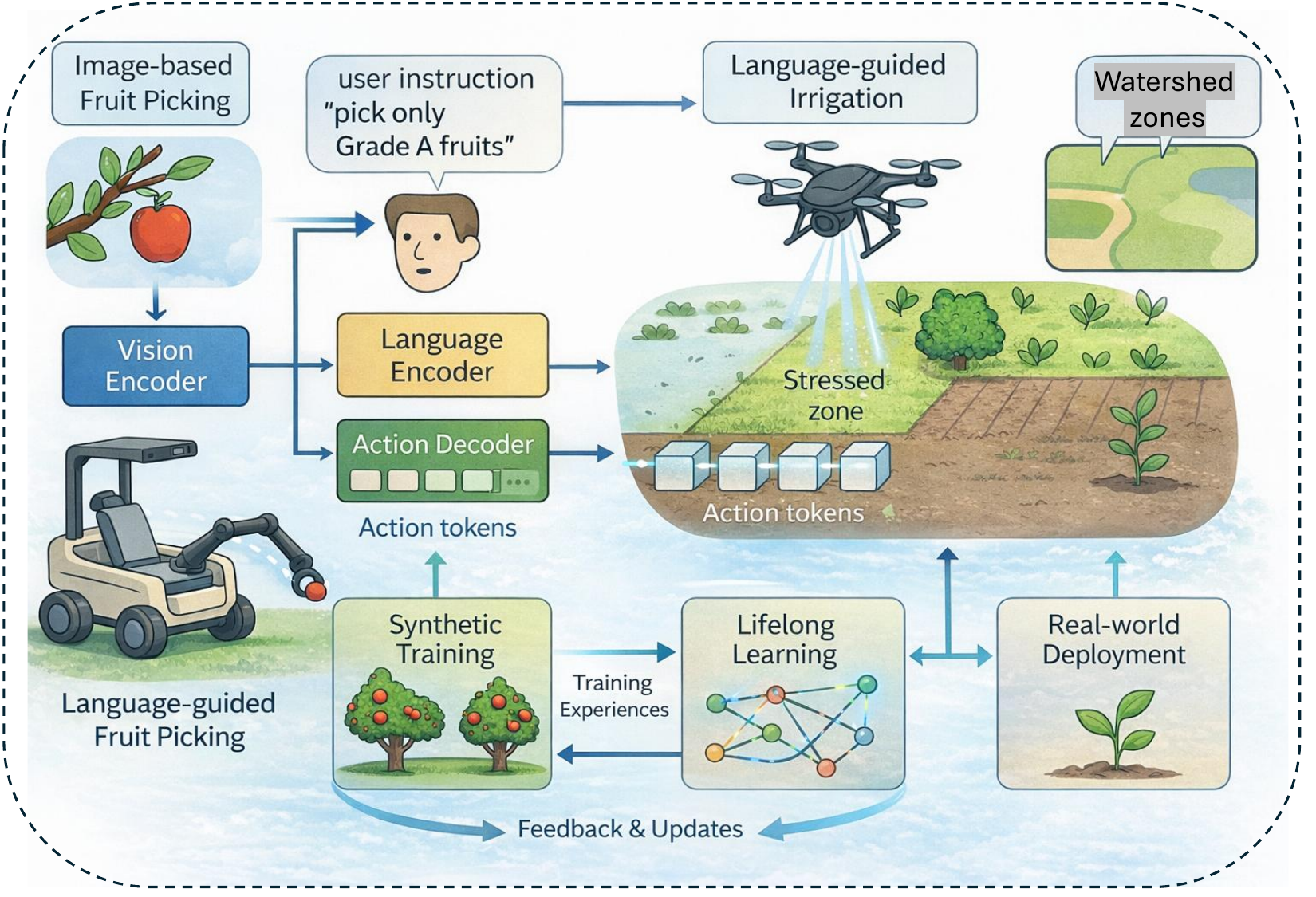}
\caption{Conceptual illustration of VLA models in precision and automated agriculture. A ground robot combines vision encoders and language instructions (e.g., “pick only Grade A fruits”) to generate action tokens for damage-free harvesting, while aerial robots use VLA reasoning for language-guided irrigation. Synthetic training, LoRA-based adaptation, and lifelong feedback enable generalization across crops, environments, and geographies, supporting sustainable, data-driven agricultural automation.}
\label{fig:conceptsagri}
\end{figure*}

As illustrated in Figure~\ref{fig:conceptsagri}, VLA models are emerging as transformative tools in precision and automated agriculture, offering intelligent, adaptive solutions for labor-intensive tasks across diverse farming landscapes \cite{gao2025vision, sautenkov2025uav}. Unlike traditional agricultural automation systems that depend on rigid, sensor-driven pipelines requiring manual reprogramming for each task or environmental variation \cite{tian2020computer, jha2019comprehensive}, VLAs integrate multi-modal perception, natural language understanding, and real-time action generation within a unified framework \cite{park2023visual, guruprasad2024benchmarking}. This unified multimodal integration enables autonomous ground robots and drones to interpret complex field scenes, follow spoken or text-based farming instructions, and generate context-aware actions such as selective fruit picking or adaptive irrigation. The ability of VLAs to dynamically adjust to occlusions, terrain irregularities, lighting variability, or varying crop types combined with training on synthetic, photorealistic datasets allows them to generalize across crop types, geographies and seasons. By leveraging action tokenization \cite{wu2024smart}, transformer-based policy generation \cite{bathula2024policy, haldar2024baku}, and techniques like LoRA fine-tuning \cite{hu2022lora}, these systems are redefining the scalability and intelligence of agricultural robotics for sustainable and precision-driven farming.

In modern fruit orchards and other crop fields, VLAs can process visual inputs from RGB-D cameras, multispectral sensors, or drones to monitor plant growth, detect diseases, and identify nutrient deficiencies. Vision transformers (e.g., ConvNeXt, DINOv2) encode spatial and semantic information from visual scenes, while LLMs (e.g., T5, LLaMA) parse natural language commands such as “inspect the east plot for powdery mildew” or “harvest ripe apples near the irrigation trench.” Through token fusion, these modalities are aligned in a shared representation space, allowing robots to execute fine-grained, context-aware actions with precision.

For instance, in fruit-picking tasks, as illustrated in Figure~\ref{fig:conceptsagri}, a VLA-equipped ground robot can identify ripe produce using image-based ripeness cues, interpret user-specified criteria such as “pick only Grade A fruits,” and execute motion sequences via action tokens that control its end-effector. This approach ensures minimal crop damage, optimizes pick rates, and allows real-time adaptation to unexpected variables like occlusions or terrain shifts. In irrigation management, drones guided by VLA models can interpret field maps and verbal instructions to selectively water stressed zones, reducing water usage \%.

Beyond immediate task execution, VLA models are expected to support dynamic reconfiguration and lifelong learning through closed-loop feedback mechanisms. In particular, execution outcomes, sensory observations, and task success signals collected during deployment can be logged and periodically incorporated into offline or incremental updates of the VLA policy. When combined with synthetic training data generated from photorealistic simulations of crop environments (e.g., 3D orchard renderings), such feedback-driven adaptation may enable models to progressively improve robustness to new crop varieties, pest conditions, and seasonal variations without extensive manual annotation. Parameter-efficient techniques such as LoRA adapters and diffusion-based policy refinement are anticipated to play a key role in facilitating this continual adaptation while limiting computational overhead.

Overall, the integration of VLA models into agricultural workflows is expected to offer several long-term benefits, including reduced reliance on skilled manual labor, improved yields through targeted intervention, and enhanced environmental sustainability via optimized input usage. As global food systems increasingly face climate variability and resource constraints, VLA-enabled agricultural technologies are anticipated to contribute to scalable, intelligent, and context-aware farming practices that better accommodate real-world complexity.

\subsubsection{Interactive AR Navigation with Vision-Language-Action Models}

Interactive Augmented Reality (AR) navigation represents a frontier where VLA models can significantly enhance human-environment interaction by providing intelligent, context-aware guidance in real-time \cite{chen2024augmented, ikeda2025marcer, xue2025reactive}. In this paradigm, VLAs process continuous streams of visual data from AR-enabled devices such as smart glasses or smartphones alongside natural language queries to generate dynamic navigational cues overlaid directly onto the user’s view of the physical world. Unlike traditional GPS-based systems that rely on rigid maps and limited user input \cite{chatzopoulos2017mobile, singh2022augmented}, VLA-based AR agents interpret complex visual scenes (e.g., intersections, indoor hallways, signage) and respond to free-form instructions such as “take me to the nearest pharmacy with a wheelchair ramp” or “show the quietest route to the conference room.”

\begin{figure}[ht!]
\centering
\includegraphics[width=0.99\linewidth]{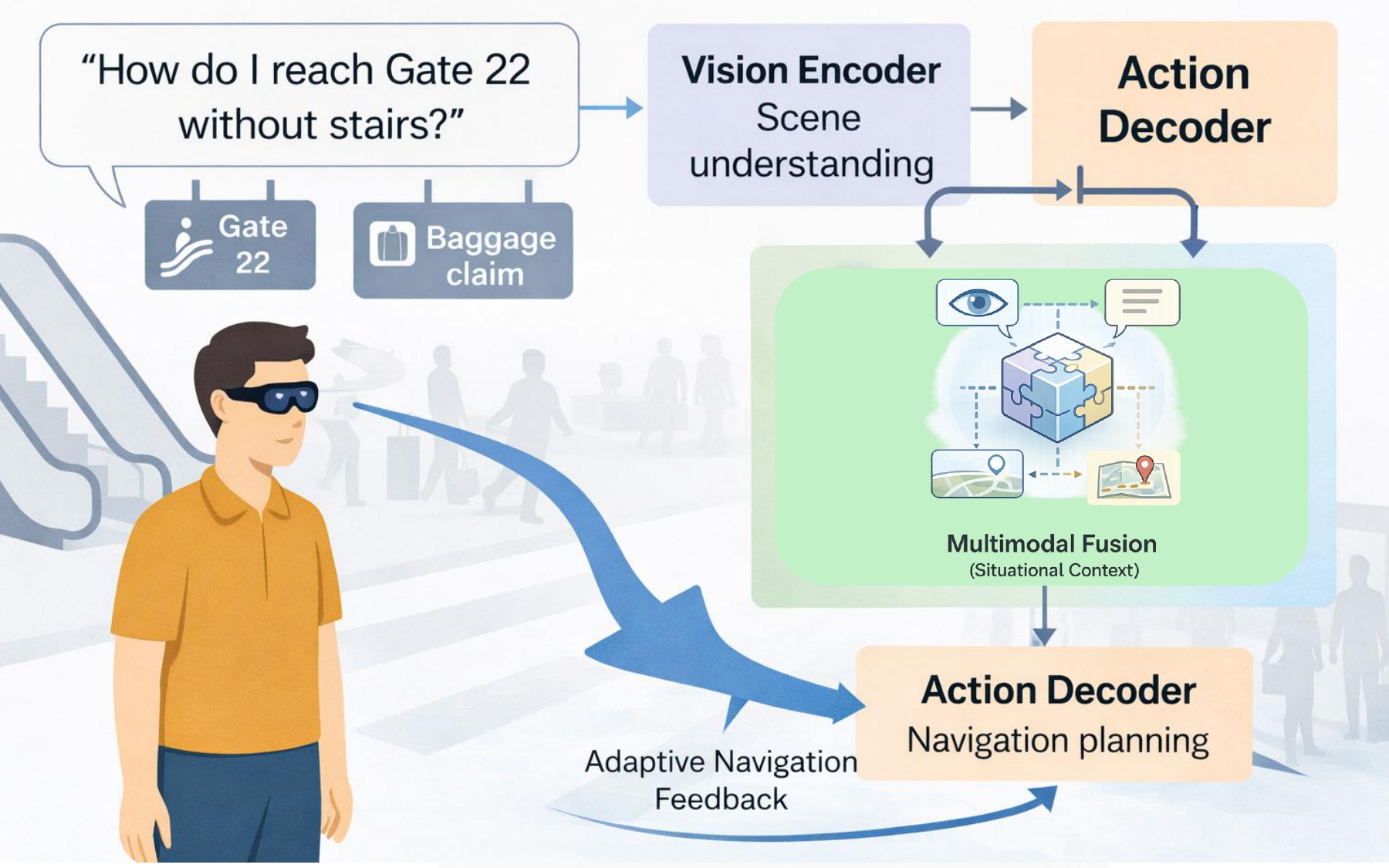}
\caption{Showing how  VLA  models enable interactive AR navigation by fusing real-time visual perception, language understanding, and action planning. In dynamic environments such as airports, VLAs interpret user queries like “avoid stairs to Gate 22,” analyze visual scenes (e.g., detecting escalators), and adjust navigational paths accordingly, supporting personalized, accessible, and context-aware mobility guidance.}
\label{fig:conceptAR}
\end{figure}

Technically, these models integrate a vision encoder (e.g., ViT, DINOv2) that extracts scene representations from RGB camera frames, a language encoder (e.g., T5 or LLaMA) that processes user prompts or voice commands, and an action decoder that predicts tokenized navigation cues such as directional overlays, waypoints, or voice instructions. A transformer-based architecture fuses these modalities to reason about both the spatial layout and semantic intent, allowing the AR agent to adaptively highlight paths, landmarks, and hazards directly within the user's field of view \cite{sun2025review, pang2025towards}. For example, as shown in Figure \ref{fig:conceptAR}, in a crowded airport, the VLA agent could visually identify escalators, gates, or baggage claims while understanding a query like “how do I reach Gate 22 without stairs?”, adjusting the route in response to real-time occupancy and obstacles.

VLAs are expected to support interactive instruction loops in which users can first issue high-level commands (e.g., ``navigate to the pharmacy'') and subsequently refine them with additional constraints such as ``avoid busy areas'' or ``take the scenic route.'' Through context-aware feedback and iterative clarification, such interaction paradigms can improve accessibility and usability for visually impaired or cognitively challenged individuals. In logistics and indoor navigation, these systems can be integrated with IoT sensors and digital twins to guide warehouse workers, maintenance teams, or delivery robots through complex environments. Furthermore, personalized navigation can be achieved through continual fine-tuning, where VLA models learn user preferences and local spatial layouts over time.

As AR hardware becomes more affordable and integrated into daily life, VLA-powered navigation systems will enable seamless spatial understanding, multi-modal interaction, and autonomous guidance in public, industrial, and assistive contexts redefining how humans perceive, explore, and interact with physical spaces.

\section{Challenges and Limitations of Vision-Language-Action Models}
VLA models face a spectrum of interrelated challenges that impede their translation from research prototypes to robust, real‐world systems. First, achieving real‐time, resource‐aware inference remains difficult: models like DeeR-VLA leverage dynamic early‐exit architectures to cut computation 5–6× on manipulation benchmarks while preserving accuracy, yet their gains diminish in complex scenarios \cite{yue2024deer}. Similarly, Uni-NaVid compresses egocentric video tokens for 5 Hz navigation but still struggles under highly ambiguous instructions and longer horizons \cite{zhang2024uni}. Moreover, these efficiency-driven designs often expose a trade-off between computational speed and representational coverage. When operating under aggressive compression or early-exit constraints, even advanced hybrid vision--language grounding methods exhibit limited object generalization; for example, ObjectVLA generalizes to only 64\,\% of novel objects, highlighting how real-time optimization can exacerbate gaps in open-world robustness \cite{zhu2025objectvla}.

Second, adapting VLA models with minimal supervision and ensuring stable policy updates under scarce, noisy data is nontrivial. ConRFT combines behavior cloning and Q-learning with human‐in‐the‐loop fine‐tuning to rapidly converge to 96.3\% success over eight contact‐rich tasks, yet it relies heavily on expert interventions and reward shaping \cite{chen2025conrft}. Hierarchical frameworks such as Hi Robot decouple high‐level reasoning from low‐level execution to improve instruction fidelity, but coordinating these modules and grounding ambiguous feedback remains challenging \cite{shi2025hi}. Likewise, Tactile-language-action model model's fusion of tactile streams with language commands achieves over 85 \% success on unseen peg‐in-hole tasks, but dataset breadth and real‐time multi‐step decoding still limit broader generalization \cite{hao2025tla}.

Furthermore, ensuring safety, generalization, and end‐to‐end reliability in dynamic environments demands new modeling and evaluation standards. Occupancy‐Language‐Action models like OccLLaMA unify 3D scene understanding with action planning, yet they need to scale to richer scene dynamics and semantic consistency across modalities \cite{wei2024occllama}. RaceVLA advances high-speed drone navigation through quantized, iterative control loops; however, its limited visual--physical generalization compared to larger VLAs and dedicated reasoning models raises safety concerns in unseen or rapidly changing environments \cite{serpiva2025racevla}. Model‐merging strategies in ReVLA recover lost out‐of‐domain visual robustness improving oriented object detection (OOD) grasp success by up to 77 \% but introduce extra computation and complexity \cite{dey2024revla}. Finally, SafeVLA formulates constraints via constrained Markov decision processes to cut unsafe behavior by over 80 \%, yet defining comprehensive, non‐restrictive safety rules for diverse real‐world tasks remains an open problem \cite{zhang2025safevla}. Addressing these intersecting limitations is critical for VLA models to achieve reliable, autonomous operation under the full complexity of real‐world robotics.

Building upon the critical limitations outlined above, it is imperative to map each challenge to targeted mitigation strategies and assess their system-level impact. Table \ref{tab:vla_challenges_extended} presents this map identifying core limitations, potential technical remedies drawn from recent advances, and articulating the anticipated benefits for real-world VLA deployment. For instance, addressing real-time inference constraints leverages parallel decoding and quantized transformer pipelines with hardware acceleration (e.g., TensorRT) to sustain control loop speed in drones and manipulators \cite{li2024improving, kim2024openvla, geens2024bringing, liu2025hybridvla}. Addressing multi-modal action representation via hybrid diffusion–autoregressive policies enriches a model’s capacity to produce varied, context-sensitive motor commands for complex tasks \cite{pertsch2025fast, ma2024survey}. To guarantee safety in open worlds, dynamic risk assessment modules and adaptive planning layers can be integrated, ensuring robust emergency stop behaviors in unpredictable settings \cite{rodriguez2025integrating, wang2024exploring, jiang2025behavior}. Similarly, dataset bias and grounding can be minimized through curated debiased corpora and advanced contrastive fine-tuning, strengthening fairness and semantic fidelity when generalizing to novel objects and scenes \cite{sahili2025scaling, bordes2024introduction, qu2025spatialvla}. Together, these strategies and other approaches spanning simulation-to-real transfer, tactile integration, and energy-efficient architectures frame a comprehensive roadmap for transitioning VLA research into reliable, scalable autonomy.

\begin{table*}[t]
\centering
\scriptsize
\setlength{\tabcolsep}{3.5pt}
\renewcommand{\arraystretch}{1.15}
\caption{Challenges, potential solutions, and expected impact of VLA models.}
\label{tab:vla_challenges_extended}
\begin{tabular}{p{3.0cm} p{7.0cm} p{6.8cm}}
\toprule
\textbf{Challenge / limitation} & \textbf{Potential solution} & \textbf{Expected impact} \\
\midrule
Real-time inference constraints &
Parallel decoding, quantized transformers, and hardware acceleration (e.g., TensorRT) \cite{li2024improving, kim2024openvla}; reduce autoregressive overhead \cite{geens2024bringing, liu2025hybridvla}. &
Enables real-time control and deployment in latency-critical domains \cite{xu2025embodied, sautenkov2025uav} (e.g., UAVs, manipulators). \\
\addlinespace

Multi-modal action representation&
Hybrid tokenization combining diffusion and autoregressive policies \cite{pertsch2025fast}; train on diverse demonstrations and multi-modal outputs \cite{ma2024survey}.&
Improves performance on complex, dynamic manipulation with multiple valid solution modes \cite{gbagbe2024bi}. \\
\addlinespace

Safety assurance in open worlds &
Dynamic risk assessment modules \cite{rodriguez2025integrating, wang2024exploring}; low-latency emergency-stop and adaptive planning layers \cite{jiang2025behavior}. &
Improves reliability and safety in unpredictable settings (homes, factories, healthcare); enhances user acceptability.\\
\addlinespace

Dataset bias and grounding &
Curate diverse/debiased datasets \cite{sahili2025scaling}; stronger grounding (e.g., CLIP fine-tuning with hard negatives) \cite{zhang2025generative, bordes2024introduction}. &
Improves fairness and semantic fidelity \cite{jiang2024solami}, and enhances generalization to novel real-world inputs \cite{vuong2023open, zhen20243d, qu2025spatialvla}. \\
\addlinespace

Limited 3D perception and reasoning &
Integrate depth/LiDAR; develop 3D-aware architectures; fuse point clouds with vision--language features. &
Enables stronger spatial reasoning for manipulation and navigation in complex environments \cite{li2024improving}. \\
\addlinespace

Cross-embodiment generalization &
Train across diverse morphologies; learn embodiment-agnostic action abstractions; apply cross-domain adaptation \cite{ye2024latent}. &
Facilitates policy transfer across robot platforms and configurations \cite{zhang2025up, kim2024openvla}. \\
\addlinespace

Annotation complexity and cost &
Weak supervision, active learning, and synthetic data generation to reduce manual labeling \cite{liu2024synthvlm}. &
Lowers development cost and accelerates scaling to new tasks/domains \cite{wang2024exploring, zhao2025cot}. \\
\addlinespace

Sim-to-real transfer gap &
Domain adaptation, sim-to-real fine-tuning, and real-world calibration \cite{sun2025prism, li2025hamster}. &
Improves reliability and consistency when deploying beyond simulation \cite{anderson2021sim, fang2025rebot}. \\
\addlinespace

Integration of physical knowledge &
Physics priors, simulation environments, and dynamics modeling in training pipelines \cite{ding2025humanoid}. &
Improves prediction and planning under real physical constraints \cite{intelligence2025pi}. \\
\addlinespace

Multi-modal integration (tactile, audio) &
Fuse tactile/audio with vision and language \cite{jones2025beyond}; extend multimodal transformer fusion. &
Improves robustness under occlusion/ambiguity and expands the task repertoire \cite{ghosh2024exploring, li2025visual, hong2025building}. \\
\addlinespace

Long-horizon multi-stage tasks &
Hierarchical policies, memory-augmented networks, and trajectory planning modules \cite{li2025visual}. &
Improves sequential planning, memory, and compositional execution \cite{li2024cogact, zhao2025cot, vuong2023open, qu2025spatialvla}. \\
\addlinespace

System integration complexity &
Unified transformer backbones \cite{zheng2025universal}; temporal alignment and sim-to-real transfer strategies \cite{noorani2025abstraction, zhang2025slim}. &
Enables tighter planning--control coordination and more robust transfer to physical robots \cite{samson2025scalable, song2025survey}. \\
\addlinespace

Energy and compute demands &
Pruning, LoRA, quantization-aware training, and low-power accelerators. &
Enables efficient embedded/mobile deployment \cite{xu2025vla, zhang2025mole, li2025pointvla, wu2025momanipvla}. \\
\addlinespace

Generalization to unseen tasks &
Compositional generalization, few-shot meta-learning, and task-agnostic pretraining \cite{polubarov2025vintix, liu2025screens}. &
Reduces overfitting and strengthens zero-/few-shot adaptation \cite{huang2025otter, wen2025tinyvla, zhao2025cot}. \\
\addlinespace

Robustness to environmental variability &
Domain randomization, sensor fusion, broader training datasets, and online recalibration \cite{ma2024survey}.&
Improves stability under changing lighting, clutter, and scene dynamics \cite{zhao2025more, wen2025tinyvla}. \\
\addlinespace

Ethical and societal implications &
Privacy via on-device processing/anonymization \cite{lu2025probing, sharshar2025vision, xu2025vla, chen2025combatvla}; fairness audits; regulatory and trust frameworks. &
Promotes equitable and trustworthy adoption across social, medical, and labor domains \cite{mumuni2025large, raza2025responsible, team2025gemini, plaat2025agentic}. \\
\bottomrule
\end{tabular}
\end{table*}

The remainder of this section is organized into five focused subsections, each examining a distinct cluster of VLA challenges identified in the literature. First, we analyze real-time inference constraints and the emerging methods to address them. Next, we explore multi-modal action representation alongside safety assurance in open-world settings. We then discuss dataset bias, grounding strategies, and generalization to unseen tasks, followed by an exploration of system integration complexity and computational demands. Finally, we consider robustness and the ethical implications of deploying VLAs in real-world applications.

\subsection{Real-Time Inference Constraints} 
Despite recent progress, deploying VLA models in latency-critical settings continues to be constrained by real-time inference requirements, particularly in applications such as robotic manipulation, autonomous driving, and aerial control. VLAs typically depend on autoregressive decoding strategies, which sequentially generate action tokens based on previous predictions. While effective for many tasks, this autoregressive decoding paradigm substantially limits inference speed, typically achieving only 3-5\,Hz when deployed on standard GPU-based research platforms (e.g., single high-end consumer or datacenter GPUs) for end-to-end VLA inference \cite{firoozi2023foundation}. This rate remains substantially below the control frequencies typically required for responsive and stable robotic operation, which often range from tens of hertz for high-level planning to higher update rates for low-level feedback control, depending on the task and hardware platform. For instance, when a robotic arm manipulates delicate objects, frequent positional updates are essential to maintain accuracy and prevent damage. Models such as OpenVLA \cite{kim2024openvla} and Pi-0 \cite{black2024pi_0} face inherent challenges with this sequential token generation approach, thereby limiting their effectiveness in dynamic environments.

Emerging solutions such as parallel decoding, exemplified by NVIDIA’s GR00T N1 model \cite{bjorck2025gr00t}, aim to accelerate inference by predicting multiple tokens simultaneously. GR00T N1 achieves approximately a 2.52× speedup over traditional decoding methods; however, this parallelism often introduces trade-offs in trajectory smoothness, resulting in suboptimal robot movements. Such movements are undesirable in sensitive applications like surgical robotics, where precision, adaptability and flexibility are paramount. Thus, achieving rapid inference without compromising output quality remains an open challenge.

Additionally, hardware limitations exacerbate real-time inference constraints. For example, processing high-dimensional visual embeddings, typically involving over 400 vision tokens at 512 dimensions each, requires approximately 1.2 GB/s memory bandwidth. This demand significantly exceeds the capacity of current embedded systems or edge-AI hardware such as NVIDIA Jetson platforms, thereby restricting practical deployment \cite{guan2025efficient, zhang2025pure}. Even with efficient quantization techniques, which reduce the precision of floating-point operations to alleviate memory constraints, models frequently experience accuracy degradation, especially in tasks demanding sub-millimeter precision, such as bimanual robotic manipulation or medical robotics.

\subsection{Multi-modal Action Representation and Safety Assurance}

\textbf{Multi-modal Action Representation:} One significant limitation of current  VLA  models is accurately representing multi-modal actions, particularly in scenarios requiring continuous and nuanced control \cite{fan2025interleavevlaenhancingrobotmanipulation, deng2025graspvlagraspingfoundationmodel}. Traditional discrete tokenization methods, such as those dividing actions into 256 distinct bins, inherently lack precision, creating substantial errors in fine-grained tasks like delicate robotic grasping or intricate surgical procedures \cite{pertsch2025fast}. For instance, during precise robotic manipulation in assembly tasks, discrete representations can result in misaligned or imprecise actions, undermining performance and reliability. On the other hand, continuous multilayer perceptron (MLP) based approaches face the risk of mode collapse \cite{nie2021mlp, wang2025roboflamingo}, where models converge prematurely to single action trajectories, despite multiple viable paths available. This diminishes the flexibility necessary for adaptive decision-making in highly dynamic environments. Emerging diffusion-based policies, exemplified by models like Pi-Zero and RDT-1B \cite{liu2024rdt}, offer richer multi-modal action representation capable of capturing diverse action possibilities. However, their substantial computational overhead, approximately three times that of conventional transformer-based decoders, renders them impractical for real-time deployment. Consequently, VLA models currently struggle with complex dynamic tasks, such as robotic navigation in densely crowded spaces or sophisticated bimanual manipulations \cite{gbagbe2024bi, xiang2025vla}, where multiple strategic actions may be equally valid and contextually dependent.

\textbf{Safety Assurance in Open World:} Another critical challenge facing VLAs is ensuring robust safety in dynamic, unpredictable environments characteristic of real-world scenarios \cite{cheng2024manipulation, zhang2025safevla}. Many current implementations depend heavily on predefined, hardcoded force and torque thresholds, significantly constraining their adaptability in encountering unforeseen or novel conditions, such as unexpected obstacles or sudden environmental changes \cite{ma2024survey}. Models used for collision prediction typically attain only about 82\% accuracy in cluttered and dynamic spaces, posing serious risks in applications such as warehouse logistics or household robotics, where safety margins are minimal \cite{zhen20243d, kim2024openvla}. Moreover, the essential safety mechanisms like emergency stops incorporate substantial latency often between 200 and 500 milliseconds due to comprehensive safety verifications \cite{patel2023pretrained, kim2024openvla}. This delay, although seemingly minor, can prove hazardous in high-speed operations or critical interventions, such as automated driving or emergency robotic responses. 

\subsection{Dataset Bias, Grounding, and Generalization to Unseen Tasks}
A significant obstacle limiting the effectiveness of  VLA  models is the pervasive presence of dataset bias and grounding deficiencies. Current training datasets, predominantly sourced from web-crawled repositories, frequently exhibit inherent biases \cite{szot2024grounding, kelly2024visiongpt}. Studies indicate that approximately 17\% of the associations within standard datasets are skewed toward stereotypical interpretations, such as disproportionately associating terms like “doctor” with male figures \cite{torres2024comprehensive, lee2023survey}. These biases propagate through training, resulting in VLAs that produce semantically misaligned or contextually inappropriate responses when deployed in diverse environments. For instance, models such as OpenVLA have been documented to overlook approximately 23\% of object references in novel settings, significantly limiting their practical utility in real-world applications where accurate interpretation of instructions is critical \cite{kim2024openvla}. This grounding issue also extends to challenges in compositional generalization, where VLAs often fail when encountering rare or unconventional combinations, such as interpreting a phrase like “yellow horse” because of underrepresentation in training corpora. These shortcomings highlight an urgent need for carefully curated, balanced, and comprehensive, domain-specific datasets, coupled with advanced grounding algorithms designed to mitigate biases and enhance semantic alignment across varied contexts.

Complementing the challenges posed by dataset bias is the broader issue of generalization to unseen tasks, a critical barrier for the practical deployment of VLAs. While existing models demonstrate proficiency in familiar environments or tasks similar to their training scenarios, their performance significantly degrades, often by as much as 40\%, when encountering entirely novel tasks or unfamiliar variations. For example, a VLA trained specifically on domestic tasks may struggle or fail when introduced into industrial or agricultural settings, largely due to discrepancies in object types, environmental dynamics, and operational constraints. This limitation arises primarily from overfitting to narrowly scoped training distributions and insufficient exposure to diverse task representations. Consequently, current VLAs exhibit limited generalization in zero-shot or few-shot learning scenarios, impeding their adaptability and scalability.

\subsection{System Integration Complexity and Computational Demands}
Integrating  VLA  models within dual-system architectures, which combine high-level cognitive planning (System 2) and real-time physical control (System 1), presents significant complexity in robotic applications. A primary challenge arises from temporal mismatches between these two systems. Typically, System 2 leverages LLMs such as GPT or LLaMA-4 for complex task decomposition and strategic planning. These models, due to their substantial computational requirements, often incur inference latencies on the order of $\sim$800\,ms or more when executed on standard GPU-based inference platforms (e.g., single high-end consumer or datacenter GPUs) commonly used for LLM deployment. Conversely, System~1 components responsible for low-level motor execution typically run within tightly constrained control loops implemented on real-time CPUs, microcontrollers, or dedicated robot controllers, where update intervals on the order of several milliseconds are common, depending on the platform and task. This stark discrepancy in operational cadence leads to synchronization difficulties, causing delays and potentially suboptimal execution trajectories. For example, NVIDIA's GR00T N1 model demonstrates an effective integration of these two systems but still suffers from occasional jerkiness in motion due to asynchronous interaction, highlighting this intrinsic challenge.

Furthermore, the feature space misalignment between high-dimensional vision encoders, such as Vision Transformers (ViT), and lower-dimensional action decoders exacerbates integration complexity. When attempting to reconcile these disparate embeddings, the coherence between perceptual understanding and actionable commands can deteriorate significantly. OpenVLA \cite{kim2024openvla} and RoboMamba \cite{liu2024robomamba}, which utilize transformer-based visual processing and subsequent action decoding, illustrate these integration challenges resulting in diminished performance when ported from simulation environments to physical hardware deployments. Such discrepancies may lead to  reduction in performance, primarily due to mismatches between simulated dynamics and real-world sensor noise or calibration issues\cite{kim2024openvla, din2025vision, hou2025dita}.

Energy and compute demands constitute another significant barrier for VLA deployment, particularly in edge computing contexts typical of autonomous drones, mobile robots, and wearable robotic systems. The substantial parameter counts typical of advanced VLAs (e.g.,models possessing upwards of 7 billion parameters) necessitate computational resources often exceeding 28 GB of VRAM in their native form. These requirements are substantially higher than the capabilities of most current edge-oriented processors and GPUs, restricting the practical applicability of sophisticated VLAs outside specialized, high-resource environments.
    
\subsection{Robustness and Ethical Challenges in VLA Deployment}

A central barrier to the real-world deployment of  VLA models lies in their limited robustness to environmental variability, which in turn raises important ethical and safety considerations. Environmental robustness refers to a system’s ability to sustain reliable perception, reasoning, and action generation under dynamically changing and partially observable conditions. In practice, real-world environments introduce significant uncertainty through factors such as fluctuating illumination, adverse weather, sensor noise, and object occlusions.

Empirical evidence highlights these limitations across multiple VLA components. For example, vision modules employed in systems such as OpenDriveVLA~\cite{zhou2025opendrivevla} experience accuracy degradations of approximately 20--30\% in low-contrast or shadow-dominated scenes, reflecting the sensitivity of current visual encoders to challenging lighting conditions. Similarly, language understanding in VLAs such as CoVLA~\cite{arai2025covla} deteriorates in acoustically noisy or semantically ambiguous settings, where instruction misinterpretation can propagate into incorrect action execution. In manipulation-centric scenarios, VLA-enabled robotic systems like RoboMamba~\cite{liu2024robomamba} struggle in cluttered environments, frequently misestimating the pose or orientation of partially occluded objects and thereby reducing task success rates.

These robustness limitations have direct ethical implications in safety-critical deployments, as performance degradation under real-world variability can lead to unintended behaviors, reduced reliability, and loss of user trust. Addressing robustness is therefore not only a technical challenge but also a prerequisite for responsible and ethical deployment of VLA systems in human-centered environments.

\begin{figure*}[ht!]
\centering
\includegraphics[width=0.75\linewidth]{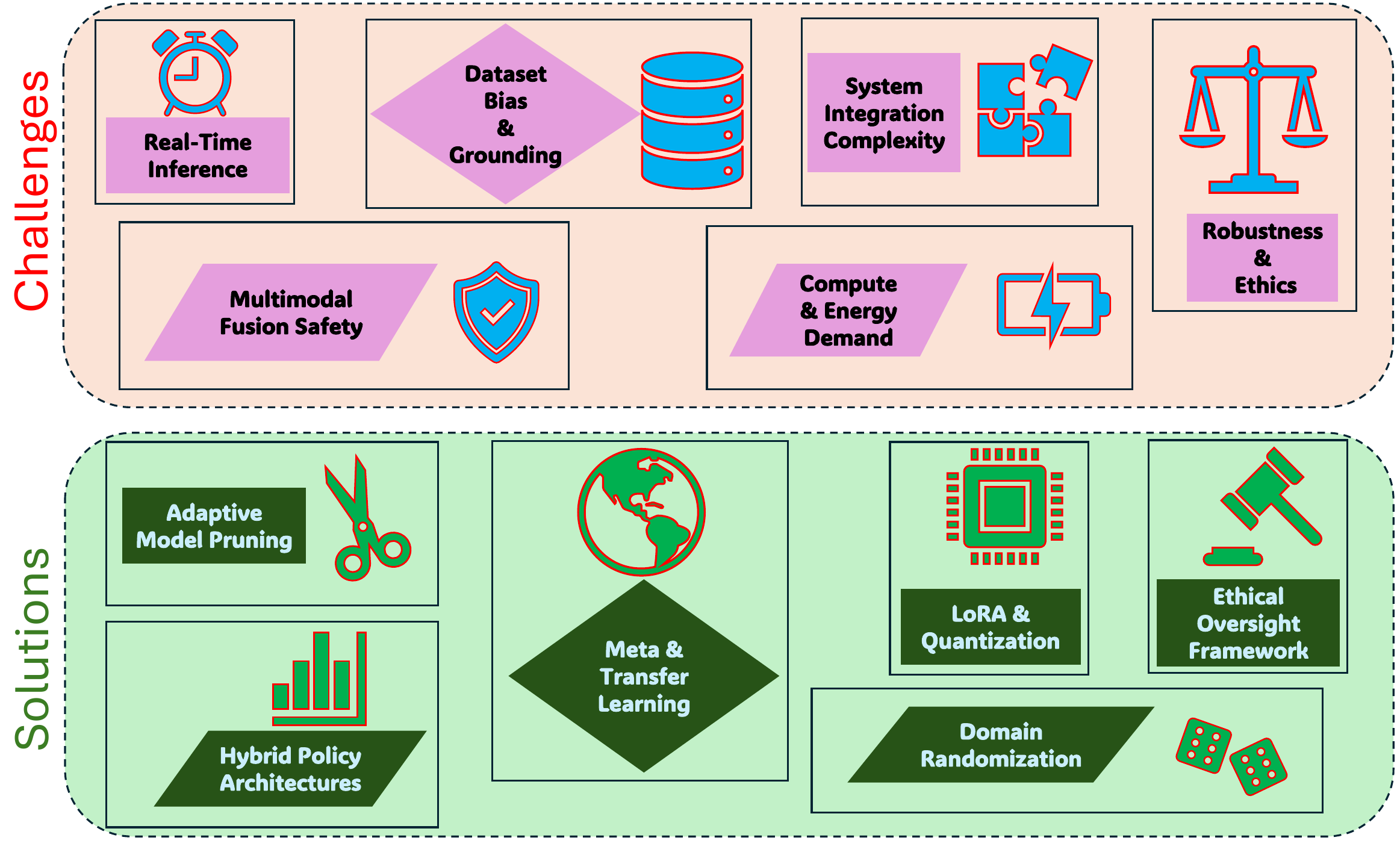}
\caption{Figure maps six core VLA challenges; namely real-time inference, multimodal fusion safety, dataset bias, integration complexity, compute demands, and robustness/ethics; against six targeted solutions: adaptive pruning, hybrid policy architectures, meta/transfer learning, LoRA/quantization, domain randomization, and ethical oversight. This systematic alignment clarifies pathways to robust, efficient, and safe VLA deployment across broader real-world robotic domains.}
\label{fig:challengesoln}
\end{figure*}

\section{Discussion}
As illustrated in Figure \ref{fig:challengesoln}, VLA models face a multifaceted set of challenges that span algorithmic, computational, and ethical dimensions. First, achieving real-time inference on resource-constrained hardware remains difficult due to the sequential nature of autoregressive decoders and the high dimensionality of multi-modal inputs. Second, fusing vision, language, and action into coherent policies introduces safety vulnerabilities when encountering unanticipated environmental changes. Third, dataset bias and grounding errors compromise generalization, often causing models to fail on out-of-distribution tasks. Fourth, integrating diverse components perception, reasoning, and control yields complex architectures that are hard to optimize and maintain. Fifth, the energy and compute demands of large VLA systems hinder the deployment on embedded or mobile platforms.  

Finally, limited robustness to environmental variability can result in unsafe or unreliable behavior, which in turn raises ethical and regulatory concerns related to safety assurance, accountability, privacy, and bias mitigation. Collectively, these limitations constrain the practical adoption of VLA models in real-world robotics, autonomous systems, and interactive applications. The potential solutions to these challenges are discussed below.
\begin{figure}[ht!]
\centering
\includegraphics[width=0.95\linewidth]{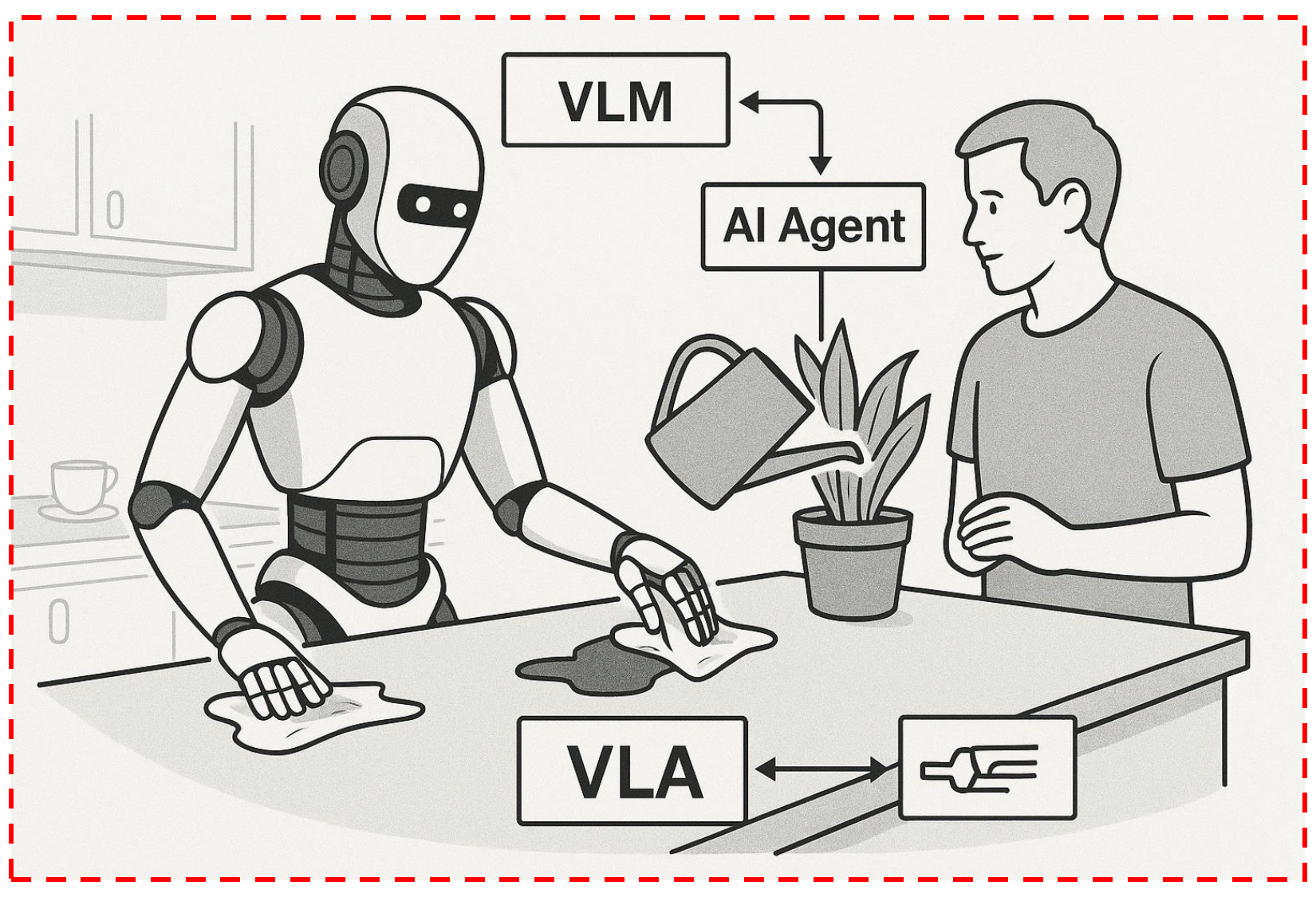}
\caption{This conceptual illustration presents “Eva,” a future humanoid assistant powered by Vision-Language Models (VLMs),  VLA  frameworks, and agentic AI systems. VLMs enable semantic scene understanding and object affordance prediction, while VLAs translate language-grounded instructions into hierarchical motor plans. Agentic AI modules ensure adaptive learning, self-refinement, and interactive decision-making in open-ended environments. Together, these components represent a foundational blueprint for Artificial General Intelligence (AGI) in robotics, where perception, language understanding, planning, and safe autonomous behavior converge in real-world, socially aware tasks.}
\label{fig:futuree}
\end{figure}

\subsection{Potential Solutions}
\begin{itemize}

  \item \textbf{Real-Time Inference Constraints.}  
  Future research must develop VLA architectures that harmonize latency, throughput, and task-specific accuracy. One promising direction is the integration of specialized hardware accelerators such as FPGA-based vision processors and tensor cores optimized for sparse matrix operations to execute convolutional and transformer layers at sub-millisecond scales \cite{kim2024openvla, li2024improving}. Model compression techniques like Low-Rank Adaptation (LoRA) \cite{hu2022lora} and knowledge distillation can shrink parameter counts by up to 90\%, reducing both memory footprint and inference time while retaining over 95\% of original performance on benchmark tasks. Progressive quantization strategies that combine mixed-precision arithmetic (e.g., FP16/INT8) with block-wise calibration can further cut computation by 2–4× with minimal accuracy loss \cite{kim2025fine}. Adaptive inference architectures that dynamically adjust network depth or width based on input complexity akin to early-exit branches in DeeR-VLA \cite{yue2024deer} can reduce average compute by selectively bypassing transformer layers when visual scenes or linguistic commands are simple. Finally, efficient tokenization schemes leveraging subword patch embeddings and dynamic vocabulary allocation can compress visual and linguistic input into compact representations, minimizing token counts without sacrificing semantic richness \cite{pertsch2025fast}. Together, these innovations can enable sub-50 ms end-to-end inference on commodity edge GPUs, paving the way for latency-sensitive applications in autonomous drone flight, real-time teleoperation, and collaborative manufacturing.

  \item \textbf{Multi-modal Action Representation and Safety Assurance.}  
  Addressing multi-modal action representation and robust safety requires end-to-end frameworks that unify perception, reasoning, and control under stringent safety constraints. Hybrid policy architectures combining diffusion-based sampling for low-level motion primitives \cite{chi2023diffusion} with autoregressive high-level planners \cite{wen2025tinyvla} enable compact stochastic representations of diverse action trajectories, improving adaptability in dynamic environments. Safety can be enforced via real-time risk assessment modules that ingest multi-sensor fusion streams including visual, depth, and proprioceptive data to predict collision probability and joint stress thresholds, triggering emergency stop circuits when predefined safety envelopes are breached \cite{rodriguez2025integrating, wang2024exploring}. Reinforcement learning algorithms augmented with constrained optimization (e.g., Lagrangian methods in SafeVLA \cite{zhang2025safevla}) can learn policies that maximize task success while strictly respecting safety constraints. Online model adaptation techniques such as rule-based RL (GRPO) and Direct Preference Optimization (DPO) further refine action selection under new environmental conditions, ensuring consistent safety performance across scenarios \cite{jiang2025behavior}. Crucially, embedding formal verification layers that symbolically analyze planner outputs before execution can guarantee compliance with safety invariants, even for neural-network–based controllers. Integrating these methodologies will produce VLA systems that not only execute complex, multi-modal actions but do so with provable safety in unstructured, real-world settings.

  \item \textbf{Dataset Bias, Grounding, and Generalization to Unseen Tasks.}  
  Robust generalization demands both broadened data diversity and advanced learning paradigms. Curating large-scale, debiased multi-modal datasets combining web-scale image–text corpora like LAION-5B \cite{schuhmann2022laion} with robot-centric trajectory archives such as Open X-Embodiment \cite{vuong2023open} lays the groundwork for equitable semantic grounding. Hard-negative sampling and contrastive fine-tuning of vision–language backbones (e.g., CLIP variants) can mitigate spurious correlations and enhance semantic fidelity \cite{bordes2024introduction, zhang2025generative}. Meta-learning frameworks enable rapid adaptation to novel tasks by learning shared priors across task families, as demonstrated in vision-language robotic navigation models \cite{qu2025spatialvla}. Continual learning algorithms with replay buffers and regularization strategies preserve old knowledge while integrating new concepts, addressing catastrophic forgetting in VLA models \cite{dey2024revla}. Transfer learning from 3D perception domains (e.g., point cloud reasoning in 3D-VLA \cite{zhen20243d}) can endow models with stronger spatial inductive biases, thereby improving robustness to out-of-distribution scenarios. Finally, simulation-to-real (sim2real) fine-tuning with domain randomization and real-world calibration such as dynamic lighting, texture, and physics variations ensures that policies learned in synthetic environments transfer effectively to physical robots \cite{anderson2021sim, fang2025rebot}. These combined strategies will empower VLAs to generalize confidently to unseen objects, scenes, and tasks in real-world deployments.

  \item \textbf{System Integration Complexity and Computational Demands.}  
  To manage the intricate orchestration of multi-modal pipelines under tight compute budgets, researchers must embrace model modularization and hardware–software co–design. Low-Rank Adaptation (LoRA) adapters can be injected into pre–trained transformer layers, enabling task-specific fine–tuning without modifying core weights \cite{hu2022lora}. Knowledge distillation from large ``teacher'' VLAs to lightweight ``student'' networks, guided by mutual information–based objectives that encourage the student to match the teacher’s intermediate representations and action distributions, produces compact models with 5--10$\times$ fewer parameters while retaining 90--95\% of task performance \cite{kim2025fine}. Mixed-precision quantization augmented by quantization-aware training can compress weights to 4–8 bits, cutting memory bandwidth and energy consumption by over 60\% \cite{kim2024openvla}. Hardware accelerators tailored for VLA workloads supporting sparse tensor operations, dynamic token routing, and fused vision–language kernels can deliver sustained 100+ TOPS throughput within a 20–30 W power envelope, meeting the demands of embedded robotic platforms \cite{pertsch2025fast, wen2025tinyvla}. Toolchains like TensorRT-LLM \cite{li2024improving} and TVM can optimize end-to-end VLA graphs for specific edge devices, fusing layers and precomputing static subgraphs. Emerging architectures such as TinyVLA demonstrate that sub–1 B parameter VLAs can achieve near–state-of-the-art performance on manipulation benchmarks with real–time inference, charting a path for widespread deployment in resource-constrained settings.

  \item \textbf{Robustness to Environmental Variability in VLA Deployment.}  
  Ensuring robust VLA performance in real-world settings requires targeted technical interventions to handle environmental uncertainty and long-term system drift. Domain randomization and synthetic data augmentation pipelines, such as UniSim’s closed-loop sensor simulator, generate photorealistic variations in lighting, occlusion, and sensor noise, thereby improving resilience to distributional shifts \cite{yang2023unisim}. In addition, adaptive recalibration modules that dynamically adjust perception thresholds and control gains based on real-time feedback can mitigate performance degradation caused by sensor aging or changing operating conditions. Together, these approaches aim to enhance the stability and reliability of VLA systems under diverse and evolving deployment scenarios.

 \item \textbf{Ethical, Privacy, and Societal Considerations in VLA Deployment.}  
  Beyond technical robustness, the deployment of VLA systems raises important ethical and societal challenges that require governance-oriented solutions. Bias auditing tools are needed to identify skewed demographic or semantic distributions in training data, followed by corrective strategies such as adversarial debiasing and counterfactual data augmentation \cite{sahili2025scaling, zhang2025generative}. Privacy-preserving inference mechanisms, including on-device processing, homomorphic encryption for sensitive data streams, and differential privacy during training, are critical for safeguarding user data in domains such as healthcare and smart homes \cite{mumuni2025large, raza2025responsible}. Furthermore, transparent impact assessments, stakeholder engagement, and workforce upskilling initiatives can help manage socioeconomic effects, while regulatory frameworks and industry standards are essential to ensure accountability and responsible VLA adoption.

\end{itemize}

\subsection{Future Roadmap}
\label{subsec:future}

The future of VLA-based systems are expected to evolved at the intersection of increasingly capable multi-modal foundations, agentic reasoning, and embodied continual learning. Over the next decade, we anticipate several converging trends that will propel VLAs from capable yet brittle task specialists toward dependable, generalist robotic intelligence. However, this trajectory will be shaped by persistent constraints/limitations highlighted earlier: (i) real-time inference bottlenecks in closed-loop control, (ii) incomplete multi-modal action representations and weak safety assurance, (iii) dataset bias and grounding failures under distribution shift, (iv) integration complexity across perception-memory-reasoning-control, (v) high compute and energy demands that hinder edge deployment, and (vi) robustness, transparency, and ethical concerns in open-world settings. To address these issues holistically, Figure~\ref{fig:future} summarizes a system-level research roadmap, while Figure~\ref{fig:futuree} provides an intuitive conceptual illustration of how VLMs, VLA architectures, and agentic AI modules may co-evolve toward embodied AGI in robotics.

\begin{figure*}[ht!]
\centering
\resizebox{\linewidth}{!}{%
\begin{tikzpicture}[
    font=\Large, 
    node distance=11mm and 16mm, 
    every node/.style={draw, rounded corners, align=center, line width=0.7pt},
    root/.style={fill=gray!20, font=\bfseries\LARGE, minimum width=4.2cm, minimum height=1.30cm},
    pillar/.style={minimum width=4.1cm, minimum height=1.05cm, fill=white, font=\bfseries\Large},
    fut/.style={draw, rounded corners, line width=0.8pt, dashed, minimum width=3.9cm,
                minimum height=1.00cm, fill=white, font=\Large},
    edge/.style={->, line width=0.85pt}
]

\node[root] (future) {Future Roadmap};

\node[pillar, fill=blue!14, below left=of future, xshift=-10mm] (eff)
{Efficient\\Deployment};
\node[pillar, fill=orange!16, below=of future] (rel)
{Reliable \&\\Safe Intelligence};
\node[pillar, fill=green!16, below right=of future, xshift=10mm] (uni)
{Unified Systems\\\& Governance};

\draw[edge] (future) -- (eff);
\draw[edge] (future) -- (rel);
\draw[edge] (future) -- (uni);

\node[fut, below=of eff] (e1) {Parameter-efficient\\VLA backbones};
\node[fut, below=of e1] (e2) {Anytime / early-exit\\inference};
\node[fut, below=of e2] (e3) {Compact action\\tokenization \& chunking};
\node[fut, below=of e3] (e4) {On-device caching\\\& episodic memory};
\node[fut, below=of e4] (e5) {Hardware-aware\\compilation (edge)};

\draw[edge] (eff) -- (e1);
\draw[edge] (e1) -- (e2);
\draw[edge] (e2) -- (e3);
\draw[edge] (e3) -- (e4);
\draw[edge] (e4) -- (e5);

\node[fut, below=of rel] (r1) {Robust multimodal\\grounding (uncertainty)};
\node[fut, below=of r1] (r2) {Calibrated abstention\\\& safe fallback};
\node[fut, below=of r2] (r3) {World models:\\physical \& causal prediction};
\node[fut, below=of r3] (r4) {Constraint-aware control\\(shields/MPC)};
\node[fut, below=of r4] (r5) {Verification \& runtime\\safety monitors};

\draw[edge] (rel) -- (r1);
\draw[edge] (r1) -- (r2);
\draw[edge] (r2) -- (r3);
\draw[edge] (r3) -- (r4);
\draw[edge] (r4) -- (r5);

\node[fut, below=of uni] (u1) {Unified 2D--temporal--3D\\representations};
\node[fut, below=of u1] (u2) {Cross-embodiment\\transfer \& adaptation};
\node[fut, below=of u2] (u3) {Sim-to-real curricula\\\& domain randomization};
\node[fut, below=of u3] (u4) {Evaluation beyond success\\(safety, energy, recovery)};
\node[fut, below=of u4] (u5) {Governance: privacy,\\bias audits, accountability};

\draw[edge] (uni) -- (u1);
\draw[edge] (u1) -- (u2);
\draw[edge] (u2) -- (u3);
\draw[edge] (u3) -- (u4);
\draw[edge] (u4) -- (u5);

\draw[dotted, line width=0.9pt] (e5.south) -- ++(0,-7mm) -| (u5.south);
\draw[dotted, line width=0.9pt] (r5.south) -- ++(0,-7mm) -| (u5.south);
\draw[dotted, line width=0.9pt] (e4.east)  -- ++(8mm,0) |- (r3.west);

\end{tikzpicture}
} 

\caption{\textbf{Future research roadmap for Vision-Language-Action (VLA) models (Fig.~\ref{fig:future}).}
Dashed boxes denote prioritized directions addressing the core limitations of VLA models identified in this work: (i) \emph{Efficient Deployment} (latency, compute, energy), (ii) \emph{Reliable \& Safe Intelligence} (grounding, uncertainty, safety assurance), and (iii) \emph{Unified Systems \& Governance} (2D-temporal-3D integration, transfer, evaluation, and responsible deployment).}
\label{fig:future}
\end{figure*}

\paragraph{Multi-modal foundation models as the ``cortex'' for embodied perception:}
Today’s VLA stacks often rely on a vision-language backbone coupled to task-specific policy heads, which limits reuse of general knowledge and increases retraining cost across domains. A plausible next step is a unified multi-modal foundation model trained on web-scale image, video, text, and interaction/affordance traces to function as a shared ``cortex'' that encodes not only static semantics but also dynamics, contact priors, and commonsense physical knowledge \cite{zhou2025chatvla, zheng2025universal, zhan2026stable}. Such a cortex can reduce failure modes caused by shallow correlations by grounding language in object-centric representations and persistent scene structure \cite{singh2025neural}. As emphasized in Figure~\ref{fig:futuree}, this foundation-model cortex would enable robots to segment environments into actionable entities (objects, regions, affordances) and provide stable semantic anchors to downstream planners and controllers. Yet, to prevent overconfidence and hallucinated grounding, these models must incorporate calibrated uncertainty and evidence-linked reasoning to ensure that perception-driven plans remain verifiable under occlusion, clutter, and ambiguous instructions \cite{liu2026vision}.

\paragraph{Agentic, self-supervised, lifelong learning and continual adaptation:}
A defining limitation of current VLAs is their static nature: policies trained once are deployed unchanged, despite operating in non-stationary environments. Future VLAs should adopt agentic learning loops where models propose exploration objectives, hypothesize outcomes, and self-correct through simulated and real rollouts, enabling continual skill growth over months or years \cite{chowa2026language, holldack2026agentic}. This direction is expected to mitigate distribution shift, dataset bias, and long-horizon brittleness by allowing models to adapt over time; however, continual policy updates introduce new risks. In particular, repeated online or incremental learning can overwrite previously acquired competencies (catastrophic forgetting), induce unintended behavioral regressions, and increase susceptibility to noisy, adversarial, or unintended environmental feedback that may corrupt the learned policy \cite{zhou2025learning, yang2025recent}. Therefore, lifelong learning must be coupled with replay and safety-aware updates, modular adapters, and verification-informed policy revisions \cite{kim2025fine}. Within the roadmap of Figure~\ref{fig:future}, this agentic lifelong learning paradigm falls naturally at the intersection of Reliable \& Safe Intelligence and Unified Systems \& Governance, where continual learning is treated as a controlled, auditable lifecycle process rather than an ad hoc fine-tuning step.

\paragraph{Hierarchical, neuro-symbolic planning for scalability and interpretability:}
Scaling from low-level motor primitives to long-horizon objectives requires explicit hierarchy \cite{yang2025lohovla, yang2025guiding}. Next-generation VLA systems will likely use language-grounded planners (LLM-style modules fine-tuned for affordances and constraints) that decompose goals into structured sub-tasks, followed by mid-level skill policies and low-level controllers that ensure compliant motion \cite{hu2025vision, xu2025language}. This neuro-symbolic blend helps bridge integration complexity by imposing interfaces that are easier to debug, monitor, and certify \cite{shivadekar2025artificial}. Importantly, hierarchy also enables selective verification: high-level plans can be checked for constraint violations (unsafe steps, forbidden regions) \cite{schakkal2025hierarchical, gao2025vla}, while low-level trajectories can be shielded by control barrier functions, MPC, and runtime safety monitors \cite{sanyal2025asma, shadab2025comparison}. These components align with Figure~\ref{fig:future} under the pillar of \emph{Reliable \& Safe Intelligence}, where safety is enforced through both planning-time constraints and execution-time guards rather than post hoc evaluation alone \cite{huang2025building, zhang2025safevla}.

\paragraph{Real-time adaptation via world models and physical/causal reasoning:}
Robust deployment in unstructured settings demands that VLAs maintain internal predictive models of objects, contacts, and dynamics. World models that forecast near-term state transitions and failure likelihoods can support counterfactual evaluation (``if I push here, what collides?'') and rapid corrective actions when reality deviates from expectation (e.g., slipping grasp, unexpected friction) \cite{shukor2025smolvla}. This capability is central to safe manipulation, navigation, and human-robot interaction, where small errors compound quickly \cite{ma2024survey}. Yet, world models must be efficient enough for on-board use and consistent with multi-sensor evidence. Thus, a major future direction is hardware-aware, memory-efficient predictive modeling such as temporal token compression \cite{ye2025token, tan2025think}, event-driven state updates \cite{xu2025stare, vinod2025sebvs}, and hybrid physics-learning models that combine differentiable physical simulators with learned dynamics to enable control-relevant update rates \cite{huang2023diffvl, ding2021dynamic}. In Figure~\ref{fig:future}, these needs appear jointly in the \emph{Efficient Deployment} pillar (real-time constraints) and the \emph{Reliable \& Safe Intelligence} pillar (physics/causality for grounded decision-making).

\paragraph{Efficiency and scalability: bridging generality with edge deployment:}
A central barrier to VLA adoption remains the mismatch between the computational footprint of large multi-modal backbones and the latency/energy constraints of closed-loop control \cite{yu2025survey, wei2025focus}. Future VLAs should prioritize parameter-efficient designs (structured sparsity, low-rank adaptation, modular experts) that preserve generalization while reducing inference cost \cite{fang2025sqap, shao2025large}. Beyond training-time efficiency, anytime/early-exit policies can allocate computation adaptively, ensuring that safety-critical steps retain high fidelity while routine steps use cheaper pathways \cite{chi2025impromptu, zhang2025safevla}. Equally important is action-space efficiency: compact action tokenization and chunked control representations shorten autoregressive horizons, enabling higher control rates without sacrificing temporal smoothness \cite{pertsch2025fast}. These model-level choices must be paired with hardware-aware compilation across GPU/NPU/edge accelerators, including quantization-aware scheduling and memory-optimized attention kernels \cite{fang2025sqap, park2025saliency}. Compute-aware caching and episodic memory further reduce redundant forward passes and improve responsiveness in long-horizon tasks \cite{chang2025survey, li2025seeing}. Collectively, these directions operationalize the \emph{Efficient Deployment} pillar of Figure~\ref{fig:future} and directly address the compute/energy limitations emphasized earlier.

\paragraph{Cross-embodiment transfer and morphology-agnostic skill representations:}
The approach of training separate VLAs for each robot morphology is unlikely to scale. A key future theme is embodiment-agnostic policy learning, where skills are expressed in abstract action spaces (e.g., contact goals, affordance-point manipulation, task-space constraints) that transfer across wheeled platforms, quadrupeds, and humanoids \cite{zhang2025pure, kawaharazuka2025vision}. Meta-learning and few-shot calibration can allow rapid bootstrapping on new robots with minutes of data rather than weeks of training \cite{chen2025vision}. This direction also mitigates dataset bias by enforcing invariances across embodiments and environments, but it demands standardized representations, common interfaces, and reproducible evaluation protocols \cite{kawaharazuka2025vision, zhang2025pure}. In Figure~\ref{fig:future}, this embodiment-agnostic skill learning paradigm sits under \emph{Unified Systems \& Governance}, linking architectural unification with principled transfer and benchmarking.

\paragraph{Evaluation beyond task success: safety, recovery, and resource-aware metrics:}
Progress in VLA requires measurement that reflects deployment realities. Task success alone does not clearly represent failure severity, temporal inconsistency, unsafe near-misses, and energy inefficiency. Future benchmarks should quantify safety violations, uncertainty calibration, recovery behavior, temporal coherence, energy consumption, and downstream utility under human constraints \cite{zhang2025safevla, gu2025safe}. Moreover, evaluations should report compute budgets, dataset composition, and deployment conditions to enable fair comparisons and diagnose bias-driven gains \cite{taherin2025cross, zhang2025vla, kawaharazuka2025vision}. Such measurement is not merely scientific hygiene: it is the foundation for auditing and governance, and it determines whether a system can be responsibly deployed at scale \cite{zhang2025safevla, zhang2025pure}. This motivation underlies the evaluation branch of Figure~\ref{fig:future} and complements the conceptual deployment narrative in Figure~\ref{fig:futuree}.

\paragraph{Safety, ethics, and human-centered alignment as first-class design objectives:}
As VLAs gain autonomy, built-in safety and value alignment become critical. Future systems should integrate real-time risk estimators that assess potential harm before executing high-risk actions, request natural-language confirmation under ambiguity, and maintain transparent logs for accountability \cite{zhang2025safevla, kawaharazuka2025vision}. Privacy-aware sensing, bias audits, and human-in-the-loop oversight must be embedded into the lifecycle, especially for assistive robotics and safety-critical autonomy \cite{zhang2025through, wang2025navigating}. Regulatory-aligned evaluation protocols and standardization efforts will be essential to translate VLA advances into trusted real-world systems \cite{pasas2025discovery, zhou2025autovla, wang2025rad}. This governance perspective is explicitly captured in Figure~\ref{fig:future} and is implicitly reflected by the socially aware humanoid setting illustrated in Figure~\ref{fig:futuree}.

\paragraph{Cross-cutting themes: continual learning, failure recovery, interaction, and control fidelity:}
Across all pillars in Figure~\ref{fig:future}, several cross-cutting themes are expected to shape the next decade of VLA research. First, continual and lifelong learning must be safe, auditable, and resistant to forgetting, enabling long-term adaptation without destabilizing deployed systems \cite{zheng2026lifelong}. Second, failure detection and recovery should be treated as first-class capabilities, incorporating introspective monitoring, uncertainty-aware perception, and structured recovery behaviors when execution deviates from expected outcomes \cite{karli2025ask, yang2025upl}. Third, improving the precision and reliability of action generation remains critical: while VLA-based planners enable flexible, language-conditioned decision making, their trajectory accuracy and control stability currently lag behind conventional analytical approaches such as model predictive control, sampling-based motion planning, and feedback-linearized controllers. As a result, hybrid architectures that combine VLA-driven high-level planning with classical or learned low-level controllers are expected to play a central role in achieving both semantic flexibility and control-level precision. Finally, human alignment and interaction require mechanisms for intent clarification, shared autonomy, and explainable action rationales, supporting trust and usability across diverse real-world settings \cite{jiang2025survey, jiang2025solami}.

In summary, Figure~\ref{fig:future} emphasizes that closing the gap between laboratory demonstrations and robust real-world deployment will require coordinated advances in efficiency, safety, data and generalization, system integration, evaluation, and governance. Complementarily, Figure~\ref{fig:futuree} illustrates how these advances may converge toward generalist embodied agents across multiple platforms including mobile robots, manipulators, assistive systems, and humanoids where multi-modal perception, hierarchical planning, continual adaptation, and human-aligned safety are integrated within a unified intelligence stack. Addressing these directions collectively is expected to transform VLAs from promising research prototypes into dependable, broadly applicable embodied systems rather than solutions limited to any single robot morphology.

\section{Conclusion}
In this comprehensive review, we systematically evaluated the recent developments, methodologies, and applications of  Vision-Language-Action (VLA)  models published over the last three years. Our analysis began with the foundational concepts of VLAs, defining their role as multi-modal systems that unify visual perception, natural language understanding, and action generation in physical or simulated environments. We traced their evolution and timeline, detailing key milestones that marked the transition from isolated perception-action modules to fully unified, instruction-following robotic agents. We highlighted how multi-modal integration has matured from loosely coupled pipelines to transformer-based architectures that enable seamless coordination between modalities.

Next, we examined tokenization and representation techniques, focusing on how VLAs encode visual and linguistic information, including action primitives and spatial semantics. We explored learning paradigms, detailing the datasets and training strategies from supervised learning and imitation learning to reinforcement learning and multi-modal pretraining that have shaped VLA performance. In 'adaptive control and real-time execution' section, we discussed how modern VLAs are optimized for dynamic environments, analyzing policies that support latency-sensitive tasks. We then categorized major architectural innovations, surveying over 50 recent VLA models. This discussion included advancements in model design, memory systems, and interaction fidelity. We further studied strategies for training efficiency improvement, including parameter-efficient methods like LoRA, quantization, and model pruning, alongside acceleration techniques such as parallel decoding and hardware-aware inference. Our analysis of real-world applications highlighted both the promise and current limitations of VLA models across six domains: humanoid robotics, autonomous vehicles, industrial automation, healthcare, agriculture, and augmented reality (AR) navigation. Across these settings, VLAs demonstrated strong capabilities in high-level semantic reasoning, instruction-following, and task generalization, particularly in structured or partially controlled environments. However, their effectiveness was often constrained by real-time inference latency, limited robustness under environmental variability, and reduced precision in long-horizon or safety-critical control when compared to conventional analytical planning and control pipelines. Moreover, application-specific adaptations and extensive data curation were frequently required to achieve reliable performance, underscoring challenges in scalability and deployment. These findings suggest that while VLAs are well-suited for semantic decision making and flexible task specification, hybrid architectures that integrate VLA reasoning with classical or learned low-level controllers remain essential for practical, real-world operation.

In addressing challenges and limitations, we focused on five core areas: real-time inference, multi-modal action representation and safety, bias and generalization, system integration and compute constraints, and ethical deployment. We proposed potential solutions drawn from current literature, including model compression, cross-modal grounding, domain adaptation, and agentic learning frameworks. Finally, our discussion and future roadmap articulated how the convergence of VLMs, VLA architectures, and agentic AI systems is steering robotics toward artificial general intelligence (AGI). This review provides a unified understanding of VLA advancements, identifies unresolved challenges, and outlines a structured path forward for developing intelligent, embodied, and human-aligned agents in the future.

\section*{Funding Declaration}
This work was supported in part by the National Science Foundation (NSF) and the United States Department of Agriculture (USDA), National Institute of Food and Agriculture (NIFA), through the ‘‘Artificial Intelligence (AI) Institute for Agriculture’’ program under Award Numbers AWD003473 and AWD004595, and USDA-NIFA Accession Number 1029004 for the project titled ‘‘Robotic Blossom Thinning with Soft Manipulators.’’ Additional support was provided through USDA/NIFA Grant Number 2024-67022-41788, Accession Number 1031712, under the project ‘‘ExPanding UCF AI Research To Novel Agricultural EngineeRing Applications (PARTNER).

\section*{Declarations}
The authors declare no conflicts of interest.

\section*{Statement on AI Writing Assistance}
ChatGPT and Perplexity were utilized to enhance grammatical accuracy and refine sentence structure; all AI-generated revisions were thoroughly reviewed and edited for relevance. 
\bibliographystyle{elsarticle-harv} 
\bibliography{example}


\end{document}